\documentclass[12pt,british]{book}
\usepackage[T1]{fontenc}
\usepackage[latin9]{inputenc}
\usepackage{geometry}
\usepackage{verbatim}
\usepackage{float}
\usepackage{amssymb}
\usepackage{graphicx}
\usepackage{setspace}
\makeatletter

\providecommand{\tabularnewline}{\\}
\floatstyle{ruled}
\newfloat{algorithm}{tbp}{loa}[chapter]
\providecommand{\algorithmname}{Algorithm}
\floatname{algorithm}{\protect\algorithmname}

\usepackage{times}
\usepackage{algorithmic}
\usepackage{setspace}

\@ifundefined{showcaptionsetup}{}{%
 \PassOptionsToPackage{caption=false}{subfig}}
\usepackage{subfig}
\makeatother

\usepackage{babel}
\begin{document}
\begin{titlepage}\centering{\includegraphics[width=0.4\columnwidth]{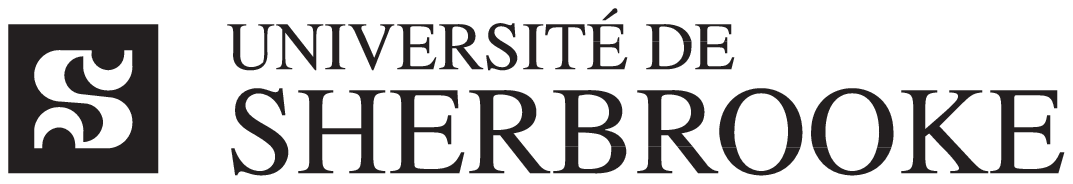}

Faculté de génie

Génie électrique et informatique

\vfill{}

{\Large{}Auditory System for a Mobile Robot}{\Large \par}

\vfill{}

Thèse de doctorat

Specialité: génie électrique

\vfill{}

Jean-Marc Valin

\vfill{}

}

Sherbrooke (Quebec) Canada\hfill{}10 août 2005

\end{titlepage}\thispagestyle{empty}\cleardoublepage

\pagenumbering{roman}

\begin{center}\large{\textbf{Abstract}}\end{center}\vspace{0.5cm}

\begin{comment}
Abstract
\end{comment}

The auditory system of living creatures provides useful information
about the world, such as the location and interpretation of sound
sources. For humans, it means to be able to focus one's attention
on events, such as a phone ringing, a vehicle honking, a person taking,
etc. For those who do not suffer from hearing impairments, it is hard
to imagine a day without being able to hear, especially in a very
dynamic and unpredictable world. Mobile robots would also benefit
greatly from having auditory capabilities. 

In this thesis, we propose an artificial auditory system that gives
a robot the ability to locate and track sounds, as well as to separate
simultaneous sound sources and recognising simultaneous speech. We
demonstrate that it is possible to implement these capabilities using
an array of microphones, without trying to imitate the human auditory
system. The sound source localisation and tracking algorithm uses
a steered beamformer to locate sources, which are then tracked using
a multi-source particle filter. Separation of simultaneous sound sources
is achieved using a variant of the Geometric Source Separation (GSS)
algorithm, combined with a multi-source post-filter that further reduces
noise, interference and reverberation. Speech recognition is performed
on separated sources, either directly or by using Missing Feature
Theory (MFT) to estimate the reliability of the speech features.

The results obtained show that it is possible to track up to four
simultaneous sound sources, even in noisy and reverberant environments.
Real-time control of the robot following a sound source is also demonstrated.
The sound source separation approach we propose is able to achieve
a 13.7 dB improvement in signal-to-noise ratio compared to a single
microphone when three speakers are present. In these conditions, the
system demonstrates more than 80\% accuracy on digit recognition,
higher than most human listeners could obtain in our small case study
when recognising only one of these sources. All these new capabilities
will allow humans to interact more naturally with a mobile robot in
real life settings.

\cleardoublepage

\begin{spacing}{1.45}

\begin{center}\large{\textbf{Sommaire}}\end{center}\vspace{0.5cm}

\begin{comment}
Résumé
\end{comment}

La plupart des êtres vivants possèdent un système auditif leur fournissant
de l'information utile sur leur environnement, comme la direction
et l'interprétation des source sonores. Cela permet de diriger notre
attention sur des événements, comme une sonnerie de téléphone, le
klaxon d'un véhicule, une personne qui parle, etc. Pour ceux qui ne
souffrent pas d'un problème auditif, il est difficile d'imaginer passer
une journée sans pouvoir entendre, particulièrement dans un environnement
dynamique et imprévisible. Les robots mobiles pourraient eux aussi
tirer un bénéfice important de capacités auditives.

Dans cette thèse, nous proposons un système d'audition artificielle
dotant un robot de la capacité de localiser et suivre des sons, ainsi
que de la capacité de séparer des sources sonores sonores simultanées
et de reconnaître ce qui est dit. Nous démontrons qu'il est possible
de réaliser ces capacités à l'aide d'un réseau de microphones, sans
la nécessité d'imiter le système auditif humain. L'algorithme de localisation
et suivi de sources sonores utilise un \emph{beamformer} dirigé pour
localiser les sources, qui sont ensuite suivies en utilisant un filtre
particulaire. La séparation de sources sonores simultanées est accomplie
par une variante de l'algorithme de séparation géometrique des sources
(\emph{Geometric Source Separation}), combinée à un post-traitement
multi-source permettant de réduire le bruit, les interférences et
la réverbération. Les sources séparées sont ensuite utilisées pour
effectuer la reconnaissance vocale, soit directement, soit en utilisant
la théorie des données manquantes (\emph{Missing Feature Theory})
qui tient compte de la fiabilité des paramètres de la parole séparée.

Les résultats obtenus montrent qu'il est possible de suivre jusqu'à
quatre sources sonores simultanées, même dans des environnements bruités
et réverbérants. Nous démontrons aussi le contrôle en temps réel d'un
robot se déplaçant pour suivre une source en mouvement. De plus, lorsque
trois personnes parlent en même temps, l'approche proposée pour la
séparation des sources sonores permet d'améliorer de 13.7 dB le rapport
signal-à-bruit si on compare à l'utilisation d'un seul microphone.
Dans ces conditions, le système obtient un taux de reconnaissance
supérieur à 80\% sur des chiffres, ce qui est supérieur à ce que la
plupart des auditeurs humains ont obtenu lors de tests simples de
reconnaissance sur une seule de ces sources. Ces nouvelles capacités
auditives permettront aux humains d'interagir de façon plus naturelle
avec un robot mobile.

\end{spacing}

\cleardoublepage

\begin{center}\large{\textbf{Acknowledgements}}\end{center}\vspace{0.5cm}

\begin{comment}
Acknowledgements
\end{comment}

I would like to express my gratitude to my thesis supervisor Fran\c{c}ois
Michaud and my thesis co-supervisor Jean Rouat for their support and
feedback during the course of this thesis. Their knowledge and insight
have greatly contributed to the success of this project.

During this research, I was supported financially by the National
Science and Engineering Research Council of Canada (NSERC), the Quebec
\emph{Fonds de recherche sur la nature et les technologies} (FQRNT)
and the Universit\'{e} de Sherbrooke. François Michaud holds the
Canada Research Chair (CRC) in Mobile Robotics and Autonomous Intelligent
Systems. This research is supported financially by the CRC Program
and the Canadian Foundation for Innovation (CFI). 

I would like to thank Hiroshi G. Okuno for receiving me at the Kyoto
University Speech Media Processing Group for an internship from August
to November 2004, during which I was introduced to the missing feature
theory. During this stay, I was supported financially by the Japan
Society for the Promotion of Science (JSPS) short-term exchange student
scholarship.

Special thanks to Brahim Hadjou for help formalising the particle
filtering notation, to Dominic L\'{e}tourneau and Pierre Lepage for
implementing what was necessary to control the robot in real time
using my algorithms and to Yannick Brosseau for performing speech
recognition experiments. Thanks to Serge Caron and Jonathan Bisson
for the design and fabrication of the microphones. Many thanks to
everyone at the Mobile Robotics and Intelligent Systems Research Laboratory
(LABORIUS) who took part in my numerous experiments. I would like
to thank Nuance Corporation for providing us with a free license to
use its Nuance Voice Platform.

I am grateful to my wife, Nathalie, for her love, encouragement and
unconditional support throughout my PhD, and to my son, Alexandre,
for giving me a reason to finish on schedule!

\tableofcontents{}

\listoffigures

\listoftables

\listof{algorithm}{List of Algorithms}

\chapter*{Lexicon}

\begin{comment}
\singlespacing
\end{comment}

\onehalfspacing
\begin{description}
\item [{ASR.}] Automatic Speech Recognition.
\item [{CMS.}] Cepstral Mean Subtraction (also cepstral mean normalisation).
\item [{DCT.}] Discrete Cosine Transform.
\item [{DSP.}] Digital Signal Processor.
\item [{FFT.}] Fast Fourier Transform.
\item [{GSS.}] Geometric Source Separation.
\item [{LSS.}] Linear Source Separation.
\item [{MCRA.}] Minima-Controlled Recursive Average.
\item [{MFCC.}] Mel-Frequency Cepstral Coefficient.
\item [{MFT.}] Missing Feature Theory.
\item [{MMSE.}] Minimum Mean Square Error.
\item [{pdf.}] Probability density function.
\item [{Reverberation~Time~($T_{60}$).}] Time for reverberation to decrease
by 60 dB.
\item [{SNR.}] Signal-to-Noise Ratio.
\item [{SRR.}] Signal-to-Reverberant Ratio.
\item [{TDOA.}] Time Delay of Arrival.
\end{description}

\chapter{Overview of the Thesis\label{cha:Overview-of-the-Thesis}}

\doublespacing\pagenumbering{arabic}

The auditory system of living creatures provides significant information
about the world, such as the location and the interpretation of sound
sources. For humans, it means to be able to focus one's attention
on events, such as a phone ringing, a vehicle honking, a person taking,
etc. For those who do not suffer from hearing impairments, it is hard
to imagine a day without being able to hear, especially in a very
dynamic and unpredictable world. 

Brooks \emph{et al.} \cite{Brooks} postulate that the essence of
intelligence lies in four main aspects: development, social interaction,
embodiment, and integration. It turns out that hearing plays a role
in three of these factors. Obviously, speech is by far the preferred
means of communication between humans. Hearing also plays a very important
role in human development. Marschark \cite{Marschark} even suggests
that although deaf children have similar IQ results compared to other
children, they do experience more learning difficulties in school.
Hearing is useful for the integration aspect of intelligence, as it
complements well other sensors such as vision by being omni-directional,
capable of working in the dark and not limited by physical structures
(such as walls). Therefore, it is important for robots to understand
spoken language and respond to auditory events. The auditory capabilities
will undoubtly improve the intelligence manifested by autonomous robots. 

The human hearing sense is very good at focusing on a single source
of interest and following a conversation even when several people
are speaking at the same time. We refer to this situation as the \emph{cocktail
party effect}. In order to operate in human and natural settings,
autonomous mobile robots should be able to do the same. This means
that a mobile robot should be able to separate and recognise sound
sources present in the environment at any time. This requires the
robots not only to detect sounds, but also to locate their origin,
separate the different sound sources (since sounds may occur simultaneously),
and process all of this data to be able to extract useful information
about the world. 

While it is desirable for an artificial auditory system to achieve
the same level of performance as the human auditory system, it is
not necessary for the approach taken to mimic the human auditory system.
Unlike humans, robots are not inherently limited to having only two
ears (microphones) and do not have the human limitations due to psychoacoustic
masking \cite{Ehmer1958} or partial insensitivity to phase. On the
other hand, the human brain is far more complex and powerful than
what can be achieved today in terms of artificial intelligence. So
it may be appropriate to compensate for this fact by adding more sensors.
Using more than two microphones provides increased resolution in three-dimensional
space. This also means increased robustness, since multiple signals
greatly helps reduce the effects of noise and improve discrimination
of multiple sound sources.

The primary objective of this thesis\footnote{Up-to-date information, as well as audio and video clips related to
the project are available at http://www.gel.usherb.ca/laborius/projects/Audible/} is to give a mobile robot auditory capabilities using an array of
microphones. We focus on three main capabilities:
\begin{enumerate}
\item Localising (potentially simultaneous) sound sources and being able
to track them over time;
\item Separating simultaneous sound sources from each other so they can
be analysed separately;
\item Performing speech recognition on separated sources.
\end{enumerate}
While some of these aspects have already been studied in the field
of signal processing, they are revisited here with a special focus
on mobile robotics and its constraints:
\begin{itemize}
\item \textbf{Limited computational capabilities}. A mobile robot has only
limited on-board computing power. For that reason, we impose that
our system be able to run on a low-power, embedded PC computer.
\item \textbf{Need for real-time processing with reasonable delay}. In order
for the extracted auditory information to be useful, it must be available
at the same time as the events are happening. For that reason, the
processing delay must be kept small. A certain delay is still acceptable,
but it depends on the particular information needed.
\item \textbf{Weight and space constraints}. A mobile robot usually has
strict weight and space constraints. The auditory system we propose
must respect the constraints imposed by the robot and not the contrary.
In practice, this means that the microphones must be placed on the
robot wherever possible. 
\item \textbf{Noisy operating environment (both point source and diffuse
sources, robot noise)}. Mobile robots are set to evolve in a wide
range of environments. Some of these environments are likely to be
noisy and reverberant. The noise may take many different forms. It
can be a stationary, diffuse noise, such as the room ventilation system
or a number of machines in the room. Noise can also be in the form
of people talking around the robot (not to the robot). In that case,
the noise is non-stationary, usually a point source, and referred
to as interference. Also, every environment is reverberant to a certain
degree. While this is usually not a problem in small rooms, our system
will certainly have to be robust to reverberation (the echo in a room)
when a robot is to operate in larger, reverberant rooms.
\item \textbf{Mobile sound sources}. A mobile robot usually evolves in a
dynamically changing environment. That implies that sound sources
around it will appear, disappear, and move around the robot. 
\item \textbf{Mobile reference system (robot can move)}. Not only can sound
sources move around the robot, but the robot itself should be able
to move in its environment. The auditory system must remain functional
even when the robot is moving.
\item \textbf{Adaptability}. It is highly desirable for the auditory system
we develop to be easily adapted to any mobile robot. For this reason,
there should be as few adjustments as possible that depend on the
robot's shape.
\end{itemize}

\section{Audition in Mobile Robotics}

Artificial hearing for robots is a research topic still in its infancy,
at least when compared to all the work already done on artificial
vision in robotics. However, the field of artificial audition has
been the subject of much research in recent years. In 2004, the IEEE/RSJ
International Conference on Intelligent Robots and Systems (IROS)
has for the first time included a special session on robot audition.
Initial work on sound localisation by Irie \cite{irie95robust} for
the Cog \cite{Brooks99} and Kismet robots can be found as early as
1995. The capabilities implemented were however very limited, partly
because of the necessity to overcome hardware limitations.

The SIG robot\footnote{http://winnie.kuis.kyoto-u.ac.jp/SIG/oldsig/}
and its successor, SIG2\footnote{http://winnie.kuis.kyoto-u.ac.jp/SIG/},
both developed at Kyoto University, have integrated increasing auditory
capabilities \cite{Nakadai00-AAAI,Nakadai,nakadai-hidai-okuno-kitano2001,OKUNO,nakadai-okuno-kitano2002,nakadai-okuno-kitano2002',nakadai-matsuura-okuno-kitano2003}
over the years (from 2000 to now). Both robots are based on binaural
audition, which is still the most common form of artificial audition
on mobile robots. Original work by Nakadai \emph{et al.} \cite{Nakadai00-AAAI,Nakadai}
on active audition have made it possible to locate sound sources in
the horizontal plane using binaural audition and active behaviour
to disambiguate front from rear. Later work has focused more on sound
source separation \cite{nakadai-okuno-kitano2002,nakadai-okuno-kitano2002'}
and speech recognition \cite{Yamamoto04a,Yamamoto04b}. 

The ROBITA robot, designed at Waseda University, uses two microphones
to follow a conversation between two people, originally requiring
each participant to wear a headset \cite{matsusaka-tojo-kubota-furukawa-tamiya-hayata-nakano-kobayashi99},
although a more recent version uses binaural audition \cite{Matsusaka2001}.

A completely different approach is used by Zhang and Weng \cite{ZHANG}
with the SAIL robot with the goal of making a robot develop auditory
capabilities autonomously. In this case, the \emph{Q-learning} unsupervised
learning algorithm is used instead of supervised learning, which is
most commonly used in the field of speech recognition. The approach
is validated by making the robot learn simple voice commands. Although
current speech recognition accuracy using conventional methods is
usually higher than the results obtained, the advantage is that the
robot learns words autonomously.

More recently, robots have started taking advantage of using more
than two microphones. This is the case of the Sony QRIO SDR-4XII robot
\cite{Fujita03} that features seven microphones. Unfortunately, little
information is available regarding the processing done with those
microphones. A service robot by Choi \emph{et al.} \cite{choi-kong-kim-bang2003}
uses eight microphones organised in a circular array to perform speech
enhancement and recognition. The enhancement is provided by an adaptive
beamforming algorithm. Work by Asano, Asoh, \emph{et al.} \cite{Asoh97,Asano99,Asano2001}
also uses a circular array composed of eight microphones on a mobile
robot to perform both localisation and separation of sound sources.
In more recent work \cite{Asoh2004}, particle filtering is used to
integrate vision and audition in order to track sound sources.

In general, human-robot interface is a popular area of audition-related
research in robotics. Works on robot audition for human-robot interface
has also been done by Prodanov \emph{et al.} \cite{Prodanov} and
Theobalt \emph{et al.} \cite{Theobalt}, based on a single microphone
near the speaker. Even though human-robot interface is the most common
goal of robot audition research, there is research being conducted
for other goals. Huang \emph{et al.} \cite{Huang1999Navig} use binaural
audition to help robots navigate in their environment, allowing a
mobile robot to move toward sound-emitting objects without colliding
with those object. The approach even works when those objects are
not visible (i.e., not in line of sight), which is an advantage over
vision.

It is possible to use audition to determine the size and characteristics
of a room, as proposed by Tesch and Zimmer \cite{Tesch}. The approach
uses a single microphone and can be used for localisation purposes
if the room has been visited before. The system works by emitting
wideband noise and measuring reverberation characteristics.

Robot audition even has military applications, where a robot uses
for instance an array of eight microphones to detect impulsive noise
events in order to locate a sniper weapon \cite{Young2003}. The approach
is based on impulse time detection and the localisation is performed
using Time Delay of Arrival (TDOA) estimates. The authors claim to
be working on a 16-microphone version of their system.

Audition has also been applied to groups of robot\footnote{http://www.inel.gov/featurestories/12-01robots.shtml}
at the Idaho National Engineering and Environmental Laboratory (INEEL).
In that case, the robots use chirp sounds and audition to communicate
with each other in a simple manner. The robots themselves have very
limited auditory capabilities, but the goal is to include a large
number of robots. Unfortunately, little information is available about
the exact auditory capabilities of the robot.

Although not strictly related to audition, work has been done to allow
robots to show emotions when talking \cite{Breazeal01}. This is done
by varying speech characteristics such as pitch, rate and intensity.
If combined with auditory capabilities, this could enhance robot interactions
with humans.

\section{Experimental Setup\label{sec:Experimental-Setup}}

\begin{figure}[th]
\begin{center}\subfloat[Electret with pre-amplifier (C1)]{

\includegraphics[height=0.2\paperwidth]{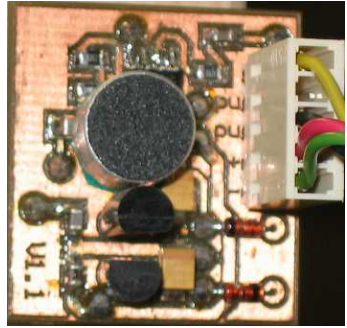}}$\qquad$\subfloat[Electret installed directly on the robot frame (C2)]{

\includegraphics[height=0.2\paperwidth]{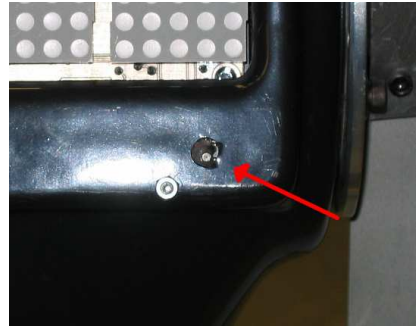}}\end{center}

\caption{Microphones used on the Spartacus robot.\label{cap:Microphones-used}}
\end{figure}

\begin{figure}[th]
\center{\subfloat[Open array configuration (C1)]{

\includegraphics[height=9cm]{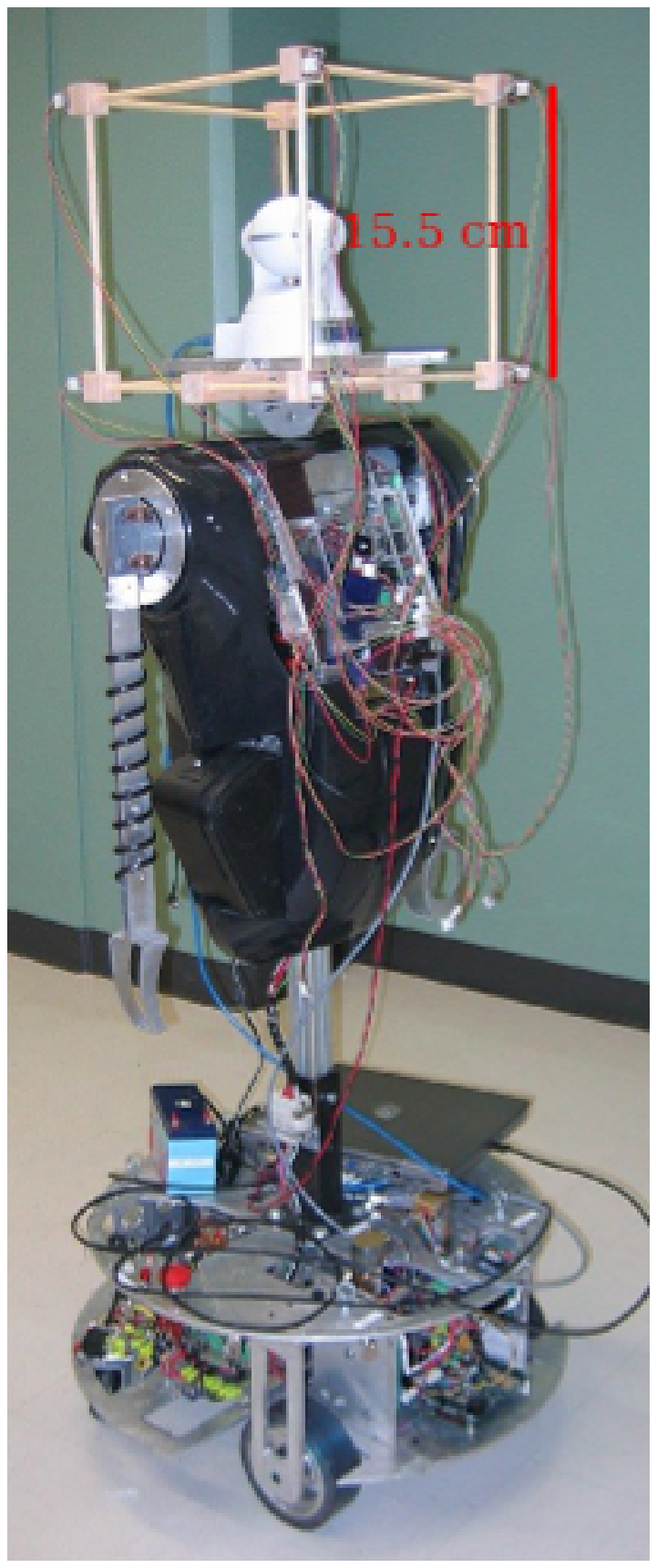}}$\;$\subfloat[Closed array configuration (C2)]{

\includegraphics[height=9cm]{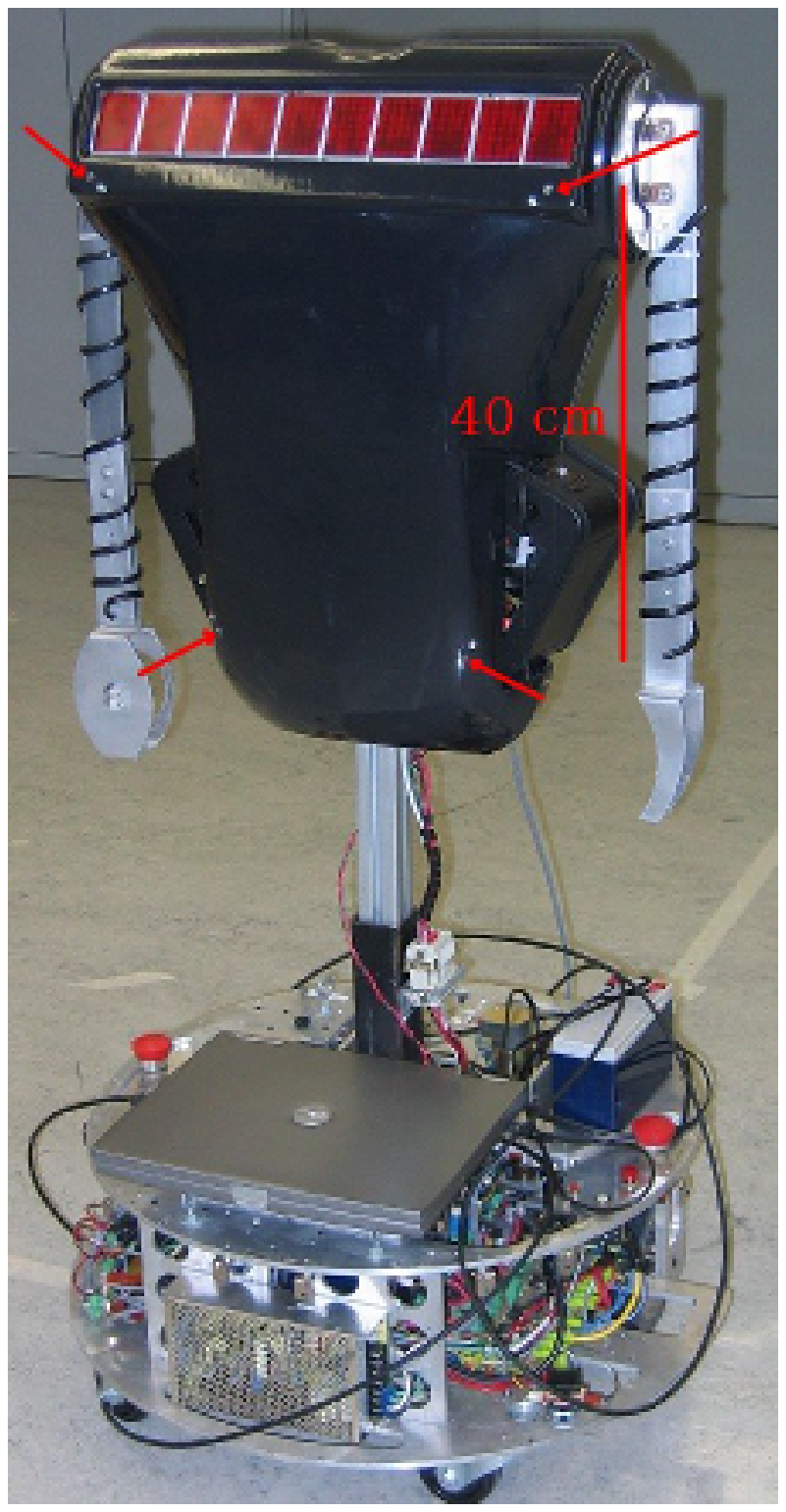}}}

\caption{Spartacus robot in configurations\label{cap:Spartacus-robot}.}
\end{figure}

The proposed artificial auditory system is tested using an array of
omni-directional microphones, each composed of an electret cartridge
mounted on a simple custom pre-amplifier (as shown in Figure \ref{cap:Microphones-used}a).
The number of microphones is set to eight, as it is the maximum number
of analog input channels on commercially available soundcards. Two
array configurations are used for the evaluation of the system. The
first configuration (C1) is an open array and consists of inexpensive
($\sim$US\$1 each) microphones arranged on the vertices of a 16 cm
cube mounted on top of the \emph{Spartacus} robot (shown in Figure
\ref{cap:Spartacus-robot}a). The second configuration (C2) is a closed
array and uses smaller, middle-range ($\sim$US\$20 each) microphones,
placed through holes (Figure \ref{cap:Microphones-used}b) at different
locations on the body of the robot (shown in Figure \ref{cap:Spartacus-robot}b).
Although we are mainly interested in the C2 configuration because
it is the least intrusive, the C1 configuration is used to demonstrate
the validity of some of the hypotheses we make, as well as the adaptability
of the system. It is reported that better localisation and separation
results are obtained when maximising the distance between the microphones
\cite{Rabinkin}. So, given the fact that we desire uniform performance
regardless of the location of the sources, we spread the microphones
evenly (and as far away from each other as possible) on the robot
surface.

For both arrays, all channels are sampled simultaneously using an
RME Hammerfall Multiface DSP connected to a notebook computer (Pentium-M
1.6 GHz CPU) through a CardBus interface. The software part is implemented
within the FlowDesigner\footnote{\texttt{http://flowdesigner.sourceforge.net/}}
environment and is designed to be connected to other components of
the robot through the MARIE\footnote{\texttt{http://marie.sourceforge.net/}}
framework, as described in \cite{Cote2004}.

Experiments are performed in two different environments. The first
environment (E1) is a medium-size room (10 m $\times$ 11 m, 2.5 m
ceiling) with a reverberation time (-60 dB) of 350 ms and moderate
noise (ventilation, computers). The second environment (E2) is a hall
(16 m $\times$ 17 m, 3.1 m ceiling, connected to other rooms) with
a reverberation time of approximately 1.0 s and a high level of background
noise. While E2 is a public area, we tried to minimise the noise (by
not having people close to the experimental setup) made by other people
in order to make the experiments repeatable.

\section{Thesis Outline}

\begin{figure}[th]
\includegraphics[width=1\columnwidth]{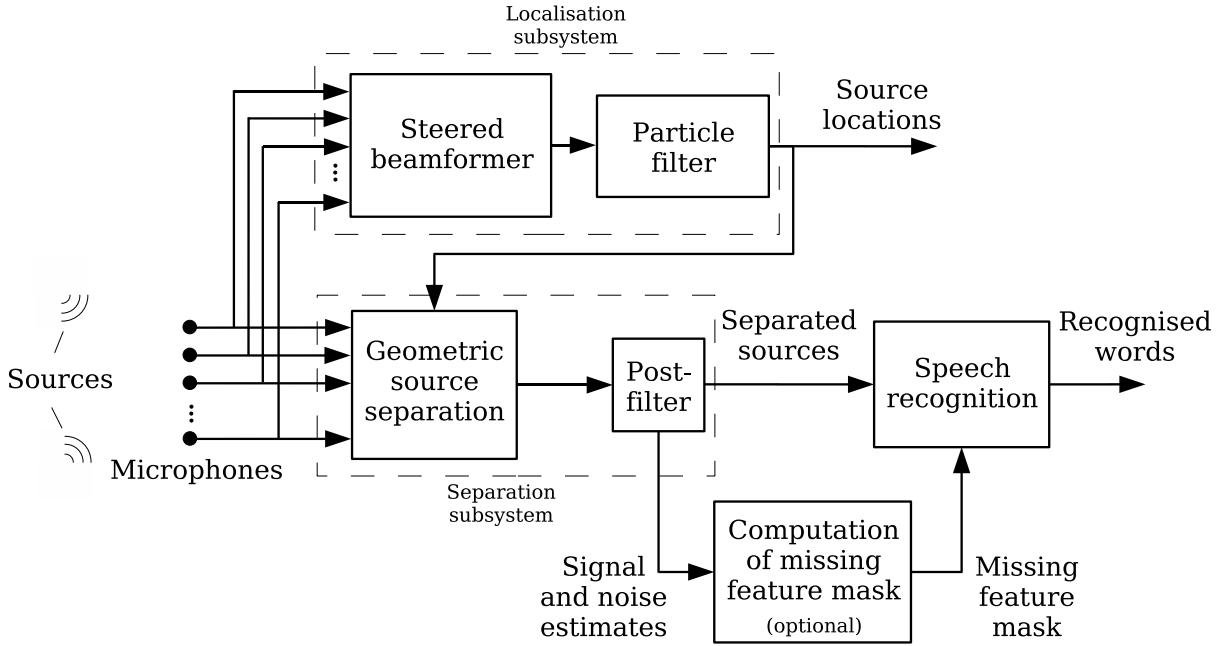}

\caption{Overview of the proposed artificial auditory system.\label{cap:Overview-global-system}}
\end{figure}

This thesis is divided in three main parts, corresponding to the three
auditory capabilities identified in the introduction: sound source
localisation, sound source separation, and integration with speech
recognition. A block diagram of the complete artificial auditory system
is shown in Figure \ref{cap:Overview-global-system}.

Chapter \ref{cha:Sound-Source-Localization} describes the localisation
subsystem. The method is based on a frequency-domain implementation
of a steered beamformer along with a particle filter-based tracking
algorithm. Results show that a mobile robot can localise and track
in real-time multiple moving sources and can detect most sound sources
reliably over a range of 7 meters. These new capabilities allow a
mobile robot to interact with people in real life settings.

Chapter \ref{cha:Sound-Source-Separation} presents a method for separating
simultaneous sound sources for the separation subsystem. The microphone
array and the sound localisation data are used in a real-time implementation
of the Geometric Source Separation \cite{Parra2002Geometric} algorithm.
A multi-source post-filter is also developed in order to further reduce
interferences from other sources as well as background noise. The
main advantage of our approach for mobile robots resides in the fact
that both the frequency-domain Geometric Source Separation algorithm
and the post-filter are able to adapt rapidly to new sources and non-stationarity.
Separation results are presented for three simultaneous interfering
speakers in the presence of noise. A reduction of log spectral distortion
(LSD) and increase of signal-to-noise ratio (SNR) of approximately
8.9~dB and 13.7~dB are observed. When the source of interest is
silent, background noise and interference is reduced by 24.5~dB.

Chapter \ref{cha:Speech-Recognition-Integration} demonstrates the
use of the sound separation system to give a mobile robot the ability
to perform automatic speech recognition with simultaneous speakers.
Two different configurations for speech recognition are tested. In
one configuration, the separated audio is sent directly to an automatic
speech recogniser (ASR). In the second configuration, the post-filter
described in Chapter \ref{cha:Sound-Source-Separation} is used to
estimate the reliability of spectral features and compute a missing
feature mask that provides the ASR with information about the reliability
of spectral features. We show that it is possible to perform speech
recognition for up to three simultaneous speakers, even in a reverberant
environment.

Chapter \ref{cha:Conclusion} concludes this thesis by summarising
the results obtained, listing possible future work, and suggesting
applications of this work to other fields.

\chapter{Sound Source Localisation\label{cha:Sound-Source-Localization}}

Sound source localisation is defined as the determination of the coordinates
of sound sources in relation to a point in space. To perform sound
localisation, our brain combines timing (more specifically delay or
phase) and amplitude information from the sound perceived by our two
ears \cite{Hartmann1999}, sometimes in addition to information from
other senses. However, localising sound sources using only two inputs
is a challenging task. The human auditory system is very complex and
resolves the problem by accounting for the acoustic diffraction around
the head and the ridges of the outer ear. Without this ability, localisation
with two microphones is limited to azimuth only, along with the impossibility
to distinguish if the sounds come from the front or the back. Also,
obtaining high-precision readings when the sound source is in the
same axis as the pair of microphones is more difficult.

One advantage with robots is that they do not have to inherit the
same limitations as living creatures. Using more than two microphones
allows reliable and accurate localisation in three dimensions (azimuth
and elevation). Also, having multiple signals provides additional
redundancy, reducing the uncertainty caused by the noise and non-ideal
conditions such as reverberation and imperfect microphones. It is
with this principle in mind that we have developed an approach allowing
to localise sound sources using an array of microphones. 

Our approach is based on a frequency-domain beamformer that is steered
in all possible directions to detect sources. Instead of measuring
TDOAs and then converting to a position, the localisation process
is performed in a single step. This makes the system more robust,
especially in the case where an obstacle prevents one or more microphones
from properly receiving the signals. The results of the localisation
process are then enhanced by probability-based post-processing, which
prevents false detection of sources. This makes the system sensitive
enough for simultaneous localisation of multiple moving sound sources.
This approach is an extension of earlier work \cite{ValinICRA2004}
and works for both far-field and near-field sound sources. Detection
reliability, accuracy, and tracking capabilities of the approach are
validated using the Spartacus robot platform, with different types
of sound sources.

This chapter is organised as follows. Section \ref{sec:Related-work}
situates our work in relation to other research projects in the field.
Section \ref{sec:System-overview} presents a brief overview of the
system. Section \ref{sec:Localization-by-steered} describes our steered
beamformer implemented in the frequency-domain. Section \ref{sec:Probabilistic-post-processing}
explains how we enhance the results from the beamformer using a probabilistic
post-processor. This is followed by experimental results in Section
\ref{sec:Results}, showing how the system behaves under different
conditions. Section \ref{sec:Conclusion} concludes the chapter and
presents future work on sound localisation.

\section{Related Work\label{sec:Related-work}}

Most of the work in relation to localisation of sound sources has
been done using only two microphones. This is the case with the SIG
robot that uses both inter-aural phase difference (IPD) and inter-aural
intensity difference (IID) to locate sounds \cite{nakadai-matsuura-okuno-kitano2003}.
The binaural approach has limitations when it comes to evaluating
elevation and usually, the front-back ambiguity cannot be resolved
without resorting to active audition \cite{Nakadai2000}.

More recently, approaches using more than two microphones have been
developed. One approach uses a circular array of eight microphones
to locate sound sources using the MUSIC algorithm \cite{Asano2001},
a signal subspace approach. In our previous work also using eight
microphones \cite{ValinIROS2003}, we presented a method for localising
a single sound source where time delay of arrival (TDOA) estimation
was separated from the direction of arrival (DOA) estimation. It was
found that a system combining TDOA and DOA estimation in a single
step improves the system's robustness, while allowing localisation
(but not tracking) of simultaneous sources \cite{ValinICRA2004}.
Kagami \emph{et al.} \cite{Kagami2004} reports a system using 128
microphones for 2D sound localisation of sound sources. Similarly,
Wang \emph{et al.} \cite{Wang2004} use 24 fixed microphones to track
a moving robot in a room. However, it would not be practical to include
such a large number of microphones on a mobile robot.

Most of the work so far on localisation of source sources does not
address the problem of tracking moving sources. It is proposed in
\cite{Bechler2004} to use a Kalman filter for tracking a moving source.
However the proposed method assumes that a single source is present.
In the past years, particle filtering \cite{Arulampalam2002} (a sequential
Monte Carlo method) has been increasingly popular to resolve object
tracking problems. Ward \emph{et al.} \cite{Ward2002b,Ward2003} and
Vermaak \cite{Vermaak2001} use this technique for tracking single
sound sources. Asoh \emph{et al.} \cite{Asoh2004} even suggested
to use this technique for mixing audio and video data to track speakers.
But again, the technique is limited to a single source due to the
problem of associating the localisation observation data to each of
the sources being tracked. We refer to that problem as the source-observation
assignment problem. Some attempts are made at defining multi-modal
particle filters in \cite{Vermaak2003}, and the use of particle filtering
for tracking multiple targets is demonstrated in \cite{MacCormick2000,Hue2001,Vermaak2005}.
But so far, the technique has not been applied to sound source tracking.
Our work demonstrates that it is possible to track multiple sound
sources using particle filters by solving the source-observation assignment
problem.

\section{System Overview\label{sec:System-overview}}

\begin{figure}[th]
\center{\includegraphics[width=0.7\columnwidth]{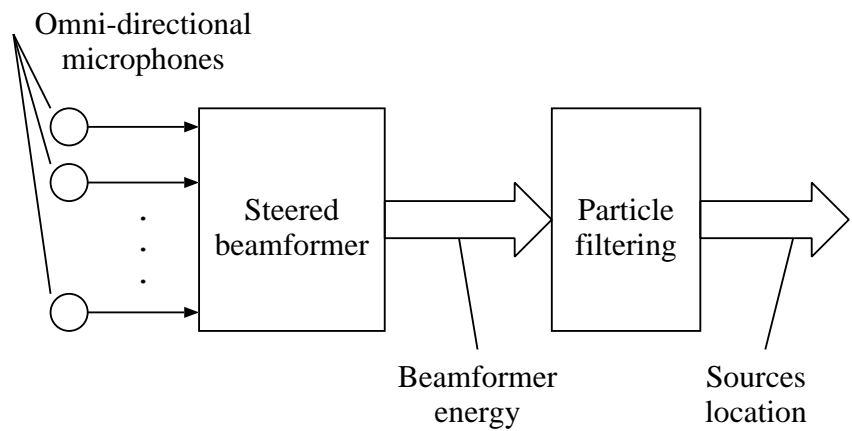}}

\caption{Overview of the Localisation subsystem.\label{cap:Overview-of-the-system}}
\end{figure}

The proposed localisation and tracking system, as shown in Figure
\ref{cap:Overview-of-the-system}, is composed of three parts:
\begin{itemize}
\item A microphone array;
\item A memoryless localisation algorithm based on a steered beamformer;
\item A particle filtering tracker.
\end{itemize}
The array is composed of up to eight omni-directional microphones
mounted on the robot. Since the system is designed to be installed
on any robot, there is no strict constraint on the position of the
microphones: only their positions must be known in relation to each
other (measured with $\sim$0.5 cm accuracy). The microphone signals
are used by a beamformer (spatial filter) that is steered in all possible
directions in order to maximise the output energy. The initial localisation
performed by the steered beamformer is then used as the input of a
post-processing stage that uses particle filtering to simultaneously
track all sources and prevent false detections. The output of the
localisation system can be used to direct the robot attention to the
source. It can also be used as part of a source separation algorithm
to isolate the sound coming from a single source, as described in
Chapter \ref{cha:Sound-Source-Separation}.

\section{Localisation Using a Steered Beamformer\label{sec:Localization-by-steered}}

\begin{comment}
The sound source localisation algorithm we use is based on the fact
that sound waves emitted from a source arrive at the microphones with
different delays
\end{comment}

The basic idea behind the steered beamformer approach to source localisation
is to direct a beamformer in all possible directions and look for
maximal output. This can be done by maximising the output energy of
a simple delay-and-sum beamformer. The formulation in both time and
frequency domain is presented in Section \ref{sub:Delay-And-Sum-Beamformer}.
Section \ref{sub:Spectral-weighting} describes the frequency-domain
weighting performed on the microphone signals and Section \ref{sub:Direction-Search}
shows how the search is performed. A possible modification for improving
the resolution is described in Section \ref{sub:Direction-Refining}.

\subsection{Delay-And-Sum Beamformer\label{sub:Delay-And-Sum-Beamformer}}

The output $y(n_{t})$of an $N$-microphone delay-and-sum beamformer
is defined as:

\begin{equation}
y(n_{t})=\sum_{n=0}^{N-1}x_{n}\left(n_{t}-\tau_{n}\right)\label{eq:delay-and-sum}
\end{equation}
where $x_{n}\left(n_{t}\right)$ is the signal from the $n^{th}$
microphone and $\tau_{n}$ is the delay of arrival for that microphone.
The output energy of the beamformer over a frame of length $L$ is
thus given by:
\begin{eqnarray}
E & = & \sum_{n_{t}=0}^{L-1}\left[y(n_{t})\right]^{2}\nonumber \\
 & = & \sum_{n_{t}=0}^{L-1}\left[x_{0}\left(n_{t}-\tau_{0}\right)+\ldots+x_{N-1}\left(n_{t}-\tau_{N-1}\right)\right]^{2}\label{eq:beamformer-energy}
\end{eqnarray}

Assuming that only one sound source is present, we can see that $E$
will be maximal when the delays $\tau_{n}$ are such that the microphone
signals are in phase, and therefore add constructively.

One problem with this technique is that energy peaks are very wide
\cite{Duraiswami2001}, which means that the accuracy on $\tau_{n}$
is poor. Moreover, in the case where multiple sources are present,
it is likely for the two or more energy peaks to overlap, making them
impossible to distinguish. One way to narrow the peaks is to whiten
the microphone signals prior to computing the energy \cite{Omologo}.
Unfortunately, the coarse-fine search method as proposed in \cite{Duraiswami2001}
cannot be used in that case because the narrow peaks can then be missed
during the coarse search. Therefore, a full fine search is necessary,
which requires increased computing power. 

It is however possible to reduce the amount of computation by calculating
the beamformer energy in the frequency domain. This also has the advantage
of making the whitening of the signal easier. To do so, the beamformer
output energy in Equation \ref{eq:beamformer-energy} can be expanded
as:

\begin{eqnarray}
E & = & \sum_{n=0}^{N-1}\sum_{n_{t}=0}^{L-1}x_{n}^{2}\left(n_{t}-\tau_{n}\right)\nonumber \\
 & + & 2\sum_{n_{1}=0}^{N-1}\sum_{n_{2}=0}^{n_{1}-1}\sum_{n_{t}=0}^{L-1}x_{n_{1}}\left(n_{t}-\tau_{n_{1}}\right)x_{n_{2}}\left(n_{t}-\tau_{n_{2}}\right)\label{eq:beamformer-energy-expand}
\end{eqnarray}
which in turn can be rewritten in terms of cross-correlations:

\begin{equation}
E=K+2\sum_{n_{1}=0}^{N-1}\sum_{n_{2}=0}^{n_{1}-1}R_{x_{n_{1}},x_{n_{2}}}\left(\tau_{n_{1}}-\tau_{n_{2}}\right)\label{eq:energy-xcorr}
\end{equation}
where $K=\sum_{n=0}^{N-1}\sum_{n_{t}=0}^{L-1}x_{n}^{2}\left(n_{t}-\tau_{n}\right)$
is nearly constant with respect to the $\tau_{m}$ delays and can
thus be ignored when maximising $E$. The cross-correlation function
can be approximated in the frequency domain as:

\begin{equation}
R_{ij}(\tau)\approx\sum_{k=0}^{L-1}X_{i}(k)X_{j}(k)^{*}e^{\jmath2\pi k\tau/L}\label{eq:TDOA_correlation_freq}
\end{equation}
where $X_{i}(k)$ is the discrete Fourier transform of $x_{i}(n_{t})$,
$X_{i}(k)X_{j}(k)^{*}$ is the cross-spectrum of $x_{i}(n_{t})$ and
$x_{j}(n_{t})$ and $(\cdot)^{*}$ denotes the complex conjugate.
The power spectra and cross-power spectra are computed on overlapping
windows (50\% overlap) of $L=1024$ samples at 48 kHz. The cross-correlations
$R_{ij}(\tau)$ are computed by averaging the cross-power spectra
$X_{i}(k)X_{j}(k)^{*}$ over a time period of 4 frames (40 ms). Once
the $R_{ij}(\tau)$ are pre-computed, it is possible to compute $E$
using only $N(N-1)/2$ lookup and accumulation operations, whereas
a time-domain computation would require $2L(N+2)$ operations. For
example, for $N=8$ microphones and $N_{g}=2562$ directions, it follows
that the complexity of the search itself is reduced from 1.2 Gflops
to only 1.7 Mflops. After counting all time-frequency transformations,
the complexity is only 48.4 Mflops, 25 times less than a time domain
search with the same resolution.

\subsection{Spectral Weighting\label{sub:Spectral-weighting}}

In the frequency domain, the whitened cross-correlation (also known
as the phase transform, or PHAT) is computed as:
\begin{equation}
R_{ij}^{(w)}(\tau)\approx\sum_{k=0}^{L-1}\frac{X_{i}(k)X_{j}(k)^{*}}{\left|X_{i}(k)\right|\left|X_{j}(k)\right|}e^{\jmath2\pi k\tau/L}\label{eq:TDOA_correlation_whitened}
\end{equation}

While it produces much sharper cross-correlation peaks, the whitened
cross-correla\-tion has one drawback: each frequency bin of the spectrum
contributes the same amount to the final correlation, even if the
signal at that frequency is dominated by noise. This makes the system
less robust to noise, while making detection of voice (which has a
narrow bandwidth) more difficult. In order to alleviate the problem,
we introduce a weighting function that acts as a mask based on the
signal-to-noise ratio (SNR). For microphone $i$, we define this weighting
function as:
\begin{equation}
\zeta_{i}^{\ell}(k)=\frac{\xi_{i}^{\ell}(k)}{\xi_{i}^{\ell}(k)+1}\label{eq:SNR-weighting}
\end{equation}
where $\xi_{i}^{\ell}(k)$ is an estimate of the \emph{a priori} SNR
at the $i^{th}$ microphone, at time frame $\ell$, for frequency
$k$. It is computed using the decision-directed approach proposed
by Ephraim and Malah \cite{EphraimMalah1984}:
\begin{equation}
\xi_{i}^{\ell}(k)=\frac{(1-\alpha_{d})\left[\zeta_{i}^{\ell-1}(k)\right]^{2}\left|X_{i}^{\ell-1}(k)\right|^{2}+\alpha_{d}\left|X_{i}^{\ell}(k)\right|^{2}}{\sigma_{i}^{2}(k)}\label{eq:decision-directed}
\end{equation}
where $\alpha_{d}=0.1$ is the adaptation rate and $\sigma_{i}^{2}(k)$
is the noise estimate for microphone $i$. It is easy to estimate
$\sigma_{i}^{2}(k)$ using the Minima-Controlled Recursive Average
(MCRA) technique \cite{CohenNonStat2001}, which adapts the noise
estimate during periods of low energy. 

It is also possible to make the system more robust to reverberation
by modifying the weighting function to include a reverberation term
$\lambda_{i}^{rev}(k,\ell)$ to the noise estimate at time frame $\ell$.
We use a simple reverberation model with exponential decay: 
\begin{equation}
\lambda_{i}^{rev}(k,\ell)=\gamma\lambda_{i}^{rev}(k,\ell)+\frac{(1-\gamma)}{\delta}\left|\zeta_{i}^{\ell}(k)X_{i}^{\ell-1}(k)\right|^{2}\label{eq:Reverb_weighting}
\end{equation}
where $\gamma$ represents the reverberation time ($T_{60}$) of the
room ($\gamma=10^{-6/T_{60}}$), $\delta$ is the Signal-to-Reverberant
Ratio (SRR) and $\lambda_{i}^{rev}(k,-1)=0$. In some sense, Equation
\ref{eq:Reverb_weighting} can be seen as modelling the \emph{precedence
effect} \cite{Huang1997Echo,Huang1999Echo} in order to give less
weight to frequency bins where a loud sound was recently present.
The resulting enhanced cross-correlation is defined as:
\begin{equation}
R_{ij}^{(e)}(\tau)=\sum_{k=0}^{L-1}\frac{\zeta_{i}(k)X_{i}(k)\zeta_{j}(k)X_{j}(k)^{*}}{\left|X_{i}(k)\right|\left|X_{j}(k)\right|}e^{\jmath2\pi k\tau/L}\label{eq:TDOA_correlation_weighted}
\end{equation}

The spectral weighting described above has similarities with the maximum
likelihood (ML) weighting described in \cite{Mungamuru2004}, with
two main differences. The first is that the weight we use requires
a lower complexity because it can be applied directly to the spectrum
of the signals. The second difference is that in high SNR conditions,
the cross-correlation peak can take very large values, and is thus
be more difficult to use for evaluating the probability that a source
is really present (see Section \ref{sub:Instantaneous-Direction-Probabilities}). 

The advantage of our approach over the simpler PHAT lies in the fact
that the PHAT does not take into account noise at all and assumes
that the signal-to-reverberant ratio (SRR) is constant across frequency
\cite{Mungamuru2004}. The latter assumption only holds when the signal
being tracked is relatively stationary (i.e., no transients) compared
to the reverberation time. For this reason, the PHAT cannot model
the precedence effect. In practice, it was found that when the presence
of a sound source is known, the localisation accuracy using our weighting
similar to that obtained using the PHAT. The main difference is that
our weighting makes it easier to estimate if a source is really present.
As far as we are aware, no other work focuses on sound source localisation
when the number of sources present is unknown (and detection becomes
important).

\subsection{Direction Search on a Spherical Grid\label{sub:Direction-Search}}

\begin{figure}[th]
\center{\subfloat[Icosahedral grid]{

\includegraphics[width=0.29\columnwidth,height=0.29\columnwidth]{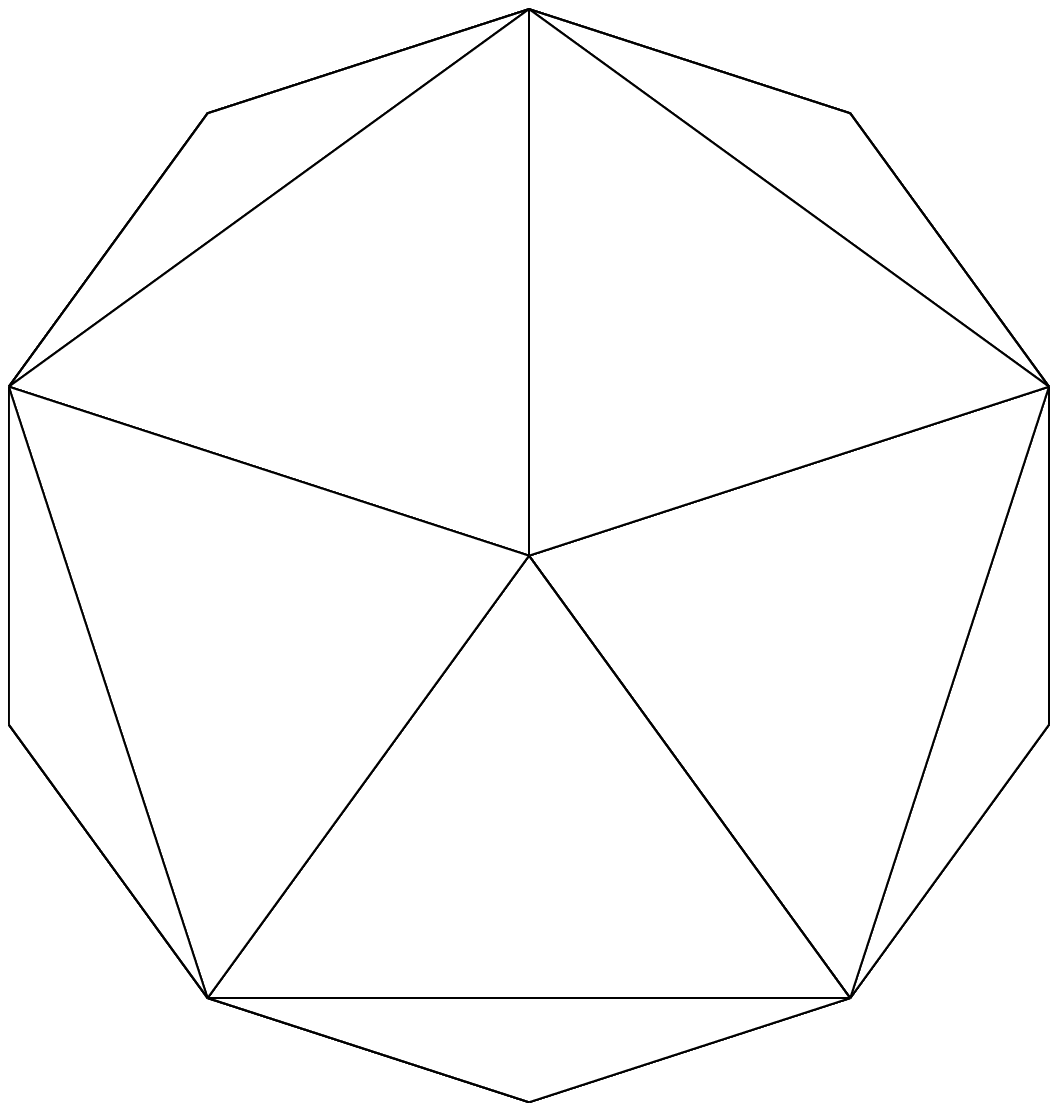}}\hspace{3mm}\subfloat[Subdivison by four (one level)]{

\includegraphics[width=0.29\columnwidth,height=0.29\columnwidth]{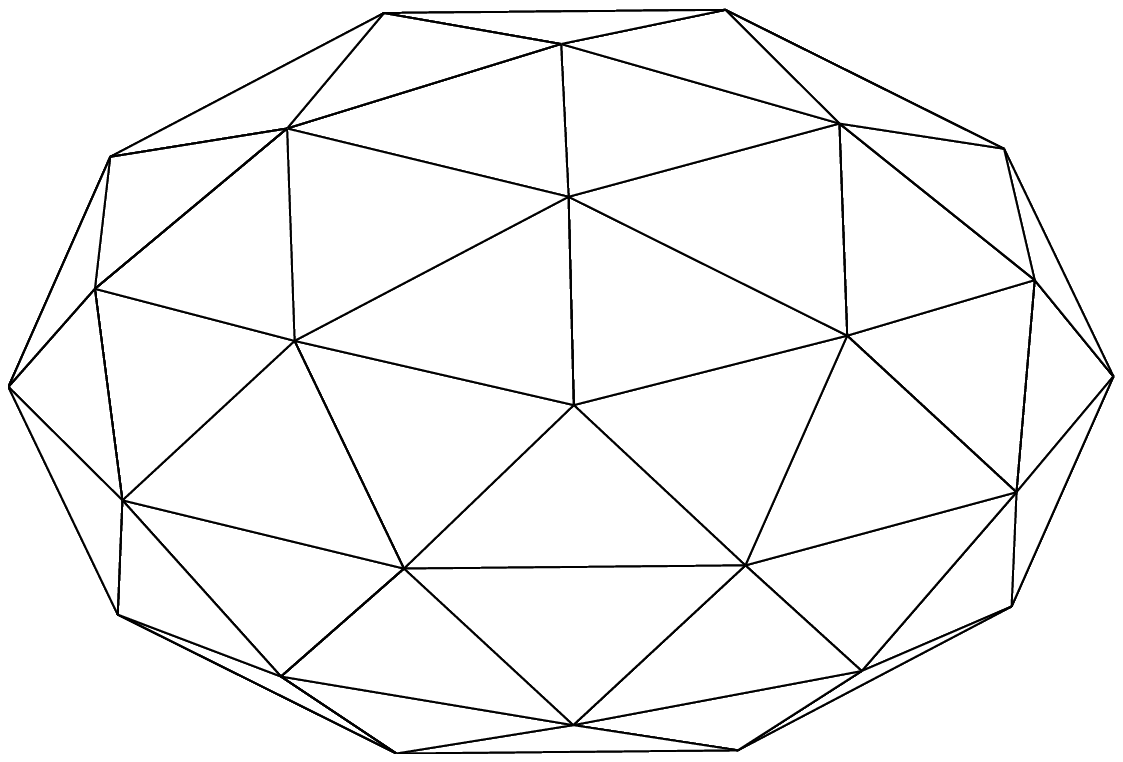}}\subfloat[Subdivison by sixteen (two levels)]{

\includegraphics[width=0.32\columnwidth,height=0.32\columnwidth]{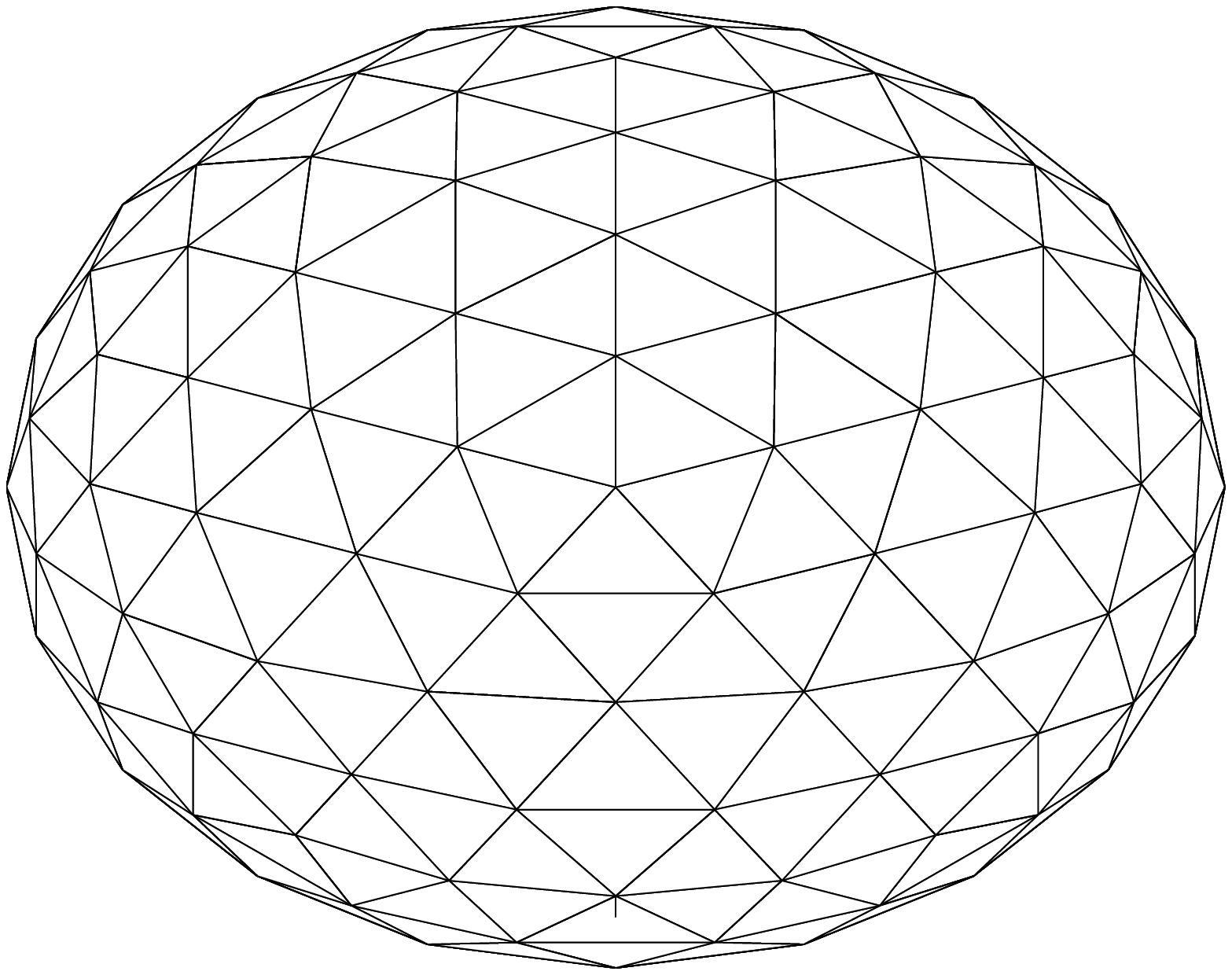}}}

\caption{Recursive subdivision of a triangular element.\label{cap:Recursive-subdivision}}
\end{figure}

In order to reduce the computation required and to make the system
isotropic, we define a uniform triangular grid for the surface of
a sphere. To create the grid, we start with an initial icosahedral
grid \cite{Giraldo97}. Each triangle in the initial 20-element grid
is recursively subdivided into four smaller triangles, as shown in
Figure \ref{cap:Recursive-subdivision}. The resulting grid is composed
of 5120 triangles and 2562 points. The beamformer energy is then computed
for the hexagonal region associated with each of these points. Each
of the 2562 regions covers a radius of about $2.5^{\circ}$ around
its centre, setting the resolution of the search.

\begin{algorithm}[th]
\begin{algorithmic}

\FORALL{grid index $k$}

\STATE $E_k \leftarrow 0$

\FORALL{microphone pair $ij$}

\STATE $\tau \leftarrow lookup(k,ij)$

\STATE $E_k \leftarrow E_k + R^{(e)}_{ij}(\tau)$

\ENDFOR

\ENDFOR

\STATE \textit{direction of source} $\leftarrow \textrm{argmax}_k\ (E_k)$

\end{algorithmic}

\caption{Steered beamformer direction search.\label{cap:Steered-beamformer-direction}}
\end{algorithm}

\begin{figure}[th]
\begin{center}\includegraphics[width=0.45\columnwidth,keepaspectratio]{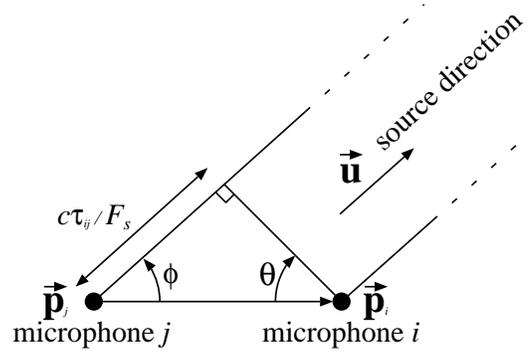}\end{center}

\caption{TDOA for the far field approximation.\label{cap:TDOA-far-field}}
\end{figure}

Once the cross-correlations $R_{ij}^{(e)}(\tau)$ are computed, the
search for the best direction on the grid is performed as described
by Algorithm \ref{cap:Steered-beamformer-direction}. The \emph{lookup}
parameter is a pre-computed table of the time delay of arrival (TDOA)
for each microphone pair and each direction on the sphere. Using the
far-field assumption as illustrated in Figure \ref{cap:TDOA-far-field},
the TDOA in samples is computed using the cosine law \cite{ValinIROS2003}:
\begin{equation}
\cos\phi=\frac{c\tau_{ij}/F_{s}}{\left\Vert \vec{\mathrm{\mathbf{p}}}_{i}-\mathrm{\vec{\mathrm{\mathbf{p}}}}_{j}\right\Vert }=\frac{\left(\vec{\mathrm{\mathbf{p}}}_{i}-\mathrm{\vec{\mathrm{\mathbf{p}}}}_{j}\right)\cdot\vec{\mathbf{u}}}{\left\Vert \vec{\mathrm{\mathbf{p}}}_{i}-\mathrm{\vec{\mathrm{\mathbf{p}}}}_{j}\right\Vert \left\Vert \vec{\mathbf{u}}\right\Vert }\label{eq:TDOA_cos_phi}
\end{equation}
where $\vec{\mathrm{\mathbf{p}}}_{i}$ is the position of microphone
$i$, $\vec{\mathbf{u}}$ is a unit-vector that points in the direction
of the source, $c$ is the speed of sound and $F_{s}$ is the sampling
rate. Isolating $\tau_{ij}$ from Equation \ref{eq:TDOA_cos_phi}
and knowing that $\vec{\mathbf{u}}$ is a unit vector, we obtain:
\begin{equation}
\tau_{ij}=\frac{F_{s}}{c}\left(\vec{\mathrm{\mathbf{p}}}_{i}-\mathrm{\vec{\mathrm{\mathbf{p}}}}_{j}\right)\cdot\vec{\mathbf{u}}\label{eq:TDOA-far-field}
\end{equation}

Equation \ref{eq:TDOA-far-field} assumes that the time delay is proportional
to the distance between the source and microphone. This is only true
when there is no diffraction involved. While this hypothesis is only
verified for an ``open'' array (all microphones are in line of sight
with the source), in practice we demonstrate experimentally (see Section
\ref{sec:Results}) that the approximation is good enough for our
system to work for a ``closed'' array (in which there are obstacles
within the array).

For an array of $N$ microphones and an $N_{g}$-element grid, the
algorithm requires $N(N-1)N_{g}$ table memory accesses and $N(N-1)N_{g}/2$
additions. In the proposed configuration ($N_{g}=2562$, $N=8$),
the accessed data can be made to fit entirely in a modern processor's
L2 cache.

\begin{algorithm}[th]
\begin{algorithmic}

\FOR{$q=0$ to $Q-1$}

\STATE $D_q \leftarrow \textrm{Steered beamformer direction search}$

\FORALL{microphone pair $ij$}

\STATE $\tau \leftarrow lookup(D_q,ij)$

\STATE $R^{(e)}_{ij}(\tau) = 0$

\ENDFOR

\ENDFOR

\end{algorithmic}

\caption{Localisation of multiple sources.\label{cap:Localization-of-multiple-sources}}
\end{algorithm}

Using Algorithm \ref{cap:Steered-beamformer-direction}, our system
is able to find the loudest source present by maximising the energy
of a steered beamformer. In order to localise other sources that may
be present, the process is repeated by removing the contribution of
the first source to the cross-correlations, leading to Algorithm \ref{cap:Localization-of-multiple-sources}.
Since we do not know how many sources are present, the number of sources
to find is set to the constant $Q$. Therefore, Algorithm \ref{cap:Localization-of-multiple-sources}
finds the $Q$ loudest sources around the array. We determined empirically
that the maximum number of sources our beamformer is able to locate
at once is four. The fact that Algorithm \ref{cap:Localization-of-multiple-sources}
always finds four sources regardless of the number of sources present
leads to a high rate of false detection, even when four or more sources
are present. That problem is handled by the particle filter described
in Section \ref{sec:Probabilistic-post-processing}.

\subsection{Direction Refining\label{sub:Direction-Refining}}

When a source is located using Algorithm \ref{cap:Steered-beamformer-direction},
the direction accuracy is limited by the size of the grid used. It
is however possible, as an optional step, to further refine the source
location estimate. In order to do so, we define a refined grid for
the surrounding of the point where a source was found. To take into
account the near-field effects, the grid is refined in three dimensions:
horizontally, vertically and over distance. Using five points in each
direction, we obtain a 125-point local grid with a maximum error of
around $1^{\circ}$. For the near-field case, Equation \ref{eq:TDOA-far-field}
no longer holds, so it is necessary to compute the time differences
as:
\begin{equation}
\tau_{ij}=\frac{F_{s}}{c}\left(\left\Vert d\vec{\mathbf{u}}-\vec{\mathbf{p}}_{j}\right\Vert -\left\Vert d\vec{\mathbf{u}}-\vec{\mathbf{p}}_{i}\right\Vert \right)\label{eq:TDOA-near-field}
\end{equation}
where $d$ is the distance between the source and the centre of the
array. Equation \ref{eq:TDOA-near-field} is evaluated for five distances
$d$ (ranging from 50 cm to 5 m) in order to find the direction of
the source with improved accuracy. Unfortunately, it was observed
that the value of $d$ found in the search is too unreliable to provide
a good estimate of the distance between the source and the array.
The incorporation of the distance nonetheless allows improved accuracy
for the near field case.

\section{Particle-Based Tracking\label{sec:Probabilistic-post-processing}}

The steered beamformer detailed in Section \ref{sec:Localization-by-steered}
provides only instantaneous, noisy information about sources being
possibly present, and no information about the behaviour of the source
in time (i.e. tracking). For that reason, it is desirable to use a
probabilistic temporal integration to track the different sound sources
based on all measurements available up to the current time. 

\begin{figure}[th]
\begin{center}\includegraphics[width=0.6\columnwidth,keepaspectratio]{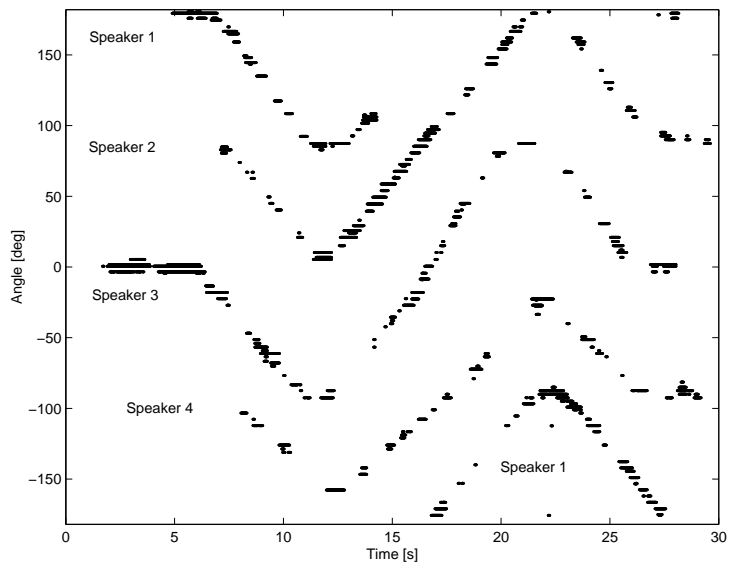}\end{center}

\caption{Probabilistic tracking of multiple sources using a fixed grid.\label{cap:Probabilistic-tracking}}
\end{figure}

It has been shown \cite{Ward2002b,Ward2003,Asoh2004} that particle
filters are an effective way of tracking sound sources. The choice
of particle filtering is further motivated by the fact that earlier
work using a fixed grid for tracking showed that the technique can
not provide continuous tracking when moving sources had short periods
of silence. This can be observed in Figure \ref{cap:Probabilistic-tracking},
taken from \cite{ValinICRA2004}. Particle filtering is also preferred
to Kalman filtering because key aspects of the proposed algorithm,
such as the handling of false detections and source-observation assignment
(see Section \ref{sub:Probabilities-for-Multiple}), cannot be adequately
modelled as a Gaussian process, as is assumed by the Kalman filter
\cite{Bechler2004}.

Let $\mathbf{S}_{j}^{(t)}$ be the state variable associated to source
$j$ ($j=0,1,\ldots,M-1$) at time $t$, a particle filter approximates
the probability density function (pdf) of $\mathbf{S}_{j}^{(t)}$
as:
\[
p\left(\mathbf{S}_{j}^{(t)}\right)\approx\sum_{i=1}^{N_{p}}w_{j,i}^{(t)}\delta\left(\mathbf{S}_{j}^{(t)}-\mathbf{s}_{j,i}^{(t)}\right)
\]
where $\delta\left(\mathbf{S}_{j}^{(t)}-\mathbf{s}_{j,i}^{(t)}\right)$
is the Dirac function for a particle of state $\mathbf{s}_{j,i}^{(t)}$,
$w_{j,i}^{(t)}$ is the particle weight and $N_{p}$ is the number
of particles. With this approach, each particle can be viewed as representing
a hypothesis about the location of a sound source and the weights
assigned to the particles represent the probability for each hypothesis
to be correct. The state vector for the particles is composed of six
dimensions, three for the position $\mathbf{x}_{j,i}^{(t)}$ and three
for its derivative: 
\begin{equation}
\mathbf{s}_{j,i}^{(t)}=\left[\begin{array}{c}
\mathbf{x}_{j,i}^{(t)}\\
\mathbf{\dot{x}}_{j,i}^{(t)}
\end{array}\right]\label{eq:particle_state}
\end{equation}

\begin{figure}[th]
\begin{center}\includegraphics[width=0.7\columnwidth,keepaspectratio]{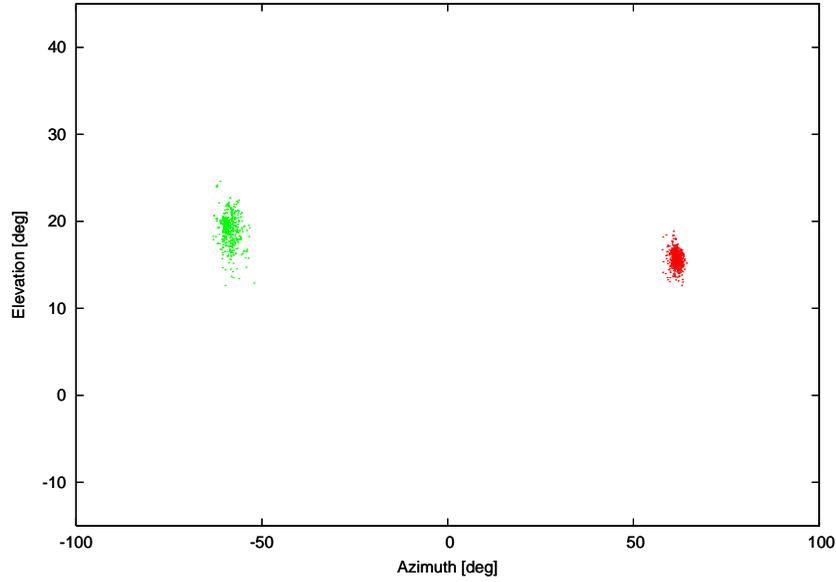}\end{center}

\caption{Example of two sources being represented by a particle filter.\label{cap:Particle-example}}
\end{figure}

Figure \ref{cap:Particle-example} illustrates the particle representation
of two sources. The source in red is located around 60 degrees azimuth
and 15 degrees elevation while the source in green is located around
-60 degrees azimuth and 20 degrees elevation. The spread of the particles
is an indicator of the uncertainty on the source position.

\begin{algorithm}[!t]
\begin{enumerate}
\item Predict the state $\mathbf{s}_{j}^{(t)}$ from $\mathbf{s}_{j}^{(t-1)}$
for each source $j$.
\item Compute instantaneous direction probabilities associated with the
steered beamformer response.
\item Compute probabilities $P_{q,j}^{(t)}$ associating beamformer peaks
to sources being tracked.
\item Compute updated particle weights $w_{j,i}^{(t)}$.
\item Add or remove sources if necessary.
\item Compute source localisation estimate $\bar{\mathbf{x}}_{j}^{(t)}$
for each source.
\item Resample particles for each source if necessary and go back to step
1.
\end{enumerate}
\begin{comment}
Particle-based tracking algorithm. Steps 1 to 7 correspond to Subsections
\ref{sub:Prediction} to \ref{sub:Resampling}.
\end{comment}

\caption[Particle-based tracking algorithm]{Particle-based tracking algorithm. Steps 1 to 7 correspond to Subsections \ref{sub:Prediction} to \ref{sub:Resampling}.}\label{cap:Particle-based-tracking-algorithm}
\end{algorithm}

Since the particle position is constrained to lie on a unit sphere
and the speed is tangent to the sphere, there are only four degrees
of freedom. The particle filtering algorithm is outlined in Algorithm
\ref{cap:Particle-based-tracking-algorithm} and generalises sound
source tracking to an arbitrary and non-constant number of sources.
The steps are detailed in Subsections \ref{sub:Prediction} to \ref{sub:Resampling}.
The particle weights are updated by taking into account observations
obtained from the steered beamformer and by computing the assignment
between these observations and the sources being tracked. From there,
the estimated location of the source is the weighted mean of the particle
positions.

\subsection{Prediction\label{sub:Prediction}}

The prediction step in particle filtering plays a similar role as
for the Kalman filter. However, instead of explicitly predicting the
mean and variance of the model, the particles are subjected to a stochastic
excitation with damping model as proposed in \cite{Ward2003}:

\begin{eqnarray}
\mathbf{\dot{x}}_{j,i}^{(t)} & = & a\mathbf{\dot{x}}_{j,i}^{(t-1)}+bF_{\mathbf{x}}\label{eq:predict_speed}\\
\mathbf{x}_{j,i}^{(t)} & = & \mathbf{x}_{j,i}^{(t-1)}+\Delta T\mathbf{\dot{x}}_{j,i}^{(t)}\label{eq:predict_pos}
\end{eqnarray}

where $a=e^{-\alpha\Delta T}$ controls the damping term, $b=\beta\sqrt{1-a^{2}}$
controls the excitation term, $F_{\mathbf{x}}$ is a normally distributed
random variable of unit variance and $\Delta T$ is the time interval
between updates. In addition, we consider three possible states:
\begin{itemize}
\item Stationary source ($\alpha=2$, $\beta=0.04$);
\item Constant velocity source ($\alpha=0.05$, $\beta=0.2$);
\item Accelerated source ($\alpha=0.5$, $\beta=0.2$).
\end{itemize}
A normalisation step\footnote{The normalisation is performed as $\mathbf{x}_{i}^{(t)}\leftarrow\mathbf{x}_{i}^{(t)}/\left\Vert \mathbf{x}_{i}^{(t)}\right\Vert $.}
ensures that $\mathbf{x}_{i}^{(t)}$ still lies on the unit sphere
($\left\Vert \mathbf{x}_{j,i}^{(t)}\right\Vert =1$) after applying
Equations \ref{eq:predict_speed} and \ref{eq:predict_pos}.

\subsection{Instantaneous Direction Probabilities from Beamformer Response\label{sub:Instantaneous-Direction-Probabilities}}

\begin{figure}[th]
\center{\includegraphics[width=3.5in,keepaspectratio]{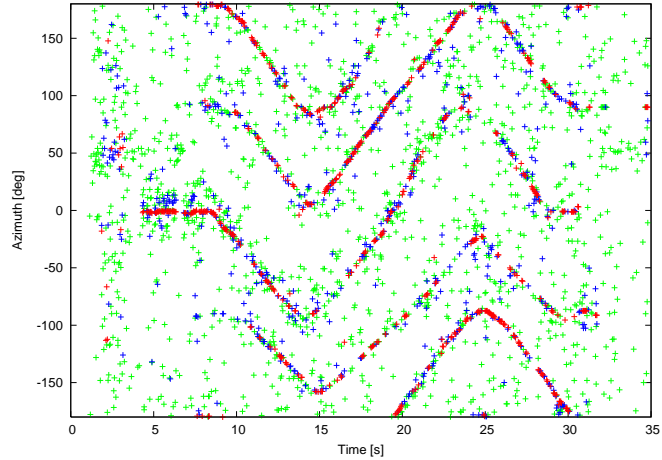}}

\caption[Beamformer output probabilities $P_q$ for azimuth as a function of time.]{Beamformer output probabilities $P_q$ for azimuth as a function of time. Observations with $P_q>0.5$ shown in red, $0.2<P_q<0.5$ in blue, $P_q<0.2$ in green.}\label{cap:Beamformer-output-probabilities}

\begin{comment}
Beamformer output probabilities $P_{q}$ for azimuth as a function
of time. Observations with $P_{q}>0.5$ shown in red, $0.2<P_{q}<0.5$
in blue, $P_{q}<0.2$ in green.\ref{eq:potential-source-reliability}
\end{comment}
\end{figure}

The steered beamformer described in Section \ref{sec:Localization-by-steered}
produces an observation $O^{(t)}$ for each time $t$. The observation
$O^{(t)}=\left[O_{0}^{(t)}\ldots O_{Q-1}^{(t)}\right]$ is composed
of $Q$ potential source locations $\mathbf{y}_{q}$ found by Algorithm
\ref{cap:Localization-of-multiple-sources}. We also denote $\mathbf{O}^{(t)}$,
the set of all observations $O^{(t)}$ up to time $t$. We introduce
the probability $P_{q}$ that the potential source $q$ is a true
source (not a false detection). The value of $P_{q}$ can be interpreted
as our confidence in the steered beamformer output. For the first
source ($q=0$), we have observed that the higher the beamformer energy,
the more likely that potential source is to be true. However, for
the other potential sources ($q>0$), false alarms are very frequent
and independent of energy. With this in mind, for the four potential
sources $q$, we define $P_{q}$ empirically as:
\begin{equation}
P_{q}=\left\{ \begin{array}{ll}
\nu^{2}/2, & q=0,\nu\leq1\\
1-\nu^{-2}/2,\qquad & q=0,\nu>1\\
0.3, & q=1\\
0.16, & q=2\\
0.03, & q=3
\end{array}\right.\label{eq:potential-source-reliability}
\end{equation}
with $\nu=E_{0}/E_{T}$, where $E_{T}$ is a threshold that depends
on the number of microphones, the frame size and the analysis window
used (we empirically found that $E_{T}=150$ is appropriate for eight
microphones). Figure \ref{cap:Beamformer-output-probabilities} shows
an example of $P_{q}$ values for potential sources found by the steered
beamformer in a case with four moving sources\footnote{Only the azimuth part of $\mathbf{y}_{q}$ is shown as a function
of time.}. It is possible to distinguish four trajectories, but it can be seen
that the observations from the steered beamformer are nonetheless
very noisy.

At time $t$, the probability density of observing $O_{q}^{(t)}$
for a source located at particle position $\mathbf{x}_{j,i}^{(t)}$
is given by:
\begin{equation}
p\left(\left.O_{q}^{(t)}\right|\mathbf{x}_{j,i}^{(t)}\right)=\mathcal{N}\left(\mathbf{y}_{q};\mathbf{x}_{j,i};\sigma^{2}\right)\label{eq:normal-pdf}
\end{equation}
where $\mathcal{N}\left(\mathbf{y}_{q};\mathbf{x}_{j,i};\sigma^{2}\right)$
is a normal distribution centred at $\mathbf{x}_{j,i}$ with variance
$\sigma^{2}$ evaluated at $\mathbf{y}_{q}$, and models the localisation
accuracy of the steered beamformer. We use $\sigma=0.05$, which corresponds
to an RMS error of 3 degrees for the location found by the steered
beamformer.

\subsection{Probabilities for Multiple Sources\label{sub:Probabilities-for-Multiple}}

Before we can derive the update rule for the particle weights $w_{j,i}^{(t)}$,
we must first introduce the concept of source-observation assignment.
For each potential source $q$ detected by the steered beamformer,
there are three possibilities:
\begin{itemize}
\item It is a false detection ($H_{0}$).
\item It corresponds to one of the sources currently tracked ($H_{1}$).
\item It corresponds to a new source that is not yet being tracked ($H_{2}$).
\end{itemize}
In the case of $H_{1}$, we need to determine which tracked source
$j$ corresponds to potential source $q$. First, we assume that a
potential source may correspond to at most one tracked source and
that a tracked source can correspond to at most one potential source. 

\begin{figure}[h]
\center{\includegraphics[width=7cm,keepaspectratio]{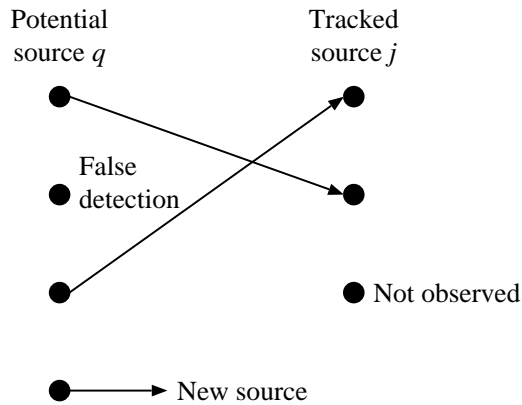}}

\caption[Assignment example where two of the tracked sources are observed, with one new source and one false detection.]{Assignment example where two of the tracked sources are observed, with one new source and one false detection. The assignment can be described as $f(\{0,1,2,3\})=\{1,-2,0,-1\}$.}\label{cap:Mapping-example}

\begin{comment}
Assignment example where two of the tracked sources are observed,
with one new source and one false detection. The assignment can be
described as $f(\{0,1,2,3\})=\{1,-2,0,-1\}$.\label{cap:Mapping-example}
\end{comment}
\end{figure}

Let $f:\{0,1,\ldots,Q-1\}\longrightarrow\{-2,-1,0,1,\ldots,M-1\}$
be a function assigning observation $q$ to the source $j$ (values
-2 is used for false detection and -1 is used for a new source). Figure
\ref{cap:Mapping-example} illustrates a hypothetical case with four
potential sources detected by the steered beamformer and their assignment
to the tracked sources. Knowing $P\left(f\left|O^{(t)}\right.\right)$
(the probability that $f$ is the correct assignment given observation
$O^{(t)}$) for all possible $f$, we can derive $P_{q,j}$, the probability
that the tracked source $j$ corresponds to the potential source $q$
as:
\begin{eqnarray}
P_{q,j}^{(t)} & = & \sum_{f}\delta_{j,f(q)}P\left(f\left|O^{(t)}\right.\right)\label{eq:Pqj}\\
P_{q}^{(t)}\left(H_{0}\right) & = & \sum_{f}\delta_{-2,f(q)}P\left(f\left|O^{(t)}\right.\right)\label{eq:PH0}\\
P_{q}^{(t)}\left(H_{2}\right) & = & \sum_{f}\delta_{-1,f(q)}P\left(f\left|O^{(t)}\right.\right)\label{eq:PH2}
\end{eqnarray}
where $\delta_{i,j}$ is the Kronecker delta. Equation \ref{eq:Pqj}
is in fact the sum of the probabilities of all $f$ that assign potential
source $q$ to tracked source $j$ and similarly for Equations \ref{eq:PH0}
and \ref{eq:PH2}.

Omitting $t$ for clarity, the probability $P(f|O)$ is given by:

\begin{equation}
P(f|O)=\frac{p(O|f)P(f)}{p(O)}\label{eq:mapping-prob}
\end{equation}
Knowing that there is only one correct assignment ($\sum_{f}P(f|O)=1$),
we can avoid computing the denominator $p(O)$ by using normalisation.
Assuming conditional independence of the observations given the mapping
function, we can decompose $p\left(\left.O\right|f\right)$ into individual
components:
\begin{equation}
p\left(\left.O\right|f\right)=\prod_{q}p\left(\left.O_{q}\right|f(q)\right)\label{eq:inverse-mapping-prob-prod}
\end{equation}
We assume that the distribution of the false detections ($H_{0}$)
and the new sources ($H_{2}$) are uniform, while the distribution
for tracked sources ($H_{1}$) is the pdf approximated by the particle
distribution convolved with the steered beamformer error pdf:
\begin{equation}
p\left(\left.O_{q}\right|f(q)\right)=\left\{ \begin{array}{ll}
1/4\pi,\qquad & f(q)=-2\\
1/4\pi, & f(q)=-1\\
\sum_{i}w_{f(q),i}p\left(\left.O_{q}\right|\mathbf{x}_{j,i}\right), & f(q)\geq0
\end{array}\right.\label{eq:indep-inverse-mapping}
\end{equation}
The \emph{a priori} probability of $f$ being the correct assignment
is also assumed to come from independent individual components, so
that:
\begin{equation}
P(f)=\prod_{q}P\left(f(q)\right)\label{eq:apriori-mapping}
\end{equation}
with:
\begin{equation}
P\left(f(q)\right)=\left\{ \begin{array}{ll}
\left(1-P_{q}\right)P_{false},\qquad & f(q)=-2\\
P_{q}P_{new} & f(q)=-1\\
P_{q}P\left(Obs_{j}^{(t)}\left|\mathbf{O}^{(t-1)}\right.\right) & f(q)\geq0
\end{array}\right.\label{eq:indep-apriori-mapping}
\end{equation}
where $P_{new}$ is the \emph{a priori} probability that a new source
appears and $P_{false}$ is the \emph{a priori} probability of false
detection. The probability $P\left(Obs_{j}^{(t)}\left|\mathbf{O}^{(t-1)}\right.\right)$
that source $j$ is observable (i.e., that it exists and is active)
at time $t$ is given by:

\begin{equation}
P\left(Obs_{j}^{(t)}\left|\mathbf{O}^{(t-1)}\right.\right)=P\left(E_{j}\left|\mathbf{O}^{(t-1)}\right.\right)P\left(\mathrm{A}_{j}^{(t)}\left|\mathbf{O}^{(t-1)}\right.\right)\label{eq:prob-observable}
\end{equation}
where $E_{j}$ is the event that source $j$ actually exists and $A_{j}^{(t)}$
is the event that it is active (but not necessarily detected) at time
$t$. By active, we mean that the signal it emits is non-zero (for
example, a speaker who is not making a pause). The probability that
the source exists is given by:
\begin{equation}
P\left(E_{j}\left|\mathbf{O}^{(t-1)}\right.\right)=P_{j}^{(t-1)}+\left(1-P_{j}^{(t-1)}\right)\frac{P_{o}P\left(E_{j}\left|\mathbf{O}^{(t-2)}\right.\right)}{1-\left(1-P_{o}\right)P\left(E_{j}\left|\mathbf{O}^{(t-2)}\right.\right)}\label{eq:prob-exist}
\end{equation}
where $P_{o}$ is the \emph{a priori} probability that a source is
not observed (i.e., undetected by the steered beamformer) even if
it exists (with $P_{0}=0.2$ in our case) and $P_{j}^{(t)}=\sum_{q}P_{q,j}^{(t)}$
is the probability that source $j$ is observed (assigned to any of
the potential sources).

Assuming a first order Markov process, we can write the following
about the probability of source activity:
\begin{eqnarray}
P\left(\mathrm{A}_{j}^{(t)}\left|\mathbf{O}^{(t-1)}\right.\right) & = & P\left(\mathrm{A}_{j}^{(t)}\left|\mathrm{A}_{j}^{(t-1)}\right.\right)P\left(\mathrm{A}_{j}^{(t-1)}\left|\mathbf{O}^{(t-1)}\right.\right)\nonumber \\
 &  & +P\left(\mathrm{A}_{j}^{(t)}\left|\neg\mathrm{A}_{j}^{(t-1)}\right.\right)\left[1-P\left(\mathrm{A}_{j}^{(t-1)}\left|\mathbf{O}^{(t-1)}\right.\right)\right]\label{eq:prob-active-t-1}
\end{eqnarray}
with $P\left(\mathrm{A}_{j}^{(t)}\left|\mathrm{A}_{j}^{(t-1)}\right.\right)$
the probability that an active source remains active (set to 0.95),
and $P\left(\mathrm{A}_{j}^{(t)}\left|\neg\mathrm{A}_{j}^{(t-1)}\right.\right)$
the probability that an inactive source becomes active again (set
to 0.05). Assuming that the active and inactive states are equiprobable,
the activity probability is computed using Bayes' rule and usual probability
manipulations:
\begin{equation}
P\left(\mathrm{A}_{j}^{(t)}\left|\mathbf{O}^{(t)}\right.\right)=\frac{1}{1+\frac{\left[1-P\left(\mathrm{A}_{j}^{(t)}\left|\mathbf{O}^{(t-1)}\right.\right)\right]\left[1-P\left(\mathrm{A}_{j}^{(t)}\left|O^{(t)}\right.\right)\right]}{P\left(\mathrm{A}_{j}^{(t)}\left|\mathbf{O}^{(t-1)}\right.\right)P\left(\mathrm{A}_{j}^{(t)}\left|O^{(t)}\right.\right)}}\label{eq:prob-active}
\end{equation}

\subsection{Weight Update}

At times $t$, the new particle weights for source $j$ are defined
as:

\begin{equation}
w_{j,i}^{(t)}=p\left(\mathbf{x}_{j,i}^{(t)}\left|\mathbf{O}^{(t)}\right.\right)\label{eq:weight_update_def}
\end{equation}

Assuming that the observations are conditionally independent given
the source position, and knowing that for a given source $j$, $\sum_{i=1}^{N_{p}}w_{j,i}^{(t)}=1$,
we obtain through Bayesian inference:

\begin{eqnarray}
w_{j,i}^{(t)} & = & \frac{p\left(\left.\mathbf{O}^{(t)}\right|\mathbf{x}_{j,i}^{(t)}\right)p\left(\mathbf{x}_{j,i}^{(t)}\right)}{p\left(\mathbf{O}^{(t)}\right)}\nonumber \\
 & = & \frac{p\left(\left.O^{(t)}\right|\mathbf{x}_{j,i}^{(t)}\right)p\left(\left.\mathbf{O}^{(t-1)}\right|\mathbf{x}_{j,i}^{(t)}\right)p\left(\mathbf{x}_{j,i}^{(t)}\right)}{p\left(\mathbf{O}^{(t)}\right)}\nonumber \\
 & = & \frac{p\left(\mathbf{x}_{j,i}\left|O^{(t)}\right.\right)p\left(\mathbf{x}_{j,i}^{(t)}\left|\mathbf{O}^{(t-1)}\right.\right)p\left(O^{(t)}\right)p\left(\mathbf{O}^{(t-1)}\right)}{p\left(\mathbf{O}^{(t)}\right)p\left(\mathbf{x}_{j,i}^{(t)}\right)}\nonumber \\
 & = & \frac{p\left(\mathbf{x}_{j,i}^{(t)}\left|O^{(t)}\right.\right)w_{j,i}^{(t-1)}}{\sum_{i=1}^{N_{p}}p\left(\mathbf{x}_{j,i}^{(t)}\left|O^{(t)}\right.\right)w_{j,i}^{(t-1)}}\label{eq:weight-update}
\end{eqnarray}

Let $I_{j}^{(t)}$ denote the event that source $j$ is observed at
time $t$ and knowing that $P\left(I_{j}^{(t)}\right)=P_{j}^{(t)}=\sum_{q}P_{q,j}^{(t)}$,
we have:

\begin{equation}
p\left(\mathbf{x}_{j,i}^{(t)}\left|O^{(t)}\right.\right)=\left(1-P_{j}^{(t)}\right)p\left(\mathbf{x}_{j,i}^{(t)}\left|O^{(t)},\neg I_{j}^{(t)}\right.\right)+P_{j}^{(t)}p\left(\mathbf{x}_{j,i}^{(t)}\left|O^{(t)},I_{j}^{(t)}\right.\right)\label{eq:instant-weight}
\end{equation}
In the case where no observation matches the source, all particles
have the same probability, so we obtain:

\begin{equation}
p\left(\mathbf{x}_{j,i}^{(t)}\left|O^{(t)}\right.\right)=\left(1-P_{j}^{(t)}\right)\frac{1}{N_{p}}+P_{j}\frac{\sum_{q=1}^{Q}P_{q,j}^{(t)}p\left(\left.O_{q}^{(t)}\right|\mathbf{x}_{j,i}^{(t)}\right)}{\sum_{i=1}^{N}\sum_{q=1}^{Q}P_{q,j}^{(t)}p\left(\left.O_{q}^{(t)}\right|\mathbf{x}_{j,i}^{(t)}\right)}\label{eq:instant-weight2}
\end{equation}
where the denominator on the right side of Equation \ref{eq:instant-weight2}
provides normalisation for the $I_{j}^{(t)}$ case, so that $\sum_{i=1}^{N}p\left(\mathbf{x}_{j,i}^{(t)}\left|O^{(t)},I_{j}^{(t)}\right.\right)=1$.

\subsection{Adding or Removing Sources}

In a real environment, sources may appear or disappear at any moment.
If, at any time, $P_{q}(H_{2})$ is higher than a threshold empirically
set\footnote{The value must be small enough for all sources to be detected, but
large enough to prevent a large number of false alarms from being
tracked.} to 0.3, we consider that a new source is present. In that case, a
set of particles is created for source $q$. Even when a new source
is created, it is only assumed to exist if its probability of existence
$P\left(E_{j}\left|\mathbf{O}^{(t)}\right.\right)$ reaches a certain
threshold, which we empirically set\footnote{The exact value does not have a significant impact on the performance
of the system} to 0.98. At this point, the probability of existence is set to 1
and ceases to be updated.

In the same way, we set a time limit (typically two seconds) on sources.
If the source has not been observed ($P_{j}^{(t)}<T_{obs}$) for a
certain amount of time, we consider that it no longer exists. In that
case, the corresponding particle filter is no longer updated nor considered
in future calculations. The value of $T_{obs}$ only determines whether
a discontinuous source will be considered as one or two sources.

\subsection{Parameter Estimation}

The estimated position $\hat{\mathbf{x}}_{j}^{(t)}$ of each source
is the mean of the pdf and can be obtained as a weighted average of
its particles position:

\begin{equation}
\hat{\mathbf{x}}_{j}^{(t)}=\sum_{i=1}^{N_{p}}w_{j,i}^{(t)}\mathbf{x}_{j,i}^{(t)}\label{eq:pos_estimation_mean}
\end{equation}

\begin{figure}[th]
\center{\subfloat[No delay]{

\includegraphics[width=2.5in]{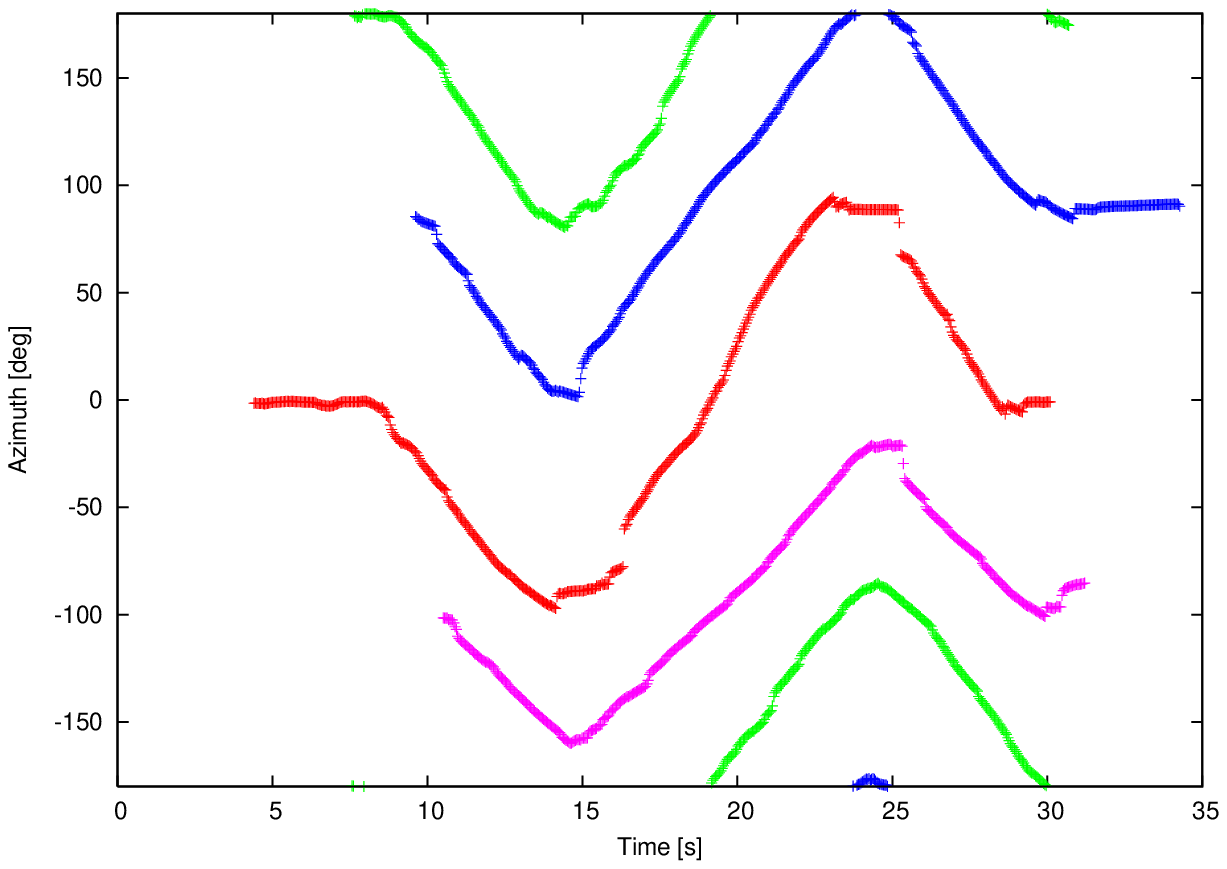}}\subfloat[Delayed estimation ($T$=500 ms)]{

\includegraphics[width=2.5in]{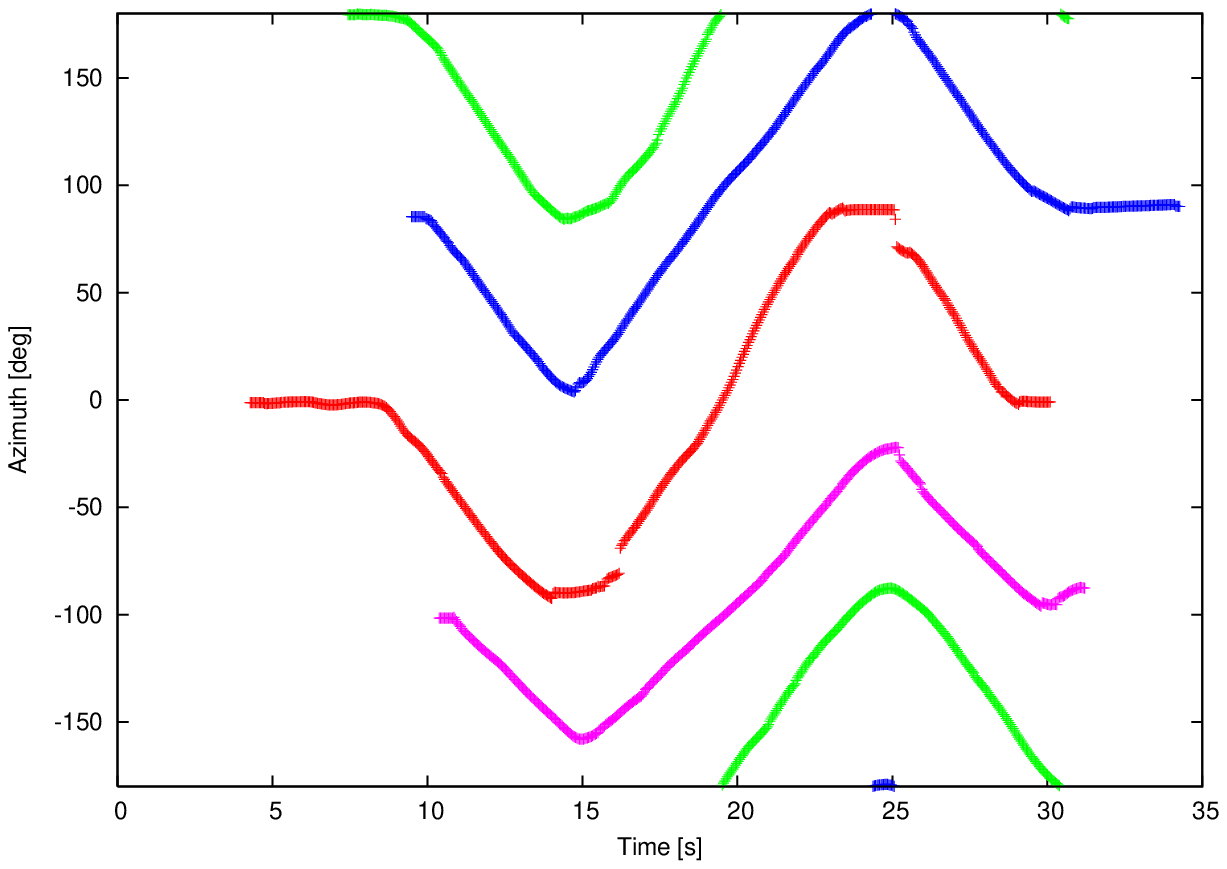}}}

\caption{Tracking of four moving sources, showing azimuth as a function of
time.\label{cap:Tracking-delay}}
\end{figure}

It is however possible to obtain better accuracy simply by adding
a delay to the algorithm. This can be achieved by augmenting the state
vector by past position values. At time $t$, the position at time
$t-T$ is thus expressed as:
\begin{equation}
\hat{\mathbf{x}}_{j}^{(t-T)}=\sum_{i=1}^{N_{p}}w_{j,i}^{(t)}\mathbf{x}_{j,i}^{(t-T)}\label{eq:pos_estimation_delayed}
\end{equation}
Using the same example as in Figure \ref{cap:Beamformer-output-probabilities},
Figure \ref{cap:Tracking-delay} represents how the particle filter
is able to remove the noise and produce smooth trajectories. The added
delay produces an even smoother result.

\subsection{Resampling\label{sub:Resampling}}

Resampling is necessary in order to prevent the filter from degenerating
to a single particle of weight 1. During the resampling stage $N_{p}$
``new'' particles are drawn from the original $N_{p}$ particles
with the probably of a particle being selected being proportional
to its weight $w_{j,i}^{(t)}$. After resampling, all particle weights
are reset to $1/N_{p}$, preserving the original pdf. The resampling
step in particle filtering can be views as survival of the fittest,
where the particles with a large weight are more likely to be selected
(and can be selected multiple times) than the particles with a small
weight.

Also, resampling is performed only when $N_{eff}\approx\left(\sum_{i=1}^{N}w_{j,i}^{2}\right)^{-1}<N_{min}$
\cite{Doucet2000} with $N_{min}=0.7N$. That criterion ensures that
resampling only occurs when new data is available for a certain source.
Otherwise, this would cause unnecessary reduction in particle diversity,
due to some particles randomly disappearing.

\section{Results\label{sec:Results}}

Results for the localisation are obtained using the robot described
in Section \ref{sec:Experimental-Setup} with the C1 and C2 configurations
in the E1 and E2 environments. Running the localisation system in
real-time currently requires 30\% of a 1.6~GHz Pentium-M CPU. Due
to the low complexity of the particle filtering algorithm, we are
able to use 1000 particles per source without noticeable increase
in complexity. This also means that the CPU time does not increase
significantly with the number of sources present. For all tasks, configurations
and environments, all parameters have the same value, except for the
reverberation decay $\gamma$, which is set to 0.65 ($T_{60}=350\:\mathrm{ms}$)
in the E1 environment and 0.85 ($T_{60}=910\:\mathrm{ms}$) in the
E2 environment. In both cases, the Signal-to-Reverberant Ratio (SRR)
$\delta$ is set to 3.3 (5.2 dB).

\subsection{Characterisation}

The system is characterised in environment E1 in terms of detection
reliability and accuracy. Detection reliability is defined as the
capacity to detect and localise sounds to within 10 degrees, while
accuracy is defined as the localisation error for sources that are
detected. We use three different types of sound: a hand clap, the
test sentence (``Spartacus, come here''), and a burst of white noise
lasting 100 ms. The sounds are played from a speaker placed at different
locations around the robot and at three different heights: 0.1 m,
1 m, 1.4 m.

\subsubsection{Detection Reliability}

Detection reliability is tested at distances (measured from the centre
of the array) ranging from $1\:\mathrm{m}$ (a normal distance for
close interaction) to $7\:\mathrm{m}$ (limitations of of the room).
Three indicators are computed: correct localisation (within 10 degrees),
reflections (incorrect elevation due to floor and ceiling), and other
errors (repeated detection or large error). For all indicators, we
compute the number of occurrences divided by the number of sounds
played. This test includes 1440 sounds at a 22.5$^{\circ}$ interval
for 1 m and 3 m, and 360 sounds at a 90$^{\circ}$ interval for 5
m and 7 m). 

Results are shown in Table \ref{cap:Detection-reliability-C1C2} for
both C1 and C2 configurations\footnote{The total for a configuration does not have to be equal to 100\% percent
because there are cases where nothing is detected and other cases
where an error occurs in addition to correct localisation.}. In configuration C1, results show near-perfect reliability even
at seven meter distance. For C2, we noted that the reliability depends
on the sound type, so detailed results for different sounds are provided
in Table \ref{cap:Detection-clap-speech-noise}, showing that only
hand clap sounds cannot be reliably detected passed one meter. We
expect that a human would have achieved a score of 100\% for this
reliability test.

Like most localisation algorithms, our system is unable to detect
pure tones. This behaviour is explained by the fact that sinusoids
occupy only a very small region of the spectrum and thus have a very
small contribution to the cross-correlations with the proposed weighting.
It must be noted that tones tend to be more difficult to localise
than wideband signals even for the human auditory system\footnote{Observed from personal experience.}. 

\begin{table}[th]
\caption{Detection reliability for C1 and C2 configurations.\label{cap:Detection-reliability-C1C2}}

\begin{center}%
\begin{tabular}{|c||c|c||c|c||c|c|}
\hline 
Distance & \multicolumn{2}{c||}{Correct (\%)} & \multicolumn{2}{c||}{Reflection (\%)} & \multicolumn{2}{c|}{Other error (\%)}\tabularnewline
\hline 
 & C1 & C2 & C1 & C2 & C1 & C2\tabularnewline
\hline 
\hline 
1 m & 100 & 94.2 & 0.0 & 7.3 & 0.0 & 1.3\tabularnewline
\hline 
 3 m & 99.4 & 80.6 & 0.0 & 21.0 & 0.3 & 0.1\tabularnewline
\hline 
5 m & 98.3 & 89.4 & 0.0 & 0.0 & 0.0 & 1.1\tabularnewline
\hline 
7 m & 100 & 85.0 & 0.6 & 1.1 & 0.6 & 1.1\tabularnewline
\hline 
\end{tabular}\end{center}
\end{table}

\begin{table}[th]
\caption{Correct localisation rate as a function of sound type and distance
for C2 configuration.\label{cap:Detection-clap-speech-noise}}

\begin{center}%
\begin{tabular}{|c|c|c|c|}
\hline 
Distance & Hand clap (\%) & Speech (\%) & Noise burst (\%)\tabularnewline
\hline 
\hline 
1 m & 88.3 & 98.3 & 95.8\tabularnewline
\hline 
3 m & 50.8 & 97.9 & 92.9\tabularnewline
\hline 
5 m & 71.7 & 98.3 & 98.3\tabularnewline
\hline 
7 m & 61.7 & 95.0 & 98.3\tabularnewline
\hline 
\end{tabular}\end{center}
\end{table}

\subsubsection{Localisation Accuracy}

In order to measure the accuracy of the localisation system, we use
the same setup as for measuring reliability, with the exception that
only distances of $1\:\mathrm{m}$ and $3\:\mathrm{m}$ are tested
(1440 sounds at a 22.5$^{\circ}$ interval) due to limited space available
in the testing environment. Neither distance nor sound type has significant
impact on accuracy. The root mean square accuracy results are shown
in Table \ref{cap:Localization-accuracy} for configurations C1 and
C2. Both azimuth and elevation are shown separately. According to
\cite{Hartmann1983,Rakerd2004}, human sound localisation accuracy
ranges between two and four degrees in similar conditions. The localisation
accuracy of our system is thus equivalent or better than human localisation
accuracy.

\begin{table}[th]
\caption{Localisation accuracy (root mean square error).\label{cap:Localization-accuracy}}

\begin{center}%
\begin{tabular}{|c|c|c|}
\hline 
Localisation error & C1 (deg) & C2 (deg)\tabularnewline
\hline 
\hline 
Azimuth & 1.10 & 1.44\tabularnewline
\hline 
Elevation & 0.89 & 1.41\tabularnewline
\hline 
\end{tabular}\end{center}
\end{table}

\subsection{Source Tracking}

\begin{figure}[th]
\center{\subfloat[Moving sources]{

\includegraphics[width=1.5in,keepaspectratio]{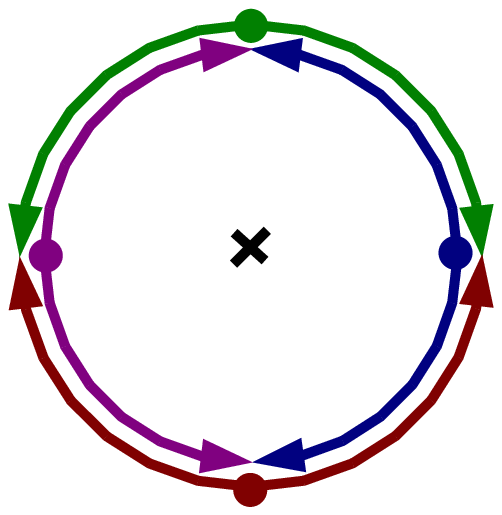}}\subfloat[Moving robot]{

\includegraphics[width=1.5in,keepaspectratio]{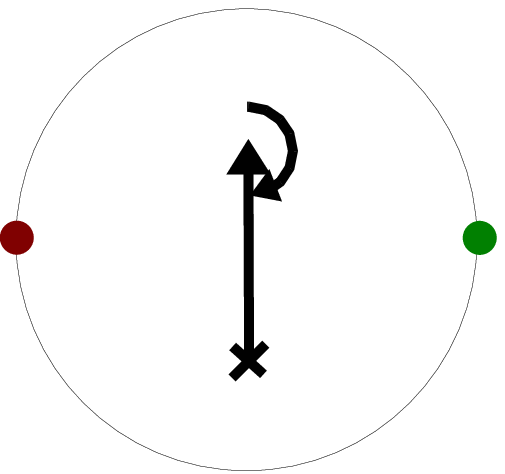}}\subfloat[Sources with intersecting trajectories]{

\includegraphics[width=1.5in,keepaspectratio]{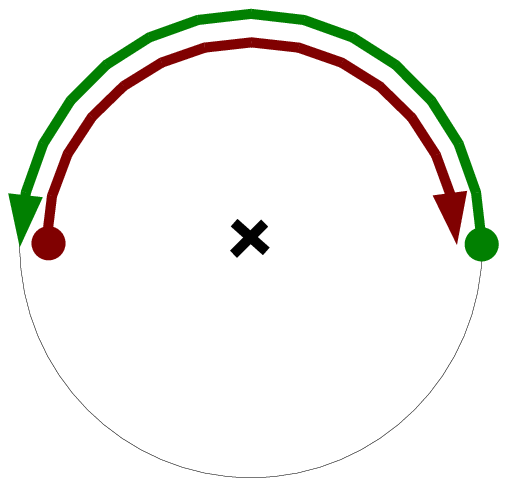}}}

\caption{Source trajectories (robot represented as an X, sources represented
with dots).\label{cap:Source-trajectories}}
\end{figure}

We measure the tracking capabilities of the system for multiple sound
sources. These are performed using the C2 configuration in both E1
and E2 environments. In all cases, the distance between the robot
and the sources is approximately two meters. The azimuth is shown
as a function of time for each source. The elevation is not shown
as it is almost the same for all sources during these tests. The trajectories
of the sources in the three experiments are shown in Figure \ref{cap:Source-trajectories}.
For each of the three cases, only one experiment was performed so
that no selection would have to be made about which trial to display.

\subsubsection{Moving Sources\label{sub:Moving-Sources}}

In a first set of trials, four people were told to talk continuously
(reading a text with normal pauses between words) to the robot while
moving, as shown on Figure \ref{cap:Source-trajectories}a. Each person
walked 90 degrees towards the left of the robot before walking 180
degrees towards the right. 

\begin{figure}[th]
\begin{center}\subfloat[E1]{

\includegraphics[width=2.5in]{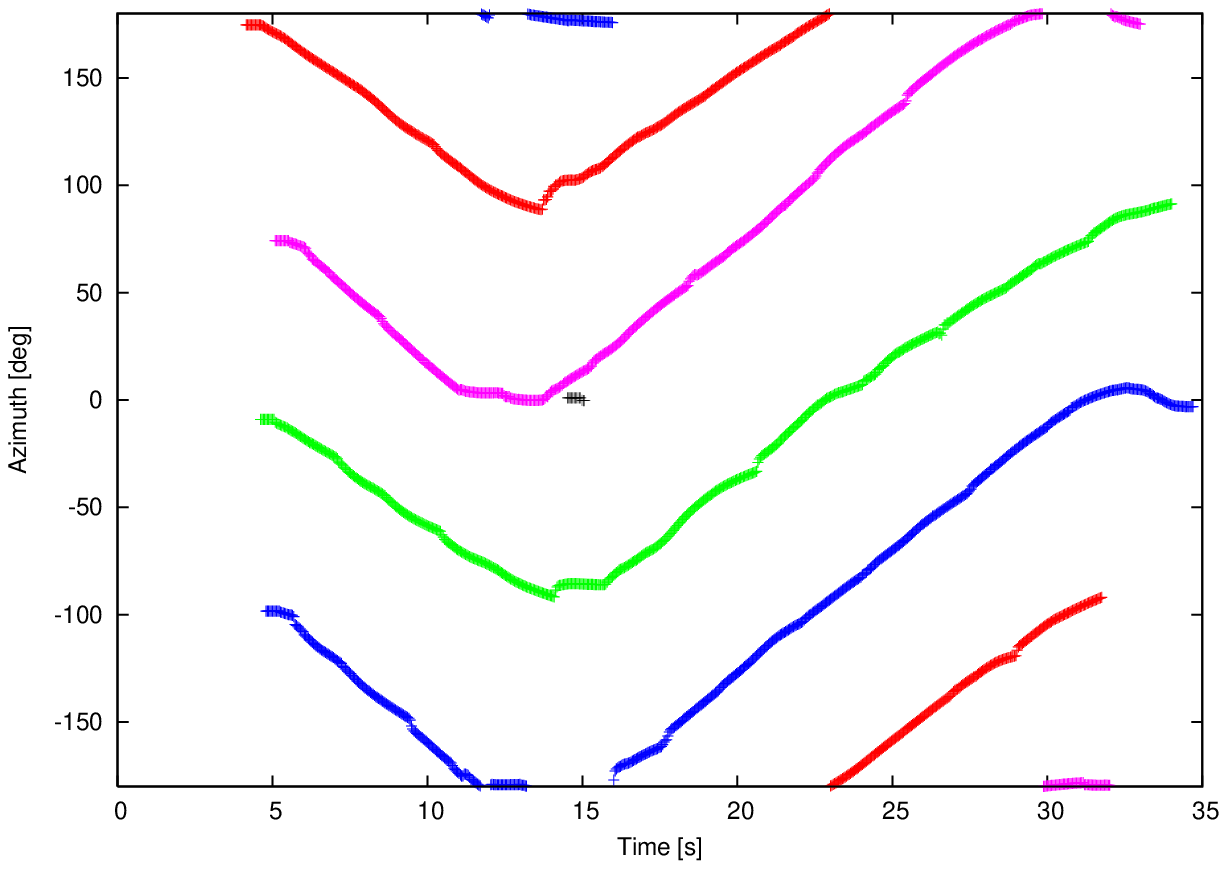}}\subfloat[E2]{

\includegraphics[width=2.5in]{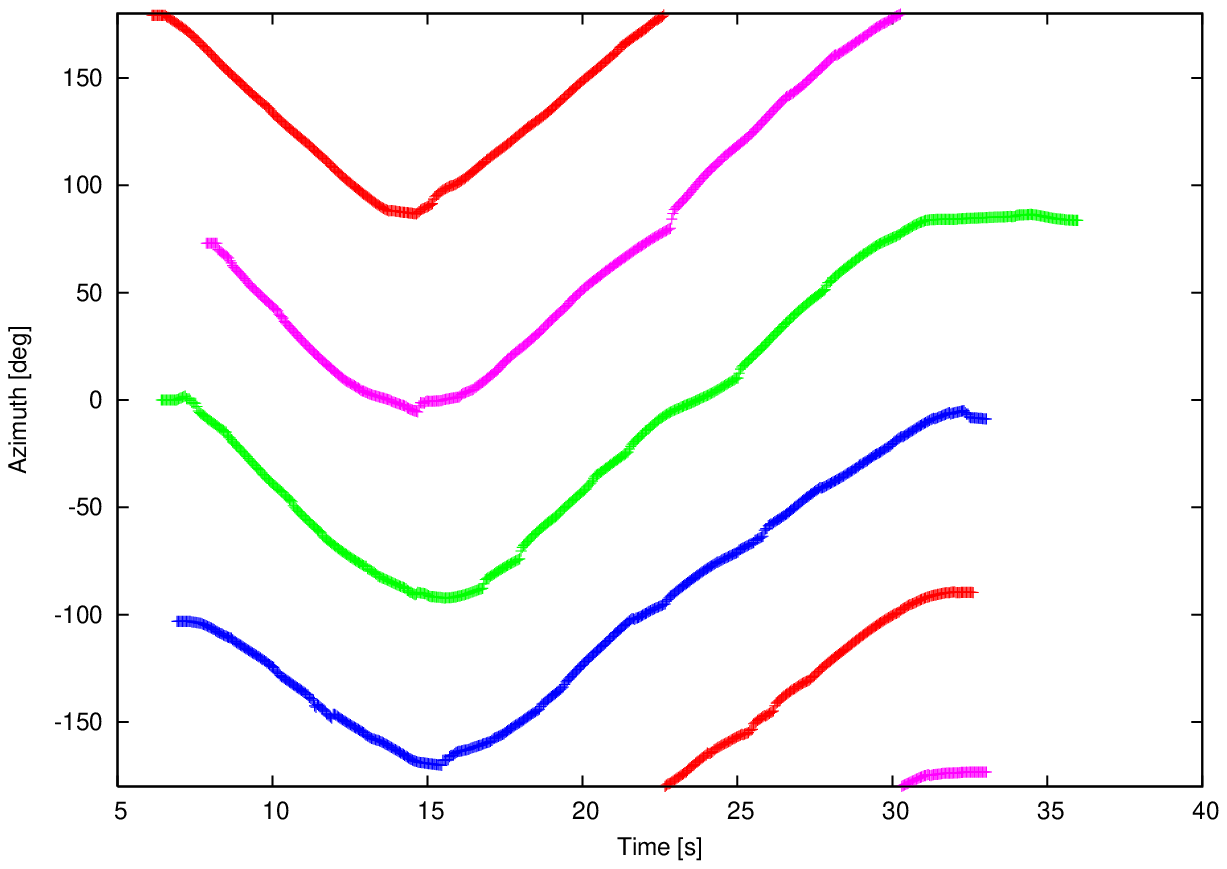}}\end{center}

\caption[Four speakers moving around a stationary robot. False detection shown in black.]{Four speakers moving around a stationary robot. False detection shown in black (around $t=15\:\mathrm{s}$ in E1).}\label{cap:Four-speakers-moving}

\begin{comment}
Four speakers moving around a stationary robot. False detection shown
in black. 
\end{comment}
\end{figure}

Results are presented in Figure \ref{cap:Four-speakers-moving} for
delayed estimation (\emph{T}=500 ms). In both environments, the source
estimated trajectories are consistent with the trajectories of the
four speakers and only one false detection was present for a short
period of time.

\subsubsection{Moving Robot}

\begin{figure}[th]
\begin{center}\subfloat[E1]{

\includegraphics[width=2.5in]{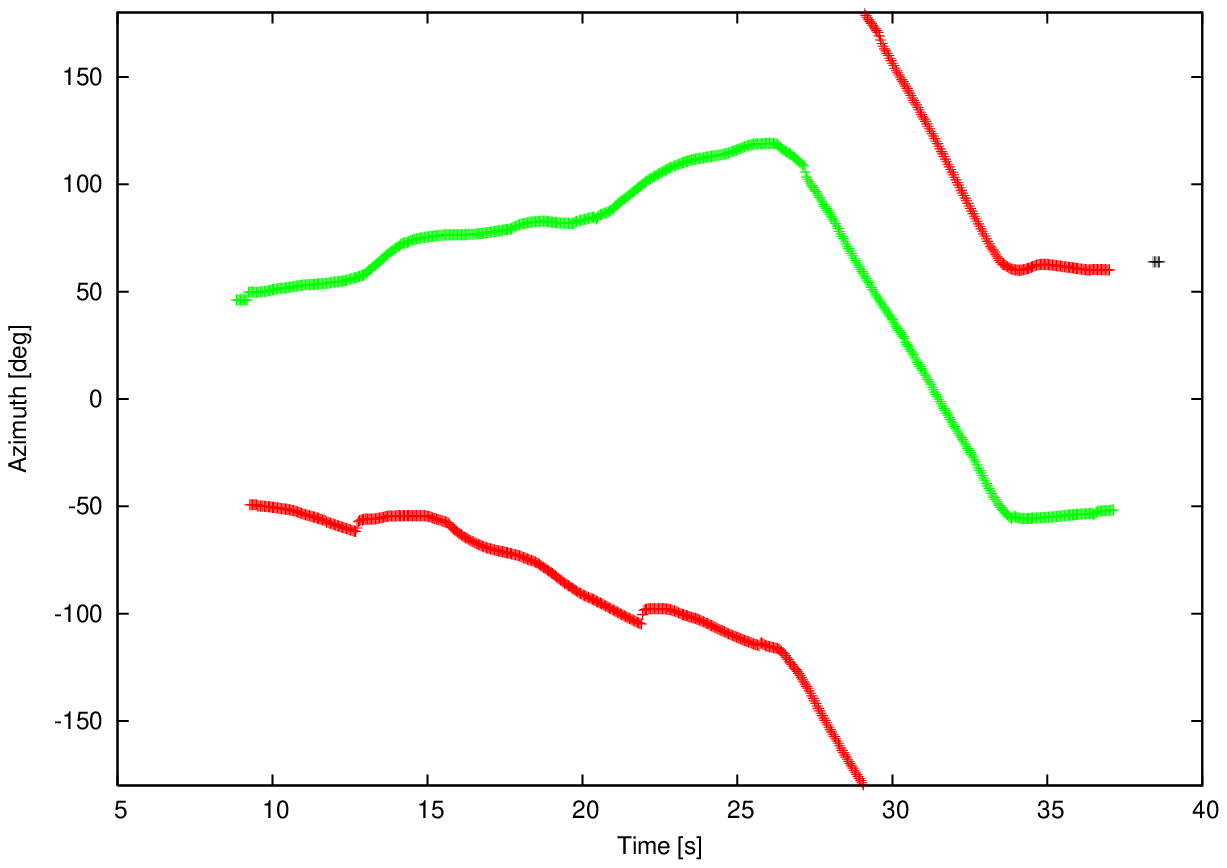}}\subfloat[E2]{

\includegraphics[width=2.5in]{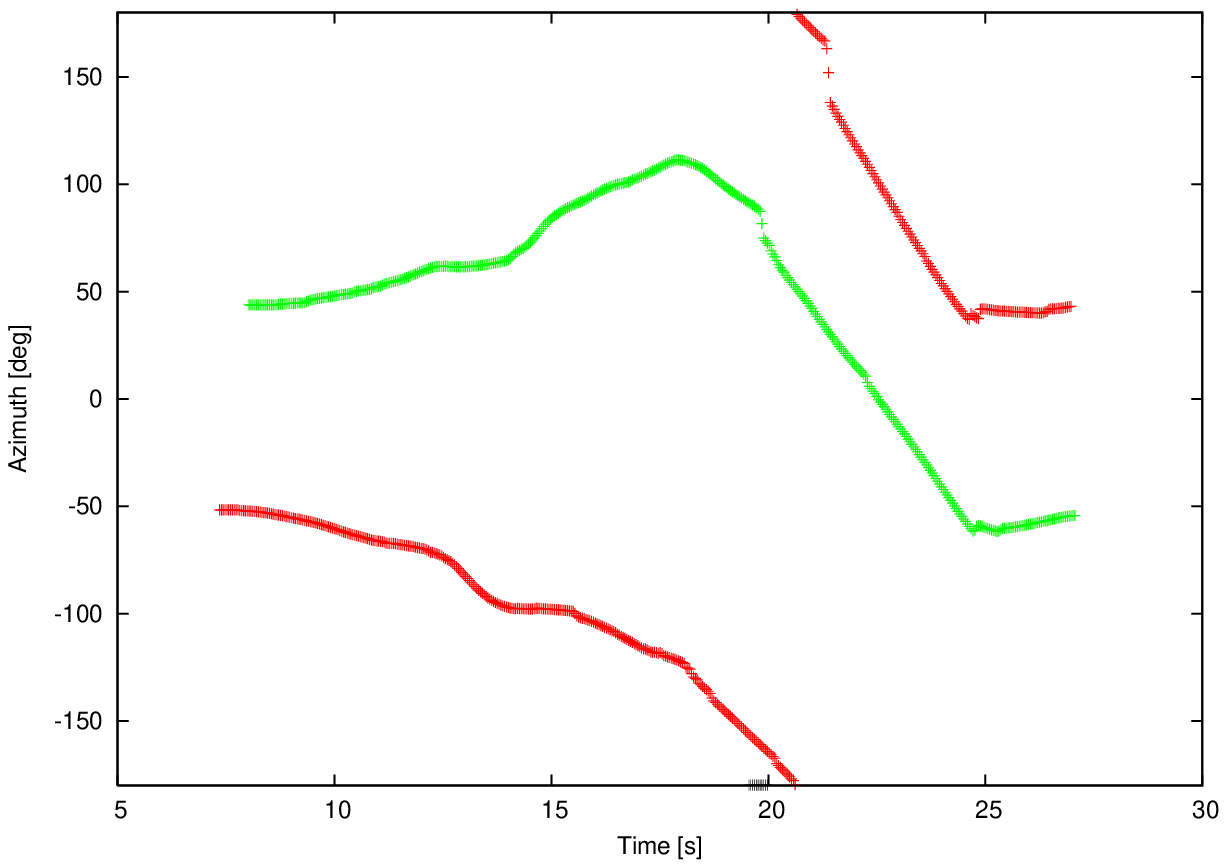}}\end{center}

\caption[Two stationary speakers with the robot moving. False detection shown in black.]{Four speakers moving around a stationary robot.Two stationary speakers with the robot moving. False detection shown in black (around $t=38\:\mathrm{s}$ in E1 and around $t=20\:\mathrm{s}$ in E2).}\label{cap:Two-speakers-robot-moving}

\begin{comment}
Two stationary speakers with the robot moving. False detection shown
in black. 
\end{comment}
\end{figure}

Tracking capabilities of our system are also evaluated in the context
where the the robot is moving, as shown on Figure \ref{cap:Source-trajectories}b.
In this experiment, two people are talking continuously to the robot
as it is passing between them. The robot then makes a half-turn to
the left. Results are presented in Figure \ref{cap:Two-speakers-robot-moving}
for delayed estimation (\emph{T}=500 ms). Once again, the estimated
source trajectories are consistent with the trajectories of the sources
relative to the robot for both environments. Only one false detection
was present for a short period of time.

\subsubsection{Sources with Intersecting Trajectories}

\begin{figure}[th]
\begin{center}\subfloat[E1]{

\includegraphics[width=2.5in]{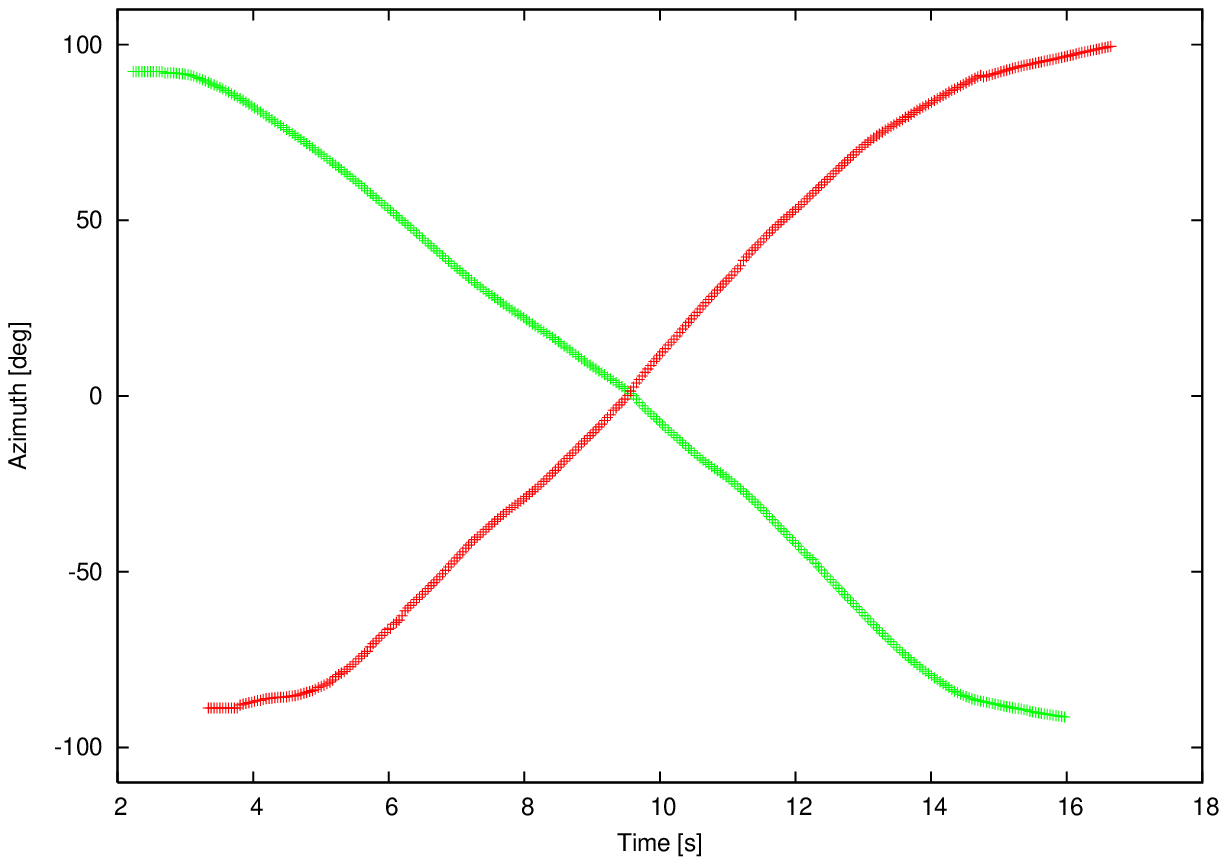}}\subfloat[E2]{

\includegraphics[width=2.5in]{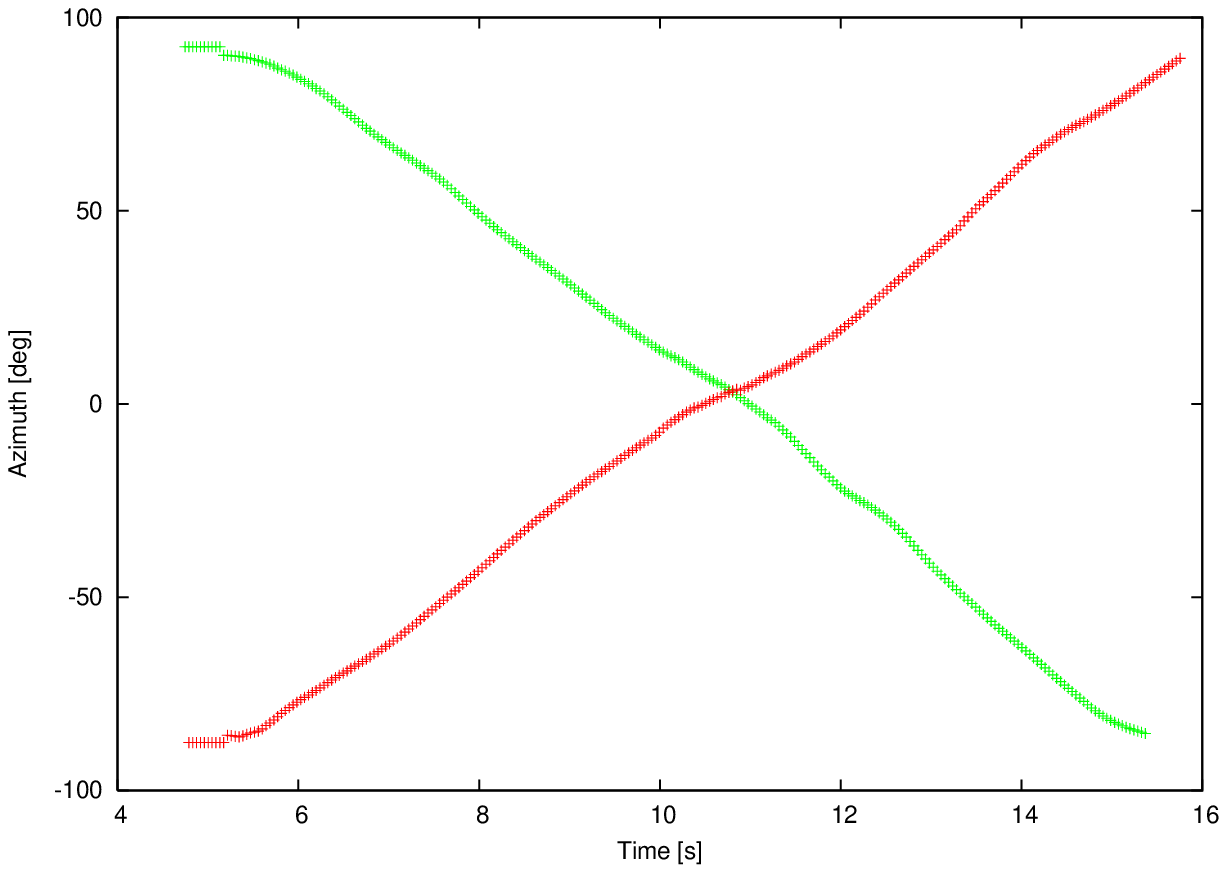}}\end{center}

\caption{Two speakers intersecting in front of the robot.\label{cap:Two-speakers-crossing}}
\end{figure}

In this experiment, two moving speakers are talking continuously to
the robot, as shown on Figure \ref{cap:Source-trajectories}c. They
start from each side of the robot, intersecting in front of the robot
before reaching the other side. Results for delayed estimation (\emph{T}=500
ms) are presented in Figure \ref{cap:Two-speakers-crossing} and show
that the particle filter is able to keep track of each source. This
result is possible because the prediction step (Section \ref{sub:Prediction})
imposes some inertia to the sources.

\subsubsection{Number of Microphones}

\begin{figure}[th]
\begin{center}\includegraphics[width=12cm,keepaspectratio]{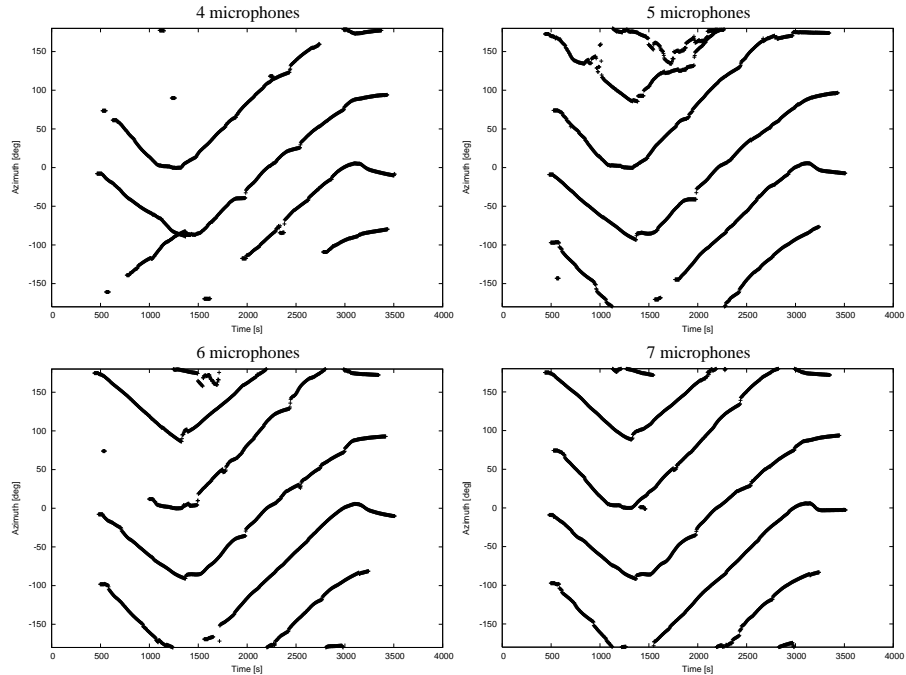}\end{center}

\caption[Tracking of four sources using C2 in the E1 environment, using 4 to 7 microphones.]{Tracking of four sources using C2 in the E1 environment, using 4 to 7 microphones ($E_T$ adjusted to be proportional to the number of microphone pairs, $T=500\:\textrm{ms}$).}\label{cap:Number-of-microphones}

\begin{comment}
Tracking of four sources using C2 in the E1 environment, using 4 to
7 microphones ($E_{T}$ adjusted for the number of microphones, \emph{T}=500
ms).
\end{comment}
\end{figure}

These results evaluate how the number of microphones used affect the
system capabilities. To do so, we use the same recording as in \ref{sub:Moving-Sources}
for C2 in E1 with only a subset of the microphone signals to perform
localisation. Since a minimum of four microphones are necessary for
localising sounds without ambiguity\footnote{With three microphones, it is not possible to distinguish sources
on either side of the plane formed by the microphones.}, we evaluate the system for four to seven microphones (selected arbitrarily
as microphones number $1$ through $N$). Comparing results of Figure
\ref{cap:Number-of-microphones} to those obtained in Figure \ref{cap:Four-speakers-moving}
for E1, it can be observed that tracking capabilities degrade gracefully
as microphones are removed. While using seven microphones makes little
difference compared to the baseline of eight microphones, the system
is unable to reliably track more than two of the sources when only
four microphones are used. Although there is no theoretical relationship
between the number of microphones and the maximum number of sources
that can be tracked, this clearly shows how the redundancy added by
using more microphones can help in the context of sound source localisation.

\subsubsection{Audio Bandwidth\label{sub:Audio-Bandwidth}}

\begin{figure}[th]
\begin{center}\subfloat[8 kHz sampling rate]{

\includegraphics[width=2.5in,keepaspectratio]{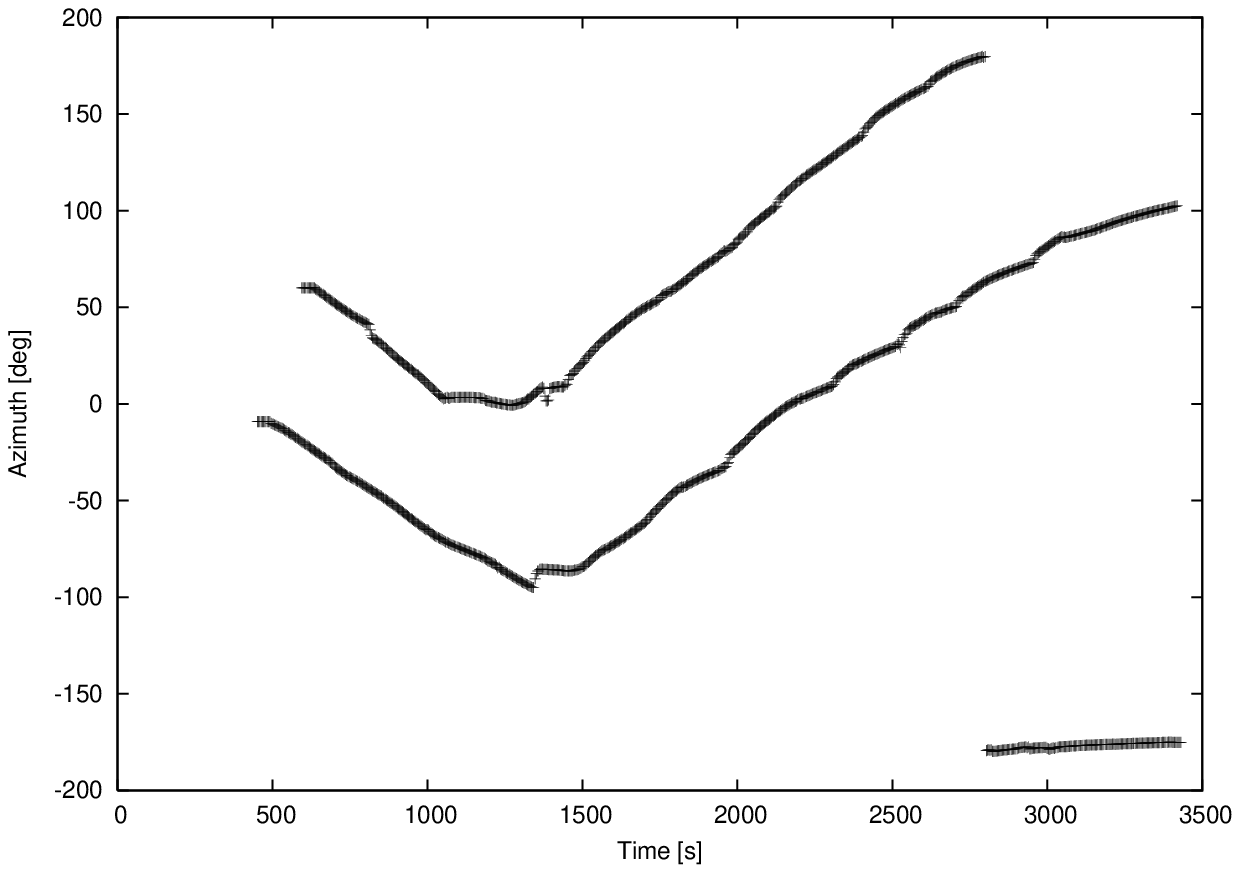}}\subfloat[16 kHz sampling rate]{

\includegraphics[width=2.5in,keepaspectratio]{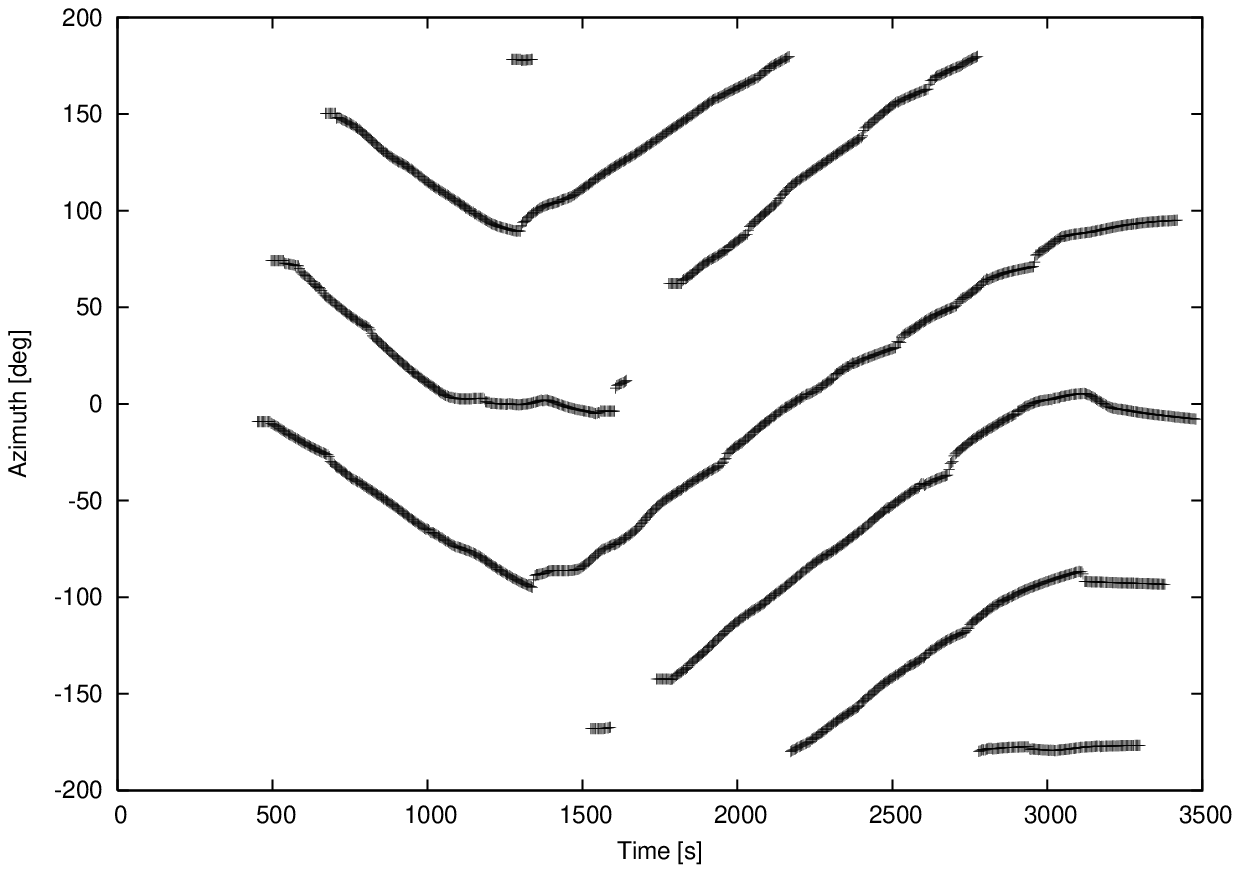}}\end{center}

\caption{Tracking of four sources using C2 in the E1 environment for reduced
audio bandwidth.\label{cap:Tracking-bandwidth}}
\end{figure}

We evaluate the impact of audio bandwidth on the tracking capabilities
using the same recording as in \ref{sub:Moving-Sources} for C2 in
E1. The input signal is low-pass filtered at different frequencies,
while the rest of the processing is still performed at 48 kHz, so
that the cross-correlations are computed at higher resolution. Figure
\ref{cap:Tracking-bandwidth} presents tracking results corresponding
to 8 kHz sampling rate and 16 kHz sampling rate, since those sampling
rate have been used in other work \cite{Asano2001,Bechler2004,Ward2002b,Mungamuru2004}.
It can be observed that results are significantly degraded when compared
to Figure \ref{cap:Four-speakers-moving}, especially in the case
of 8 kHz sampling, where only two of the four sources could be tracked.
This can be explained by the fact that the spectrum of speech extends
beyond 8 kHz and that those higher frequencies can still contribute
to the localisation process. Figure \ref{cap:Tracking-bandwidth}
justifies our use of the full 20 kHz audio bandwidth.

\subsection{Localisation and Tracking for Robot Control}

This experiment is performed in real-time and consists of making the
robot follow the person speaking to it. At any time, only the source
present for the longest time is considered. When the source is detected
in front (within 10 degrees) of the robot, it is made to go forward.
At the same time, regardless of the angle, the robot turns toward
the source in such a way as to keep the source in front. Using this
simple control system, it is possible to control the robot simply
by talking to it, even in noisy and reverberant environments. 

This has been tested by controlling the robot going from environment
E1 to environment E2, having to go through corridors and an elevator,
speaking to the robot with normal intensity at a distance ranging
from one meter to three meters. The system worked in real-time\footnote{A video of the experiment can be downloaded from http://www.gel.usherb.ca/laborius/projects/Audible/JMLocalization.mov},
providing tracking data at a rate of 25 Hz with no additional delay
(\emph{T}=0) on the estimator. The robot reaction time is limited
mainly by the inertia of the robot. One problem we encountered during
the experiment is that when going through corridors, the robot would
sometimes detect reflections on the walls in addition to the real
sources. Fortunately, the fact that the robot considers only the oldest
source present reduces problems from both reflections and noise sources.

\section{Discussion\label{sec:Conclusion}}

Using an array of eight microphones, we have implemented a system
that is able to localise and track simultaneous moving sound sources
in the presence of noise and reverberation, at distances up to seven
meters. We have also demonstrated that the system can successfully
control the motion of a robot in real-time, using only the direction
of sounds. The tracking capabilities demonstrated result from combining
our frequency-domain steered beamformer with a particle filter tracking
multiple sources. Moreover, the solution we found to the source-observation
assignment problem is also applicable to other multiple objects tracking
problems, like visual tracking.

A robot using the proposed system has access to a rich, robust and
useful set of information derived from its acoustic environment. This
can certainly affect its ability of making autonomous decisions in
real life settings, and show higher intelligent behaviour. Also, because
the system is able to localise multiple sound sources, it can be exploited
by a sound separation algorithm and enable speech recognition to be
performed. This allows to identify the sound sources so that additional
relevant information can be obtained from the acoustic environment. 

The main aspect of the localisation algorithm that could be improved
is robustness to sound reflections on hard surfaces (such as the floor,
the walls and the ceiling) that sometimes cause a source to be detected
at multiple locations at the same time. For this, it may be possible
to use knowledge of the environment to eliminate impossible hypotheses
(such as a speech source coming from below the floor).

\chapter{Sound Source Separation\label{cha:Sound-Source-Separation}}

In this chapter, we address the problem of separating individual sound
sources from a mixture of sounds. The human hearing sense is very
good at focusing on a single source of interest despite all kinds
of interferences. We generally refer to this situation as the \emph{cocktail
party effect}, where a human listener is able to follow a conversation
even when several people are speaking at the same time. For a mobile
robot, it means being able to separate each sound source present in
the environment at any moment.

Working toward that goal, we have developed a two-step approach for
performing sound source separation on a mobile robot equipped with
an array of eight microphones. The first step consists of linear separation
based on a simplified version of the Geometric Source Separation (GSS)
approach, proposed by Parra and Alvino \cite{Parra2002Geometric},
with a faster stochastic gradient estimation and shorter time frames
estimations. The second step is a generalisation of beamformer post-filtering
for multiple sources and uses adaptive spectral estimation of background
noise and interfering sources to enhance the signal produced during
the initial separation. The novelty of this post-filter resides in
the fact that, for each source of interest, the noise estimate is
decomposed into stationary, reverberation, and transient components
assumed to be due to leakage between the outputs of the initial separation
stage.

The chapter is organised as follows. Section \ref{sec:Related-Work-SSS}
presents prior art in multi-microphone separation of sound sources.
Section \ref{sec:Separation-overview} gives an overview of the complete
separation algorithm. Section \ref{sec:LSS} presents the linear separation
algorithm and Section \ref{sec:Post-filter} describes the proposed
post-filter. Results are presented in Section \ref{sec:Separation-Results},
followed by a discussion in Section \ref{sec:Separation-Discussion}.

\section{Related Work\label{sec:Related-Work-SSS}}

There are several approaches to sound source separation from multiple
microphones. Most fall either into the Blind Source Separation (BSS)
or beamforming categories. Blind Source Separation and Independent
Component Analysis (ICA) are ways to recover the original (unmixed)
sources with no prior knowledge, other than the fact that all sources
are statistically independent. Several criteria exist for expressing
the independence of the sources, either based on information theory
(e.g., Kullback-Leiber divergence) \cite{HaykinNN} or based on statistics
(e.g., maximum likelihood) \cite{CARDOSO}. Blind source separation
has been applied to audio \cite{SARUWATARI,KOUTRAS2}, often in the
form of frequency domain ICA \cite{ARAKI,MUKAI,RICKARD}. Recently,
Araki \emph{et al.} \cite{Araki2004} have applied ICA to the separation
of three sources using only two microphones.

One drawback of BSS and ICA techniques is that they require assumptions
to be made about the statistics of the sources to be separated. These
assumptions are usually contained in the contrast function \cite{HaykinBSS_3}
or other simplifying hypotheses \cite{HaykinBSS_5}. Also, while the
weak assumptions made by BSS and ICA can be considered as an advantage
of the method for certain applications, we consider it to be a weakness
in the context of microphone array sound source separation because
these techniques do not use the knowledge about the location of the
sound sources.

A more ``classical'' way of isolating a source coming from a certain
direction is beamforming. Unlike BSS, beamforming assumes that the
transfer function between the source of interest and the different
microphones is known approximately. It is thus possible to optimise
the beamformer parameters in such a way as to minimise noise while
conserving perfect response for the signal of interest. This is known
as the Minimum Variance Distortionless Response (MVDR) criterion \cite{HaykinAFT}.

The beamforming technique most widely used today is the Generalised
Sidelobe Canceller (GSC), originally proposed by Griffiths and Jim
\cite{Griffiths}. The GSC algorithm uses a fixed beamformer (delay
and sum) to produce an initial estimation of the source of interest.
Also, a blocking matrix is used to produce noise reference signals
(that do not contain the source of interest) than can be used by a
multiple-input canceller to further reduce noise at the output of
the fixed beamformer. The GSC algorithm can be implemented in the
frequency domain \cite{Gannot} where its components are matrices,
or in the time domain \cite{Hoshuyama} where the components are adaptive
filters.

Recently, a method has been proposed to combine the advantages of
both the BSS and the beamforming approach. The Geometric Source Separation
(GSS) approach proposed by Parra and Alvino \cite{Parra2002Geometric}
uses the assumption that all sources are independent, while using
information about source position through a geometric constraint.
Unlike most variants of BSS, the GSS algorithm uses only second order
statistics as its independence criterion, making it simpler and more
robust. The GSS has been extended to take into account the Head-Related
Transfer Function (HRTF) in order to improve separation \cite{Pedersen2004}.
However, we choose not to apply this technique because of the complexity
involved and the fact that it would make the system more difficult
to adapt to different robots (i.e., different configurations for placing
the microphones).

All methods previously listed can be called Linear Source Separation
(LSS) methods, in that once the demixing parameters are fixed, each
output is a Linear Time Invariant (LTI) transformation from the microphone
inputs. In real-life environments with background noise, reverberation
and imperfect microphones, it is not possible to achieve perfect separation
using LSS methods, so further noise reduction is required. Several
techniques have been developed to remove background noise, including
spectral subtraction \cite{Boll79} and optimal spectral amplitude
estimation \cite{EphraimMalah1984,EphraimMalah1985,CohenNonStat2001}.
Techniques have also been developed specifically to reduce noise at
the output of LSS algorithms, generally beamformers. Most of these
post-filtering techniques address reduction of stationary background
noise \cite{zelinski-1988,McCowan2000,mccowan-icassp2002}. Recently,
a multi-channel post-filter taking into account non-stationary interferences
was proposed by Cohen \cite{CohenArray2002}.

\section{System Overview\label{sec:Separation-overview}}

\begin{figure}[th]
\begin{center}\includegraphics[width=0.45\paperwidth,keepaspectratio]{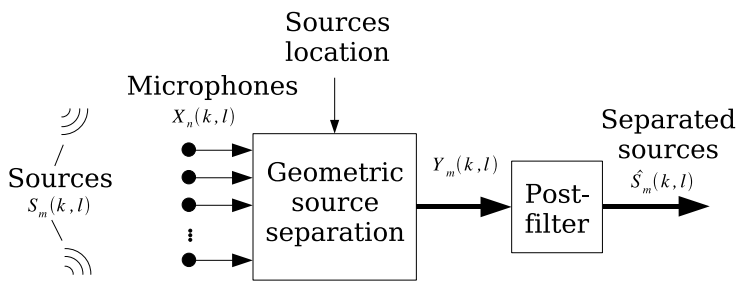}\end{center}

\caption{Overview of the separation system.\label{cap:Separation-Overview}}
\end{figure}

The proposed sound separation algorithm, as shown in Figure \ref{cap:Separation-Overview},
is composed of three parts:
\begin{enumerate}
\item A microphone array;
\item A linear source separation algorithm (LSS) implemented as a variant
of the Geometric Source Separation (GSS) algorithm;
\item A multi-source post-filter.
\end{enumerate}

The microphone array is composed of a number of omni-directional elements
mounted on the robot. The microphone signals are combined linearly
in a first-pass separation algorithm. The output of this initial separation
is then enhanced by a (non-linear) post-filter designed to optimally
attenuate the remaining noise and interference from other sources.

We assume that these sources are detected and that their position
is estimated by the localisation and tracking algorithm described
in Chapter \ref{cha:Sound-Source-Localization}. We also assume that
sources may appear, disappear or move at any time. It is thus necessary
to maximise the adaptation speed for both the LSS and the multi-source
post-filter.

\section{Linear Source Separation\label{sec:LSS}}

The LSS algorithm we propose in this section is based on the Geometric
Source Separation (GSS) approach proposed by Parra and Alvino \cite{Parra2002Geometric}.
Unlike the Linearly Constrained Minimum Variance (LCMV) beamformer
that minimises the output power subject to a distortionless constraint,
GSS explicitly minimises cross-talk, leading to faster adaptation.
The method is also interesting for use in the mobile robotics context
because it allows easy addition and removal of sources. Using some
approximations described in Subsection \ref{sub:Stochastic-Gradient-Adaptation},
it is also possible to implement separation with relatively low complexity
(i.e., complexity that grows linearly with the number of microphones),
as described in \cite{ValinIROS2004}.

\subsection{Geometric Source Separation\label{sub:Geometric-Source-Separation}}

The method operates in the frequency domain on overlapped frames of
21 ms (1024 samples at 48 kHz). Let $S_{m}(k,\ell)$ be the real (unknown)
sound source $m$ at time frame $\ell$ and for discrete frequency
$k$. We denote as $\mathbf{s}(k,\ell)$ the vector corresponding
to the sources $S_{m}(k,\ell)$ and matrix $\mathbf{A}(k)$ is the
transfer function leading from the sources to the microphones. The
signal received at the microphones is thus given by:
\begin{equation}
\mathbf{x}(k,\ell)=\mathbf{A}(k)\mathbf{s}(k,\ell)+\mathbf{n}(k,\ell)\label{eq:GSS_mixing}
\end{equation}
where $\mathbf{n}(k,\ell)$ is the non-coherent background noise received
at the microphones. The matrix $\mathbf{A}(k)$ can be estimated using
the result of a sound localisation algorithm. Assuming that all transfer
functions have unity gain, the elements of $\mathbf{A}(k)$ can be
expressed as:
\begin{equation}
a_{ij}(k)=e^{-\jmath2\pi k\delta_{ij}}\label{eq:GSS_a_ij}
\end{equation}
where $\delta_{ij}$ is the time delay (in samples) to reach microphone
$i$ from source $j$. 

The separation result is then defined as $\mathbf{y}(k,\ell)=\mathbf{W}(k,\ell)\mathbf{x}(k,\ell)$,
where $\mathbf{W}(k,\ell)$ is the separation matrix that must be
estimated. This is done by providing two constraints (the index $\ell$
is omitted for the sake of clarity):
\begin{enumerate}
\item Decorrelation of the separation algorithm outputs, expressed as $\mathbf{R}_{\mathbf{yy}}(k)-\mathrm{diag}\left[\mathbf{R}_{\mathbf{yy}}(k)\right]=\mathbf{0}$\footnote{Assuming non-stationary sources, second order statistics are sufficient
for ensuring independence of the separated sources.}.
\item The geometric constraint $\mathbf{W}(k)\mathbf{A}(k)=\mathbf{I}$,
which ensures unity gain in the direction of the source of interest
and places zeros in the direction of interferences.
\end{enumerate}
In theory, constraint 2) could be used alone for separation (the method
is referred to as LS-C2 in \cite{Parra2002Geometric}), but in practice,
the method does not take into account reverberation or errors in localisation.
It is also subject to instability if $\mathbf{A}(k)$ is not invertible
at a specific frequency. When used together, constraints 1) and 2)
are too strong. For this reason, we use a ``soft'' constraint (referred
to as GSS-C2 in \cite{Parra2002Geometric}) combining 1) and 2) in
the context of a gradient descent algorithm.

Two cost functions are created by computing the square of the error
associated with constraints 1) and 2). These cost functions are respectively
defined as:
\begin{eqnarray}
J_{1}(\mathbf{W}(k)) & = & \left\Vert \mathbf{R}_{\mathbf{yy}}(k)-\mathrm{diag}\left[\mathbf{R}_{\mathbf{yy}}(k)\right]\right\Vert ^{2}\label{eq:Cost_J1}\\
J_{2}(\mathbf{W}(k)) & = & \left\Vert \mathbf{W}(k)\mathbf{A}(k)-\mathbf{I}\right\Vert ^{2}\label{eq:Cost_J2}
\end{eqnarray}
where the matrix norm is defined as $\left\Vert \mathbf{M}\right\Vert ^{2}=\mathrm{trace}\left[\mathbf{M}\mathbf{M}^{H}\right]$
and is equal to the sum of the square of all elements in the matrix.
The gradient of the cost functions with respect to $\mathbf{W}(k)$
is equal to \cite{Parra2002Geometric}:
\begin{eqnarray}
\frac{\partial J_{1}(\mathbf{W}(k))}{\partial\mathbf{W}^{*}(k)} & = & 4\mathbf{E}(k)\mathbf{W}(k)\mathbf{R}_{\mathbf{xx}}(k)\label{eq:Grad_J1}\\
\frac{\partial J_{2}(\mathbf{W}(k))}{\partial\mathbf{W}^{*}(k)} & = & 2\left[\mathbf{W}(k)\mathbf{A}(k)-\mathbf{I}\right]\mathbf{A}^{\mathrm{H}}(k)\label{eq:Grad_J2}
\end{eqnarray}
where $\mathbf{E}(k)=\mathbf{R}_{\mathbf{yy}}(k)-\mathrm{diag}\left[\mathbf{R}_{\mathbf{yy}}(k)\right]$.

The separation matrix $\mathbf{W}(k)$ is then updated as follows:
\begin{equation}
\mathbf{W}^{n+1}(k)=\mathbf{W}^{n}(k)-\mu\left[\alpha(k)\frac{\partial J_{1}(\mathbf{W}(k))}{\partial\mathbf{W}^{*}(k)}+\frac{\partial J_{2}(\mathbf{W}(k))}{\partial\mathbf{W}^{*}(k)}\right]\label{eq:W_update}
\end{equation}
where $\alpha(f)$ is an energy normalisation factor equal to $\left\Vert \mathbf{R}_{\mathbf{xx}}(k)\right\Vert ^{-2}$
and $\mu$ is the adaptation rate. We use $\mu=0.01$, which performs
adequately for both stationary and moving sources.

\subsection{Proposed Improvements to the GSS algorithm}

We propose two improvements to the GSS algorithm as it is described
in \cite{Parra2002Geometric}. The first modification simplifies the
computation of the correlation matrices $\mathbf{R}_{\mathbf{xx}}(k)$
and $\mathbf{R}_{\mathbf{yy}}(k)$, while the second one introduces
regularisation to prevent large values to appear in the demixing matrix
during the adaptation process.

\subsubsection{Stochastic Gradient Adaptation\label{sub:Stochastic-Gradient-Adaptation}}

Instead of estimating the correlation matrices $\mathbf{R}_{\mathbf{xx}}(k)$
and $\mathbf{R}_{\mathbf{yy}}(k)$ on several seconds of data, our
approach uses instantaneous estimations. This is analogous to the
approximation made in the Least Mean Square (LMS) adaptive filter
\cite{HaykinAFT}. We thus assume that:
\begin{eqnarray}
\mathbf{R}_{\mathbf{xx}}(k) & = & \mathbf{x}(k)\mathbf{x}(k)^{H}\label{eq:Rxx_instant}\\
\mathbf{R}_{\mathbf{yy}}(k) & = & \mathbf{y}(k)\mathbf{y}(k)^{H}\label{eq:Ryy_instant}
\end{eqnarray}

It is then possible to rewrite the gradient of Equation \ref{eq:Grad_J1}
as:
\begin{equation}
\frac{\partial J_{1}(\mathbf{W}(k))}{\partial\mathbf{W}^{*}(k)}=4\left[\mathbf{E}(k)\mathbf{W}(k)\mathbf{x}(k)\right]\mathbf{x}(k)^{H}\label{eq:Grad_J1_instant}
\end{equation}
which only requires matrix-by-vector products, greatly reducing the
complexity of the algorithm. The normalisation factor $\alpha(k)$
can also be simplified as $\left[\left\Vert \mathbf{x}(k)\right\Vert ^{2}\right]^{-2}$.
Although we use instantaneous estimation, the fact that the update
rate is small means that many frames of data are still necessary for
adapting the demixing matrix, so the averaging is performed implicitly.
We have observed that the modification did not cause significant degradation
in performance, while it greatly facilitates real-time implementation.

\subsubsection{Regularisation Term}

Another modification we propose to the GSS algorithm is the addition
of a regularisation. Since it is desirable for the demixing matrix
to be as small as possible (while still respecting the other constraints),
we define the regularisation term as the cost:
\begin{equation}
J_{R}(\mathbf{W}(k))=\lambda\left\Vert \mathbf{W}(k)\right\Vert ^{2}\label{eq:Cost_regul}
\end{equation}

The gradient of the regularisation cost is equal to:
\begin{equation}
\frac{\partial J_{R}(\mathbf{W}(k))}{\partial\mathbf{W}^{*}(k)}=\lambda\mathbf{W}(k)\label{eq:Grad_Regul}
\end{equation}
so we can rewrite Equation \ref{eq:W_update} as:
\begin{equation}
\mathbf{W}^{n+1}(k)=\mathbf{W}^{n}(k)-\mu\left[\alpha(k)\frac{\partial J_{1}(\mathbf{W}(k))}{\partial\mathbf{W}^{*}(k)}+\frac{\partial J_{2}(\mathbf{W}(k))}{\partial\mathbf{W}^{*}(k)}+\frac{\partial J_{R}(\mathbf{W}(k))}{\partial\mathbf{W}^{*}(k)}\right]\label{eq:W_update_Regul}
\end{equation}
or:
\begin{equation}
\mathbf{W}^{n+1}(k)=(1-\lambda\mu)\mathbf{W}^{n}(k)-\mu\left[\alpha(k)\frac{\partial J_{1}(\mathbf{W}(k))}{\partial\mathbf{W}^{*}(k)}+\frac{\partial J_{2}(\mathbf{W}(k))}{\partial\mathbf{W}^{*}(k)}\right]\label{eq:W_update_Regul2}
\end{equation}
where we set the regularisation parameter $\lambda$ to 0.5.

\subsection{Initialisation}

The fact that sources can appear or disappear at any time imposes
constraints on the initialisation of the separation matrix $\mathbf{W}(k)$.
The initialisation must provide the following:
\begin{itemize}
\item The initial weights for a new source;
\item Acceptable separation (before adaptation).
\end{itemize}
Furthermore, when a source appears or disappears, other sources must
be unaffected.

One easy way to satisfy both constraints is to initialise the column
of $\mathbf{W}(k)$ corresponding to the new source $m$ as:
\begin{equation}
w_{m,i}(k)=\frac{a_{i,m}^{*}(k)}{N}\label{eq:GSS_I1}
\end{equation}

where $N$ is the number of microphones. This weight initialisation
corresponds to a delay-and-sum beamformer, referred to as the I1 initialisation
method \cite{Parra2002Geometric}. Such initialisation ensures that
prior to adaptation, the performances are at worst equivalent to a
delay-and-sum beamformer.

\section{Multi-Source Post-Filter\label{sec:Post-filter}}

\begin{figure}[th]
\begin{center}\includegraphics[width=0.6\paperwidth,keepaspectratio]{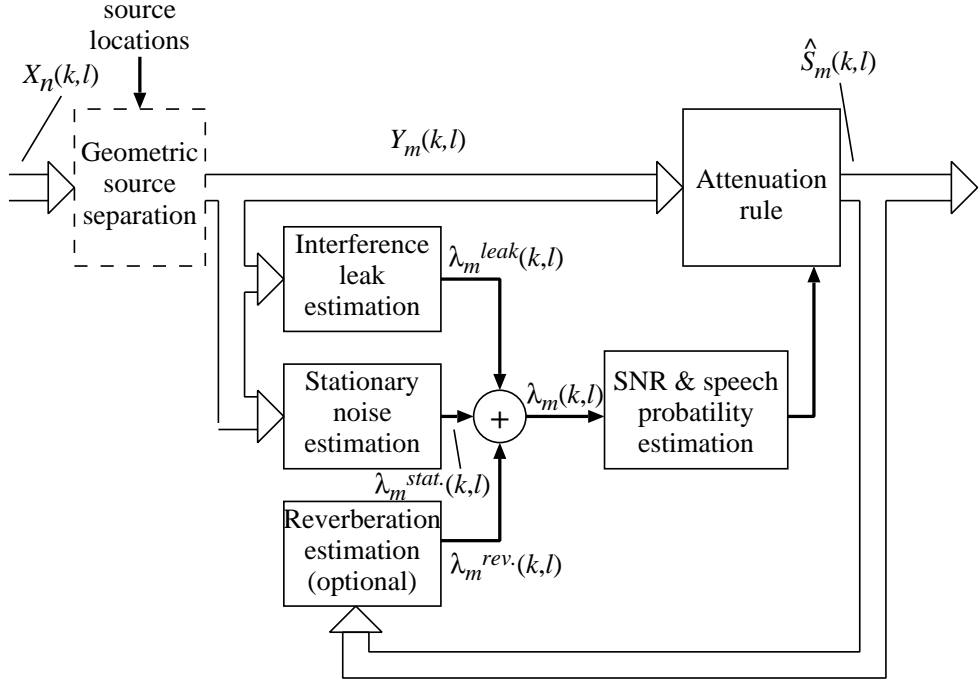}\end{center}

\caption{Overview of the complete separation system. }

$X_{n}(k,\ell),n=0\ldots N-1$: Microphone inputs, $Y_{m}(k,\ell),\:m=0\ldots M-1$:
Inputs to the post-filter, $\hat{S}_{m}(k,\ell)=G_{m}(k,\ell)Y_{m}(k,\ell),\:m=0\ldots M-1$:
Post-filter outputs.\label{cap:Overview-Post-filter}
\end{figure}

In order to enhance the output of the GSS algorithm presented in Section
\ref{sec:LSS}, we derive a frequency-domain post-filter that is based
on the optimal estimator originally proposed by Ephraim and Malah
\cite{EphraimMalah1984,EphraimMalah1985} and further improved by
Cohen and Berdugo \cite{CohenNonStat2001,CohenArray2002,Cohen2004}.
The novelty of our approach, as proposed in \cite{ValinICASSP2004,ValinIROS2004},
resides in the fact that, for a given output of the GSS, the transient
components of the corrupting sources is assumed to be due to leakage
from the other channels during the GSS process. Furthermore, for a
given source, the stationary and the transient components are combined
into a single noise estimator used for noise suppression, as shown
in Figure \ref{cap:Overview-Post-filter}. 

For this post-filter, we consider that all interferences (except the
background noise) are localised (detected) sour\-ces and we assume
that the leakage between outputs is constant. This leakage is due
to reverberation, localisation error, differences in microphone frequency
responses, near-field effects, etc. In addition, the post-filter is
tuned in such a way as to maximise speech recognition accuracy, as
opposed to perceptual quality or objective quality measurement.

\subsection{Noise Estimation}

This section describes the estimation of noise variances that are
used to compute the weighting function $G_{m}(k,\ell)$ by which the
outputs $Y_{m}(k,\ell)$ of the LSS is multiplied to generate a cleaned
signal whose spectrum is denoted $\hat{S}_{m}(k,\ell)$. The noise
variance estimation $\lambda_{m}(k,\ell)$ is expressed as:
\begin{equation}
\lambda_{m}(k,\ell)=\lambda_{m}^{stat.}(k,\ell)+\lambda_{m}^{leak}(k,\ell)\label{eq:noise_estim}
\end{equation}
 where $\lambda_{m}^{stat.}(k,\ell)$ is the estimate of the stationary
component of the noise for source $m$ at frame $\ell$ for frequency
$k$, and $\lambda_{m}^{leak}(k,\ell)$ is the estimate of source
leakage. We compute the stationary noise estimate $\lambda_{m}^{stat.}(k,\ell)$
using the Minima Controlled Recursive Average (MCRA) technique proposed
by Cohen and Berdugo \cite{CohenNonStat2001}.

To estimate $\lambda_{m}^{leak}(k,\ell)$, we assume that the interference
from other sources is reduced by a factor $\eta$ (typically $-10\:\mathrm{dB}\leq\eta\leq-5\:\mathrm{dB}$)
by the separation algorithm (LSS). The leakage estimate is thus expressed
as:
\begin{equation}
\lambda_{m}^{leak}(k,\ell)=\eta\sum_{i=0,i\neq m}^{M-1}Z_{i}(k,\ell)\label{eq:leak_estim}
\end{equation}
 where $Z_{m}(k,\ell)$ is the smoothed spectrum of the $m^{th}$
source $Y_{m}(k,\ell)$, and is recursively defined (with $\alpha_{s}=0.2$,
from informal listening tests) as:
\begin{equation}
Z_{m}(k,\ell)=\alpha_{s}Z_{m}(k,\ell-1)+(1-\alpha_{s})\left|Y_{m}(k,\ell)\right|^{2}\label{eq:smoothed_spectrum}
\end{equation}

In the case where reverberation is present, we can add a third term
to Equation \ref{eq:noise_estim}, so that:
\begin{equation}
\lambda_{m}(k,\ell)=\lambda_{m}^{stat.}(k,\ell)+\lambda_{m}^{leak}(k,\ell)+\sum_{i=0}^{M-1}\lambda_{i}^{rev}(k,\ell)\label{eq:noise_estim_reverb}
\end{equation}
where the reverberation estimate $\lambda_{i}^{rev}(k,\ell)$ for
source $i$ is given by (similarly to Equation \ref{eq:Reverb_weighting}):
\begin{equation}
\lambda_{i}^{rev}(k,\ell)=\gamma\lambda_{i}^{rev}(k,\ell-1)+\frac{(1-\gamma)}{\delta}\left|\hat{S}_{i}(k,\ell-1)\right|^{2}\label{eq:pf-reverb-term}
\end{equation}
This method of taking into account reverberation is similar to the
work by Wu and Wang \cite{Wu2005} but it is used in such a way that
reverberation from the source of interest and from the interfering
sources is considered at the same time. Unlike the work of Wu and
Wang, we do not need inverse filtering to compensate for spectral
coloration because artificial speech recognition (ASR) engines typically
normalise the channel response using cepstral mean subtraction (CMS).

\subsection{Suppression Rule in the Presence of Speech}

We now derive the suppression rule under $H_{1}$, the hypothesis
that speech is present. From here on, unless otherwise stated, the
$m$ and $\ell$ arguments are omitted for clarity and the equations
are given for each $m$ and for each $\ell$. The proposed noise suppression
rule is based on minimum mean-square error (MMSE) estimation of the
log-spectral amplitude. Although earlier work \cite{ValinICASSP2004,ValinIROS2004}
proposed an optimal estimator in the loudness domain ($\left|S(k)\right|^{1/2}$)
to maximise perceptual quality, the choice of the log-domain here
is based on optimal speech recognition accuracy.

Assuming that speech is present ($H_{1}$), the amplitude estimator
$\hat{A}_{H_{1}}(k)$ is defined as:
\begin{equation}
\hat{A}_{H_{1}}(k)=\exp\left(E\left[\log\left|S(k)\right|\:\left|Y(k)\right.,H_{1}\right]\right)=G_{H_{1}}(k)\left|Y(k)\right|\label{eq:amplitude_estim}
\end{equation}
where $G_{H_{1}}(k)$ is the spectral gain assuming that speech is
present. 

The optimal spectral gain is given in \cite{EphraimMalah1985}:
\begin{equation}
G_{H_{1}}(k)=\frac{\xi(k)}{1+\xi(k)}\left\{ \frac{1}{2}\int_{\upsilon(k)}^{\infty}\frac{e^{-t}}{t}dt\right\} \label{eq:Gain_H1}
\end{equation}
where $\gamma(k)\triangleq\left|Y(k)\right|^{2}/\lambda(k)$ and $\xi(k)\triangleq E\left[\left|X(k)\right|^{2}\right]/\lambda(k)$
are respectively the \emph{a posteriori} SNR and the \emph{a priori}
SNR, and $\upsilon(k)\triangleq\gamma(k)\xi(k)/\left(\xi(k)+1\right)$
\cite{EphraimMalah1984,EphraimMalah1985}.

Using the modifications proposed in \cite{CohenNonStat2001} to take
into account speech presence uncertainty, the \emph{a priori} SNR
$\xi(k)$ is estimated recursively as:
\begin{equation}
\hat{\xi}(k,\ell)=(1-\alpha_{p})G_{H_{1}}^{2}(k,\ell-1)\gamma(k,\ell-1)+\alpha_{p}\max\left\{ \gamma(k,\ell)-1,0\right\} \label{eq:xi_decision_directed}
\end{equation}
where the update rate $\alpha_{p}$ is dependent on the \emph{a priori}
SNR as proposed by Cohen \cite{Cohen2004}:
\begin{equation}
\alpha_{p}=\left(\frac{\hat{\xi}(k,\ell)}{1+\hat{\xi}(k,\ell)}\right)^{2}+\alpha_{pmin}\label{eq:variable_ap}
\end{equation}
with $\alpha_{pmin}=0.07$ determined from informal listening tests
to force a minimal amount of adaptation.

\subsection{Optimal Gain Modification Under Speech Presence Uncertainty\label{sec:Optimal-gain-modification}}

In order to take into account the probability of speech presence,
we derive the estimator for the log-domain:
\begin{equation}
\hat{A}(k)=\exp\left(E\left[\log\left|S(k)\right|\:\left|Y(k)\right.\right]\right)\label{eq:optimal-gain}
\end{equation}

Considering $H_{1}$, the hypothesis that speech is present for a
source (source index $m$ and frame index $\ell$ are omitted for
clarity), and $H_{0}$, the hypothesis that speech is absent (as defined
in \cite{CohenNonStat2001}), we obtain:
\begin{eqnarray}
E\left[\left.\log\:A(k)\right|Y(k)\right] & = & p(k)E\left[\left.\log\:A(k)\right|H_{1},Y(k)\right]\nonumber \\
 & + & \left[1-p(k)\right]E\left[\left.\log\:A(k)\right|H_{0},Y(k)\right]\label{eq:optimal-gain2}
\end{eqnarray}
where $p(k)$ is the probability of speech at frequency $k$.

The optimally modified gain is thus given by:
\begin{equation}
G(k)=\exp\left[p(k)\log\:G_{H_{1}}(k)+(1-p(k))\log\:G_{min}\right]\label{eq:optimal_gain3}
\end{equation}
where $G_{H_{1}}(k)$ is defined in Equation \ref{eq:Gain_H1}, and
$G_{min}$ is the minimum gain allowed when speech is absent, which
is set to -20 dB (similarly as in \cite{CohenNonStat2001}) to limit
distortion to the signal. Equation \ref{eq:optimal_gain3} further
simplifies as:
\begin{equation}
G(k)=\left\{ G_{H_{1}}(k)\right\} ^{p(k)}\cdot G_{min}^{1-p(k)}\label{eq:optimal_gain_final}
\end{equation}

The probability of speech presence is computed as:
\begin{equation}
p(k)=\left\{ 1+\frac{\hat{q}(k)}{1-\hat{q}(k)}\left(1+\xi(k)\right)\exp\left(-\upsilon(k)\right)\right\} ^{-1}\label{eq:def_speech_prob}
\end{equation}
where $\hat{q}(k)$ is the \emph{a priori} probability of speech absence
for frequency $k$ and is defined as \cite{CohenNonStat2001}:
\begin{equation}
\hat{q}(k)=\min\left(1-P_{local}(k)P_{global}(k)P_{frame},0.9\right)\label{eq:def_q_a_priori}
\end{equation}
where $P_{local}(k)$, $P_{global}(k)$ and $P_{frame}$ are speech
presence probabilities computed respectively on a local frequency
window, a large frequency window and on the whole frame.

The computation of $P_{local}(k)$, $P_{global}(k)$ and $P_{frame}$
is inspired from Cohen and Berdugo \cite{CohenNonStat2001} and Choi
\cite{Choi2005} so that:
\begin{equation}
P_{\psi}(k)=\frac{1}{1+\left(\frac{\theta}{\zeta_{\psi}(k)}\right)^{2}}\label{eq:apriori-update-rate}
\end{equation}
where $\psi$ can be either \emph{local}, \emph{global} or \emph{frame},
$\theta$ is a soft-decision threshold that we set to -5 dB, and $\zeta_{\psi}(k)$
is a recursive average of the estimated \emph{a priori} SNR:
\begin{equation}
\zeta_{\psi}(k)=\left(1-\alpha_{\zeta}\right)\zeta(k)+\alpha_{\zeta}\sum_{j=-w_{1}}^{w_{2}}h_{\psi}(j)\hat{\xi}(k+j)\label{eq:apriori-average}
\end{equation}
with $\alpha_{\zeta}=0.3$ (determined from listening tests). In Equation
\ref{eq:apriori-average}, $h_{\psi}(j)$ is a Hanning window covering
140 Hz for $P_{local}$, 1400 Hz for $P_{global}$ and the full band
for $P_{frame}$.

\subsection{Post-filter Initialisation}

When a new source appears, post-filter state variables need to be
initialised. Most of these variables may safely be set to zero. The
exception is $\lambda_{m}^{stat.}\left(k,\ell_{0}\right)$, the initial
stationary noise estimation for source $m$. The MCRA algorithm requires
several seconds to produce its first estimate for source $m$, so
it is necessary to find another way to estimate the background noise
until a better estimate is available. This initial estimate is thus
computed using noise estimations at the microphones. Assuming the
delay-and-sum initialisation of the weights from Equation \ref{eq:GSS_I1},
the initial background noise estimate is thus:
\begin{equation}
\lambda_{m}^{stat.}\left(k,\ell_{0}\right)=\frac{1}{N^{2}}\sum_{n=0}^{N-1}\sigma_{x_{n}}^{2}\left(k\right)\label{eq:noise_init}
\end{equation}
where $\sigma_{x_{n}}^{2}\left(k\right)$ is the noise estimation
for microphone $n$.

\section{Results\label{sec:Separation-Results}}

The separation system is tested using the robot described in Section
\ref{sec:Experimental-Setup} in both C1 and C2 configurations. In
order to test the system, three streams of voice were recorded separately,
in a quiet environment. The speech consists of 251 sequences of four
connected digits taken from the AURORA database \cite{Pearce01} played
from speakers located at 90 degrees to the left, in front, and 135
degrees to the right of the robot. The background noise is recorded
on the robot and includes the room ventilation and the internal robot
fans. All four signals are recorded using the same microphone array
and subsequently mixed together. This procedure is required in order
to compute the distance measures (such as SNR) presented in this section.
It is worth noting that although the signals were mixed artificially,
the result still represents real conditions with background noise,
interfering sources, and reverberation. Because a clean reference
signal is required, only the E1 environment is considered (it is not
possible to make clean recordings in the E2 environment). 

In evaluating our source separation system, we use the conventional
signal-to-noise ratio (SNR) and the log spectral distortion (LSD),
that is defined as:
\begin{equation}
\mathrm{LSD}=\frac{1}{L}\sum_{\ell=0}^{L-1}\left[\frac{1}{K}\sum_{k=0}^{K-1}\left(10\log_{10}\frac{\max\left(\left|S(k,\ell)\right|^{2},\epsilon(k)\right)}{\max\left(\left|\hat{S}(k,\ell)\right|^{2},\epsilon(k)\right)}\right)^{2}\right]^{\frac{1}{2}}\label{eq:def_LSD}
\end{equation}
where $L$ is the number of frames, $K$ is the number of frequency
bins and $\epsilon(k)$ is meant to prevent extreme values for spectral
regions of low energy. In both cases, the reference signal is estimated
by applying the GSS algorithm on multi-channel recording containing
only the source of interest. Because the reference signal is only
available with reverberation, the reverberation suppression algorithm
is not used in these tests. Only the narrowband part of the signal
(300 Hz to 3400 Hz) is considered since most of the speech information
is contained in that band.

Measurements are made for the following processing:
\begin{itemize}
\item unprocessed microphone inputs;
\item delay-and-sum (fixed) beamformer;
\item separation with GSS only;
\item separation with GSS followed by a conventional single-source post-filter
(removing the interference estimation from our post-filter);
\item separation with GSS followed by the proposed multi-source post-filter.
\end{itemize}
\begin{comment}
Tables \ref{cap:Signal-to-noise-ratio-(SNR)} and \ref{cap:Log-spectral-distortion-(LSD)}
compare the results obtained for different configurations: unprocessed
microphone inputs, delay-and-sum algorithm, GSS algorithm, GSS algorithm
with single-channel post-filter, and GSS algorithm with multi-channel
post-filter (proposed). 
\end{comment}

\begin{figure}[!h]
\begin{center}\subfloat[Signal-to-noise ratio (SNR)]{

\includegraphics[width=0.8\columnwidth,keepaspectratio]{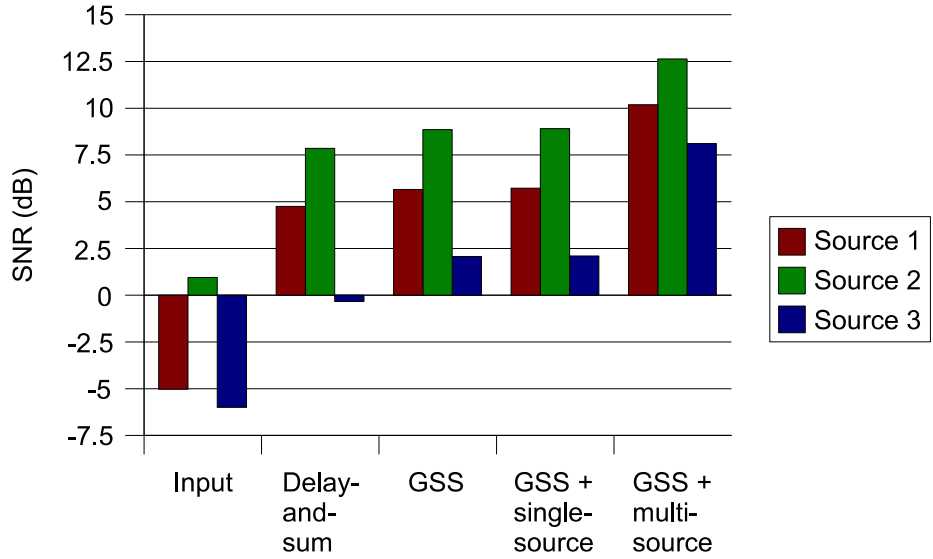}}\end{center}

\begin{center}\subfloat[Log-spectral distance (LSD)]{

\includegraphics[width=0.8\columnwidth,keepaspectratio]{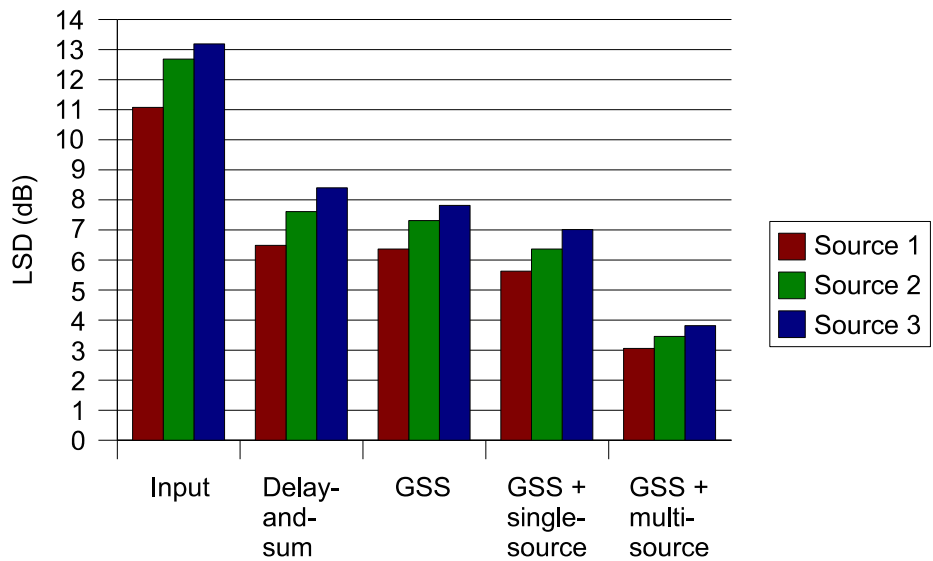}}\end{center}

\caption{Signal-to-noise ratio (SNR) and log-spectral distortion (LSD) for
each source of interest.\label{cap:SNR+LSD}}
\end{figure}

The SNR and LSD results for each source and for each processing is
shown in Figure \ref{cap:SNR+LSD}. These results demonstrate that
when three sources are present, the complete source separation system
we propose provides an average SNR improvement of 13.7 dB and an average
LSD improvement 8.9 dB. It can be observed that both the GSS algorithm
and the post-filter play an important role in the separation and that
our proposed multi-source post-filter performs significantly better
than single-source noise removal. When combined together, the GSS
algorithm and the post-filter require 25\% of a 1.6 GHz Pentium-M
to run in real-time when three sources are present.

\begin{figure}[!h]
\begin{center}\subfloat[Attenuation]{

\includegraphics[width=0.8\columnwidth,keepaspectratio]{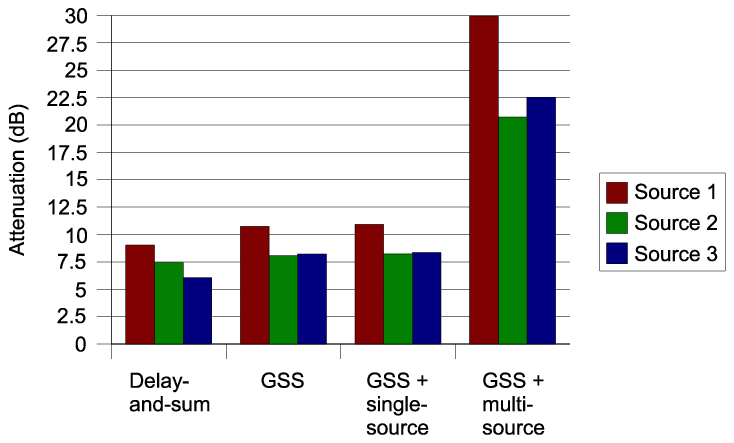}}\end{center}

\caption{Attenuation of noise and interference in the direction of each source
of interest.\label{cap:Attenuation}}
\end{figure}

There are two factors that affect the SNR and LSD measures: the amount
of residual noise and interference, and the distortion caused to the
signal of interest. It is not possible to separate those two factors.
For that reason, we measured the attenuation (relative to the input
signal) of the noise and interference in the direction of each source
when it is silent. Attenuation for the different processing is shown
in Figure \ref{cap:Attenuation} (the input signal has 0 dB attenuation
by definition). The GSS algorithm alone provides an average attenuation
of 9 dB, while our proposed multi-source post-filter further attenuates
the noise and interference by 15.5 dB (compared to GSS alone), for
a total average attenuation of 24.5 dB.

The results in SNR, LSD and attenuation show that the delay-and-sum
algorithm performs reasonably well when compared to the GSS algorithm.
Because our implementation of the GSS is initialised in a manner equivalent
to a delay-and-sum beamformer, this indicates that we can expect the
separation algorithm to work well even when adaptation is incomplete.
Also, it shows that the delay-and-sum beamformer could be used instead
of GSS in a case where very limited computing power is available.

\begin{figure}[!h]
\begin{center}\subfloat[Signal captured by one microphone]{

\includegraphics[width=0.75\columnwidth,height=0.25\paperwidth]{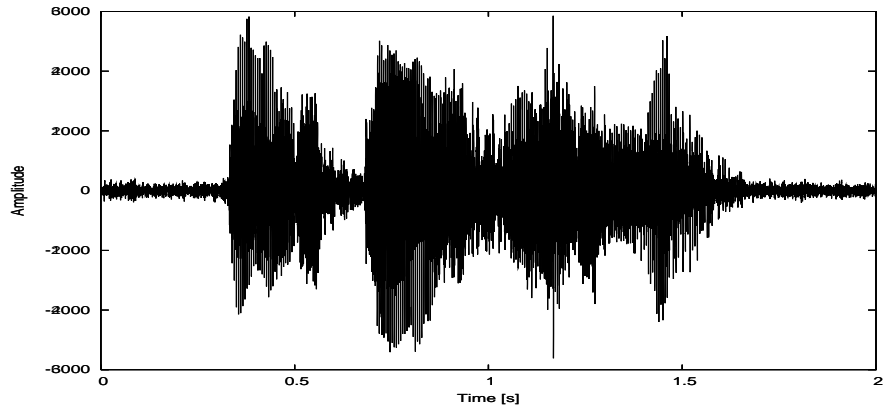}}\end{center}

\begin{center}\subfloat[Separated signal]{

\includegraphics[width=0.75\columnwidth,height=0.25\paperwidth]{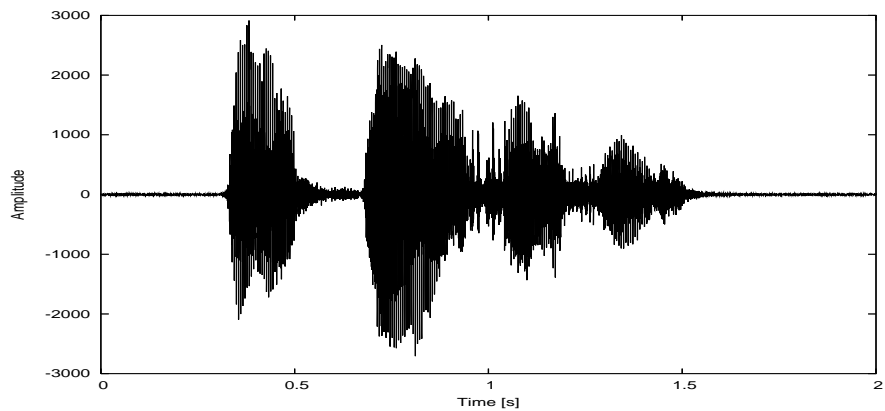}}\end{center}

\begin{center}\subfloat[Reference signal]{

\includegraphics[width=0.75\columnwidth,height=0.25\paperwidth]{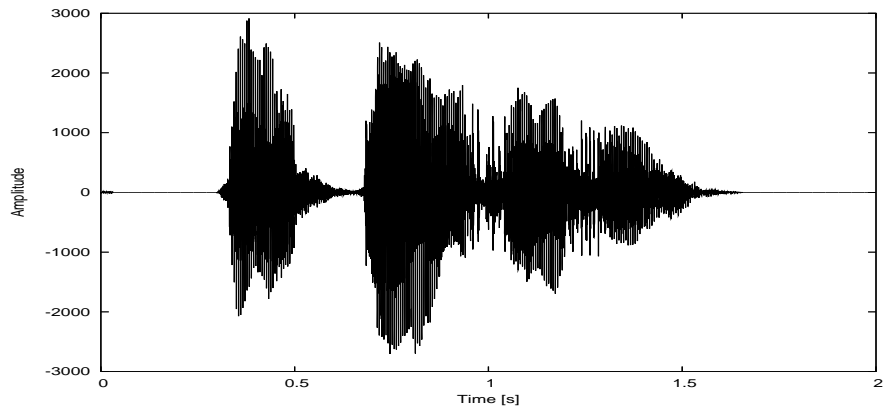}}\end{center}

\caption{Temporal signals for separation of first source.\label{cap:Signal-amplitude}}
\end{figure}

\begin{figure*}[!h]
\includegraphics[width=1\columnwidth,keepaspectratio]{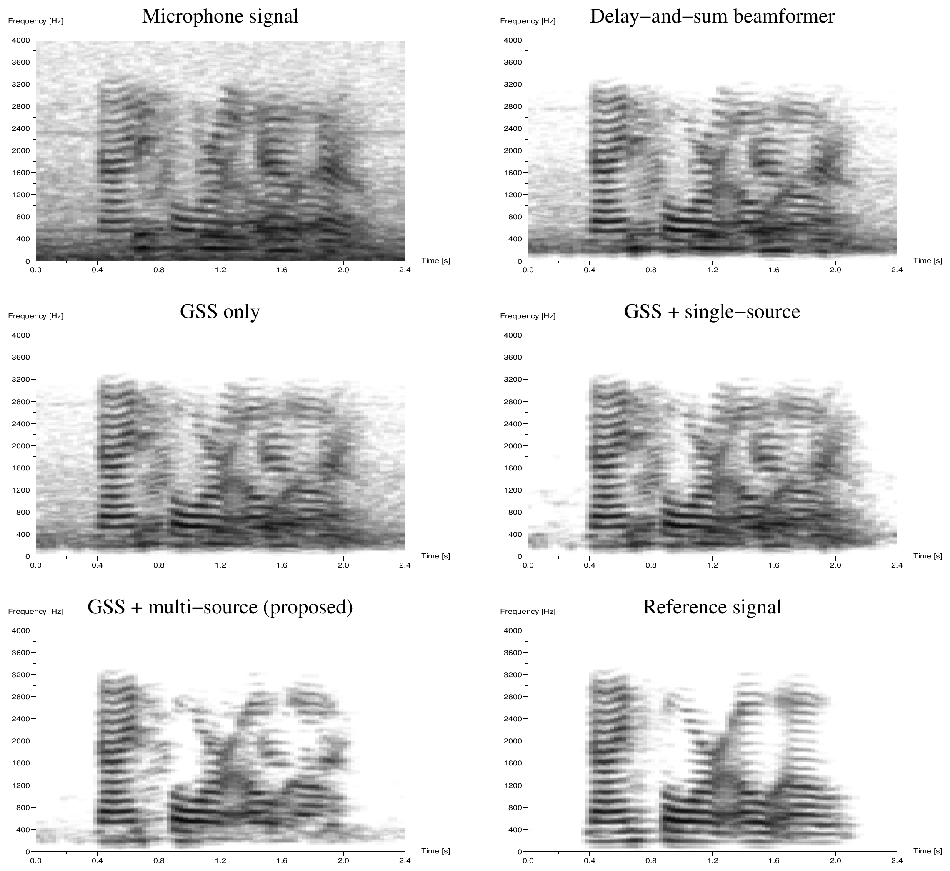}

\caption{Spectrograms for separation of first source comparing different processing.\label{cap:Spectrogram-for-separation}}
\end{figure*}

Time-domain signal plots for the first source are shown in Figure
\ref{cap:Signal-amplitude} and spectrograms are shown in Figure \ref{cap:Spectrogram-for-separation}.
Even though the task involves non-stationary interference with the
same frequency content as the signal of interest, it can be observed
that our proposed post-filter (unlike the single-source post-filter)
is able to remove most of the interference, while not causing excessive
distortion to the signal of interest. Informal subjective evaluation
confirms that the post-filter has a positive impact on both quality
and intelligibility of the speech\footnote{Audio signals and spectrograms for all three sources are available
at: \texttt{http://www.gel.usherb.ca/laborius/projects/Audible/separation/}}.

\section{Discussion\label{sec:Separation-Discussion}}

In this chapter, we described a microphone array linear source separator
and a post-filter in the context of multiple and simultaneous sound
sources. The linear source separator is based on a simplification
of the geometric source separation algorithm that performs regularisation
and instantaneous estimation of the correlation matrix $\mathbf{R}_{\mathbf{xx}}(k)$.
The post-filter is based on an optimal log-spectral MMSE estimator
where the noise estimate is computed as the sum of a stationary noise
estimate, an estimation of leakage from the geometric source separation
algorithm, and a reverberation estimate.

Experimental results show an average reduction in log spectral distortion
of 8.9 dB and an average increase of the signal-to-noise ratio of
13.7 dB compared to the noisy signal inputs. This is significant;y
better than all other methods evaluated. A significant part of that
improvement is due to our multi-source post-filter, which is shown
to perform better than a traditional single-source post-filter. Informal
evaluation and visual inspection of spectrograms indicate that the
distortion introduced by the system is acceptable to most listeners. 

A possible improvement to the algorithm would be to derive a method
that computes reverberation parameters so that the system can automatically
adapt to changing environments. It would also be interesting to explore
the use of the signal phase as an additional source of information
to the post-filter, as proposed by Aarabi and Shi \cite{AARABI2004}.
Finally, the multi-source post-filter we have developed makes few
assumptions about the nature of the signals or the linear separation
algorithm used. It may thus be applicable to other multi-sensor systems.

\chapter{Speech Recognition of Separated Sound Sources\label{cha:Speech-Recognition-Integration}}

Robust speech recognition usually assumes source separation and/or
noise removal from the audio or feature vectors. When several people
speak at the same time, each separated speech signal is severely distorted
in spectrum from its original signal. 

We propose two different approaches for speech recognition on a mobile
robot. Both approaches make use of the separated sources computed
by the algorithm presented in Chapter \ref{cha:Sound-Source-Separation}.
The first is a conventional approach and consists of sending the separated
audio directly to a speech recognition engine for further processing.
This is the most common method for speech recognition using a microphone
array \cite{asano-goto-itou-asoh2001} and, in some variants, the
acoustic models are adapted to the separated speech \cite{Nakadai02-IROS}.

The second approach provides closer integration between the sound
source separation algorithm and the speech recognition engine by estimating
the reliability of spectral features, and providing that information
to the speech recognition engine so it can be used during recognition
\cite{Renevey01,Barker00}. This second approach, as originally described
in \cite{YamamotoValin2005}, was developed in collaboration with
Kyoto University and builds on an original proof-of-concept by Yamamoto
\cite{Yamamoto04a,Yamamoto04b}. Unlike the original proof-of-concept,
which required access to the clean speech data, the system we now
propose can be used in a real environment by computing the missing
feature mask only from the data available to the robot. This is done
by using interference information provided by the multi-source post-filter
described in Section \ref{sec:Post-filter}.

This chapter is organised as follows. Section \ref{sec:Related-Work-ASR}
discusses the state of the art and limitations of speech enhancement
and missing feature-based speech recognition. Section \ref{sec:ASR-overview}
gives an overview of the system. Speech recognition integration and
computation of the missing feature mask are addressed in Section \ref{sec:Integration-With-ASR}.
Results are presented in Section \ref{sec:ASR-Results}, followed
by a discussion.

\section{Related Work in Robust Speech Recognition\label{sec:Related-Work-ASR}}

Robustness against noise in conventional\footnote{We use conventional in the sense of speech recognition for applications
where a single microphone is used in a static environment such as
a vehicle or an office.} automatic speech recognition (ASR) is being extensively studied,
particularly in the AURORA project \cite{Pearce01}. In order to realise
noise-robust speech recognition, \emph{multi-condition training} (training
on a mixture of clean speech and noises) has been studied \cite{Lippmann87,Blanchet92}.
This is currently the most common method for vehicle and telephone
applications. Because an acoustic model obtained by multi-condition
training reflects the noise in a specific condition, ASR's use of
the acoustic model is effective as long as the noise is stationary.
This assumption holds for background noise in a vehicle and on a telephone.
However, multi-condition training may not be effective for mobile
robots, since those usually work in dynamically changing noisy environments.
Source separation and speech enhancement algorithms for robust recognition
is another potential alternative for ASR on mobile robots. However,
this method is not always effective, since most source separation
and speech enhancement techniques add some distortion to the separated
signals and consequently degrade features, reducing the recognition
rate, even if the signal is perceived to be cleaner by na\"{\i}ve
listeners \cite{OShaughnessy2003}. The work of Seltzer \emph{et al.}
\cite{seltzer2003} on microphone arrays addresses the problem of
optimising the array processing specifically for speech recognition.

Although the source separation and speech enhancement system in Chapter
\ref{cha:Sound-Source-Separation} are designed for optimal speech
recognition, it is desirable to make speech recognition more robust
to uncertain or distorted features.

\subsection{Missing Feature Theory Overview}

Research of confidence islands in the time-frequency plane representation
has been shown to be effective in various applications and can be
implemented with different strategies. One of the most effective is
the missing feature strategy. Cooke \textit{et al.} \cite{cooke1994,cooke2001b}
propose a probabilistic estimation of a mask in regions of the time-frequency
plane where the information is not reliable. Then, after masking,
the parameters for speech recognition are generated and can be used
in conventional speech recognition systems. Using this method, it
is possible to obtain a significant increase in recognition rates
without any modelling of the noise \cite{barker2001}. In this scheme,
the mask is essentially based on the signal-to-interference ratio
(SIR) and a probabilistic estimation of the mask is used.

A missing feature theory-based ASR uses a Hidden Markov Model (HMM)
where acoustic model probabilities are modified to take into account
only the reliable features. According to the work by Cooke \textit{et
al.} \cite{cooke2001b}, HMMs are trained on clean data. Density in
each state $S$ is modelled using mixtures of $M_{g}$ Gaussians with
diagonal-only covariance.

Let $f(\mathbf{x}|S)$ be the output probability density of feature
vector $\mathbf{x}$ in state $S$, and $P(j|S)$ represent the weight
of mixture $j$ expressed as a probability. The output probability
density is given by: 
\begin{equation}
f(\mathbf{x}|S)=\sum_{j=1}^{M_{g}}P(j|S)f(\mathbf{x}|j,S)\label{eq:missingFeatre}
\end{equation}

Cooke \textit{et al.} \cite{cooke2001b} propose to transform Equation
\ref{eq:missingFeatre} to take into consideration the only reliable
features from $\mathbf{x}$ and to remove unreliable features. This
is equivalent to use the marginalisation probability density functions
$f(x_{r}|j,S)$ instead of $f(\mathbf{x}|j,S)$ by simply implementing
a binary mask. Consequently, only reliable features are used in the
probability calculation, and the recogniser can avoid undesirable
effects due to unreliable features.

Conventional ASR usually uses Mel Frequency Cepstral Coefficients
(MFCC) \cite{Picone} that capture the characteristics of speech.
However, the missing feature mask is usually computed in the spectral
domain and it is not easy to convert to the cepstral domain. Automatic
generation of missing feature mask needs prior information about which
spectral regions of a separated sound are distorted. This information
can be obtained by our sound source separation and post-filter system
\cite{ValinIROS2004,ValinICASSP2004}. We use the post-filter gains
to automatically generate the missing feature mask. Since we use a
vector of 48 spectral features, the missing feature mask is a vector
comprising the 48 corresponding values. The value may be discrete
(1 for reliable, or 0 for unreliable) or continuous between 0 and
1.

\subsection{Applications of Missing Feature Theory}

Hugo Van hamme \cite{vanhamme2003} formulates the missing feature
approach for speech recognisers using conventional parameters such
as MFCC. He uses data imputation according to Cooke \cite{cooke2001b}
and proposes a suitable transformation to be used with MFCC for missing
features. The acoustic model evaluation of the unreliable features
is modified to express that their clean values are unknown or confined
within bounds. In a more recent paper, Hugo Van hamme \cite{vanhamme2004}
presents speech recognition results by integrating harmonicity in
the signal to noise ratio for noise estimation. He uses only static
MFCC as, according to him, dynamic MFCC do not increase sufficiently
the speech recognition rate when used in the context of missing features
framework. The needs to estimate pitch and voiced regions in the time-space
representation is a limit to this approach. In a similar approach,
Raj, Seltzer and Stern \cite{raj2004} propose to modify the spectral
representation to derive cepstral vectors. They present two missing
feature algorithms that reconstruct spectrograms from incomplete noisy
spectral representations (\textit{masked} representations). Cepstral
vectors can be derived from the reconstructed spectrograms for missing
feature recognition. Seltzer \textit{et al.} \cite{seltzer2004SpeechComm}
propose the use of a Bayesian classifier to determine the reliability
of spectrographic elements. Ming, Jancovic and Smith \cite{ming2002,ming2003}
propose the \textit{probabilistic union model} as an alternative to
the missing feature framework. According to the authors, methods based
on the missing feature framework usually require the identification
of the noisy bands. This identification can be difficult for noise
with unknown, time-varying spectral characteristics. They designed
an approach for speech recognition involving partial, unknown corrupted
frequency-bands. In their approach, they combine the local frequency-band
information based on the union of random events, to reduce the dependence
of the model on information about the noise. Cho and Oh \cite{cho2004}
apply the \textit{union model} to improve robust speech recognition
based on frequency bands selection. From this selection, they generate
``channel-attentive'' Mel frequency cepstral coefficients. Even
if the use of missing features for robust recognition is relatively
recent, many applications have already been designed.

In order to avoid the use of multi-condition training, we propose
to merge a multi-microphone source separation and speech enhancement
system with the missing feature approach. Very little work has been
done with arrays of microphones in the context of missing feature
theory. To our knowledge, only McCowan \emph{et al.} \cite{mccowan-rr-02-09-proc}
applies the missing feature framework to microphone arrays. The proposed
approach defines a missing feature mask based on the input-to-output
ratio of a post-filter. The approach is however only validated on
stationary noise. 

\begin{comment}
From our knowledge, only Seltzer et al. \cite{seltzer2003} apply
a subband approach to an array of microphones. They incorporate the
recogniser into the filter optimisation to make sure that signal features
important for recognition are emphasised. Most missing feature based
speech recognisers use only one input channel, sometimes two at most.
This is a strong limit to the segregation of simultaneous sound sources.
With an array of microphones, it is usually possible to segregate
sources and enhance speech signals with fewer distortion, yielding
higher recognition accuracy.
\end{comment}

\section{ASR on Separated Sources\label{sec:ASR-overview}}

\begin{figure}[th]
\begin{center}\includegraphics[width=0.65\paperwidth,keepaspectratio]{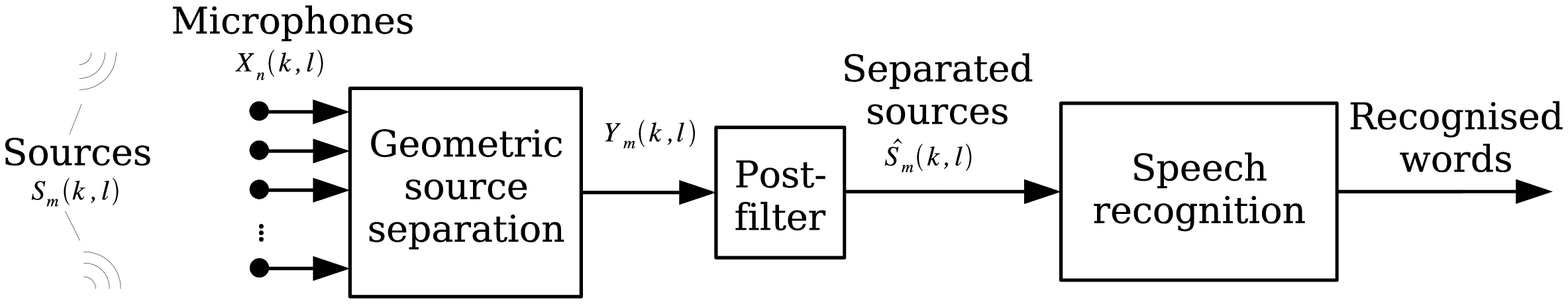}\end{center}

\caption{Direct integration of speech recognition.\label{cap:Speech-recognition-Direct-overview}}
\end{figure}

In the first proposed approach, the sources separated using the algorithm
described in Chapter \ref{cha:Sound-Source-Separation} are sent directly
to a speech recognition engine, as shown in Figure \ref{cap:Speech-recognition-Direct-overview}. 

\begin{figure}[th]
\begin{center}\includegraphics[width=0.65\paperwidth,keepaspectratio]{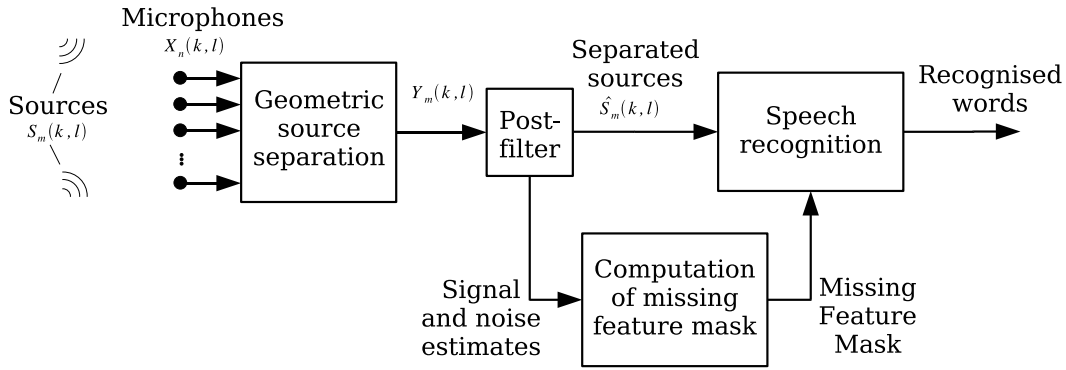}\end{center}

\caption{Speech recognition integration using missing feature theory\label{cap:Speech-recognition-MFT-overview}}
\end{figure}

The second approach, illustrated in Figure \ref{cap:Speech-recognition-MFT-overview},
aims to integrate the different steps of source separation, speech
enhancement and speech recognition as closely as possible in order
to maximise recognition accuracy by using as much of the available
information as possible. The missing feature mask is generated in
the time-frequency plane since the separation module and the post-filter
already use this signal representation. 

Not only can the multi-source post-filter reduce the amount of noise
and interference, but its behaviour provides useful information that
can be used to evaluate the reliability of different regions of the
time-frequency plane for the separated signals. That information is
used to compute a missing feature mask, which the ASR can use to further
improve accuracy. This also has the advantage that acoustic models
trained on clean data can be used (no multi-condition training is
required).

Some missing feature mask techniques can also require the estimation
of prior characteristics of the corrupting sources or noise. They
usually assume that the noise or interference characteristics vary
slowly with time. This is not possible in the context of a mobile
robot. We propose to estimate quasi-instantaneously the mask (without
preliminary training) by exploiting the post-filter outputs along
with the local gains (in the time-frequency plane representation)
of the post-filter. These local gains are used to generate the missing
feature mask. Thus, the ASR with clean acoustic models can adapt to
the distorted sounds by consulting the post-filter feature missing
masks. This approach is also a partial solution to the automatic generation
of simultaneous missing feature masks (one for each speaker). It allows
the use of simultaneous speech recognisers (one for each separated
sound source) with their own mask.

\section{Missing Feature Recognition\label{sec:Integration-With-ASR}}

The post-filter uses adaptive spectral estimation of background noise,
interfering sources and reverberation to enhance the signal produced
during the initial separation. The main idea lies in the fact that,
for each source of interest, the noise estimate is decomposed into
stationary and transient components assumed to be due to leakage between
the output channels of the initial separation stage. It also provides
useful information concerning the amount of noise present at a certain
time, at a particular frequency. Hence, we use the post-filter to
estimate a missing feature mask that indicates how reliable each region
of the spectrum is when performing recognition.

\subsection{Computation of Missing Feature Masks\label{sec:Computation-of-Mask}}

The missing feature mask is a matrix representing the reliability
of each feature in the time-frequency plane. More specifically, this
reliability is computed for each frame and for each Mel-frequency
band. This reliability can be either a continuous value from 0 to
1, or a discrete value of 0 or 1. In this work, discrete masks are
used.

It is worth mentioning that computing the mask in the Mel-frequency
bank domain means that it is not possible to use MFCC features, since
the effect of the Discrete Cosine Transform (DCT) cannot be applied
to the missing feature mask. For this reason, the log-energy of the
Mel-frequency bands are used directly as features.

Each Mel-frequency band is considered reliable if the ratio of the
output energy over the input energy is greater than a threshold $T_{m}$.
The reason for this choice is that it is assumed that the more noise
present in a certain frequency band, the lower the post-filter gain
will be for that band.

We proceed in two steps. For each frame $\ell$ and for each MEL frequency
band $i$, do: 
\begin{enumerate}
\item Compute a continuous mask $m_{\ell}(i)$ that reflects the reliability
of the band: 
\begin{equation}
m_{\ell}(i)=\frac{S_{\ell}^{out}(i)+N_{\ell}(i)}{S_{\ell}^{in}(i)}\label{eq:cont_mask}
\end{equation}

where $S_{\ell}^{in}(i)$ and $S_{\ell}^{out}(i)$ are respectively
the post-filter input and output energy for frame $\ell$ at Mel-frequency
band $i$, and $N_{\ell}(i)$ is the background noise estimate (summation
of the $\lambda_{m}^{stat.}(k,\ell)$ for band $i$).

\item Deduce a binary mask $M_{\ell}(i)$ used to remove the unreliable
MEL frequency bands at frame $\ell$:

\begin{equation}
M_{\ell}(i)=\left\{ \begin{array}{ll}
1, & m_{\ell}(i)>T_{m}\\
0,\quad & \mathrm{otherwise}
\end{array}\right.\label{eq:mask_threshold}
\end{equation}
 where $T_{m}$ is an arbitrary threshold (we use $T_{m}=0.25$ based
on speech recognition experiments).

\end{enumerate}
In comparison to McCowan \emph{et al.} \cite{mccowan-rr-02-09-proc},
the use of the multi-source post-filter allows a better reliability
estimation by distinguishing between interference and background noise.
We include the background noise estimate $N_{\ell}(i)$ in the numerator
of Equation~\ref{eq:cont_mask} to ensure that the missing feature
mask equals 1 when no speech source is present (as long as there is
no interference). Without this condition, all of the features would
be considered unreliable in the case where no speech is present for
the source of interest. This would, in turn, prevent the silence model
of the ASR from performing adequately (this has been observed in practice).
Using a more conventional post-filter as proposed by McCowan \emph{et
al.} \cite{mccowan-rr-02-09-proc} and Cohen \emph{et al.} \cite{CohenArray2002}
would not allow the mask to preserve silence features, which is known
to degrade ASR accuracy. The distinction between background noise
and interference also reflects the fact that background noise cancellation
is generally much better than interference cancellation. 

An example of missing feature masks with the corresponding separated
signals after post-filtering is shown in Figure~\ref{cap:Spectrograms-with-MFM}.
It is observed that the mask indeed preserves the silent periods and
considers unreliable regions of the spectrum as dominated by other
sources. The missing feature mask for delta-features (time derivative
of the Mel spectral features) is computed using the mask for the static
features. The dynamic mask $\Delta M_{\ell}(i)$ is computed as:
\begin{equation}
\Delta M_{\ell}(i)=\prod_{k=-2}^{2}M_{\ell-k}(i)\label{eq:delta_mask}
\end{equation}
 and is non-zero only when all the MEL features used to compute the
delta-features are deemed reliable.

\subsection{Speech Analysis for Missing Feature ASR}

Since MFCC cannot be easily used directly with a missing feature mask
and as the post-filter gains are expressed in the time--frequency
plane, we use spectral features that are derived from MFCC features
with the Inverse Discrete Cosine Transform (IDCT). The detailed steps
for feature generation are as follows:
\begin{enumerate}
\item {[}FFT{]} The speech signal sampled at 16 kHz is analysed using an
FFT with a 400-point window and a 160 frame shift. 
\item {[}Mel{]} The spectrum is analysed by a Mel-scale filter bank to obtain
the Mel-scale spectrum of the $24^{\mathrm{th}}$ order. 
\item {[}Log{]} The Mel-scale spectrum of the $24^{\mathrm{th}}$ order
is converted to log-energies. 
\item {[}DCT{]} The log Mel-scale spectrum is converted by Discrete Cosine
Transform to the cepstrum. 
\item {[}Lifter{]} Cepstral features 0 and 13-23 are set to zero so as to
make the spectrum smoother. 
\item {[}CMS{]} Convolutive effects are removed using Cepstral Mean Subtraction. 
\item {[}IDCT{]} The normalised cepstrum is transformed back to the log
Mel-scale spectral domain by means of an Inverse DCT. 
\item {[}Differentiation{]} The features are differentiated in the time
domain. Thus, we obtain 24 log spectral features as well as their
first-order time derivatives. 
\end{enumerate}
The {[}CMS{]} step is necessary in order to remove the effect of convolutive
noise, such as reverberation and microphone frequency response.

The same features are used for training and evaluation. Training is
performed on clean speech, without any effect from the post-filter.
In practice, this means that the acoustic model does not need to be
adapted in any way to our method and the only difference with a conventional
ASR is the use of the missing feature mask as represented in Equation~\ref{eq:missingFeatre}.

\subsection{Automatic Speech Recognition Using Missing Feature Theory }

Once the ASR features and the missing feature mask are computed, the
last step consists of making use of the missing feature mask in the
probability model of the ASR. Let $f(x|s)$ be the output probability
density of feature vector $x$ in state $S$. The output probability
density is defined by Equation~\ref{eq:missingFeatre} and becomes:
\begin{equation}
f(\mathbf{x}|S)=\sum_{j=1}^{M_{g}}P(j|S)f(\mathbf{x}_{r}|j,S)\label{eq:MFT-marginalisation}
\end{equation}
 where $\mathbf{x}_{r}=\{x_{i}|M(i)=1\}$ contains only the reliable
features of $\mathbf{x}$. This means that only reliable features
are used in probability calculation, and thus the recogniser can avoid
undesirable effects due to unreliable features.

\section{Results\label{sec:ASR-Results}}

In this section, we present speech recognition results for the two
coupling methods: direct, and using missing feature theory. The testing
conditions include recognition of two and three stationary simultaneous
speakers, as well as recognition on two moving speakers.

\subsection{Direct Coupling\label{sub:ASR-Results-Direct-Integration}}

For the case where speech recognition is performed directly on the
separated speech, we use the same setup as described in Section \ref{sec:Experimental-Setup}.
We use the Nuance\footnote{\texttt{http://www.nuance.com/}} speaker-independent
commercial speech recognition engine, which only accepts speech sampled
at 8 kHz. The test data is composed of 251 utterances of four connected
English digits (male and female) selected from the AURORA clean database\footnote{Files were taken from the \texttt{testa/clean1/} to \texttt{testa/clean4/}
directories.}. In each test, the utterances are played from two or three speakers
at the same time. In the condition where three speakers are used,
one speaker is placed at 90 degrees on the left, one is in front of
the robot, and the other is placed 90 degrees to the right. In the
case of two speakers, only the speakers in front and on the right
are used.

Three processing conditions are compared:
\begin{enumerate}
\item Use of the GSS algorithm only with no post-filtering;
\item Use of the GSS with post-filter, but without enabling reverberation
cancellation (no dereverb., Equation \ref{eq:noise_estim});
\item The proposed system with GSS, post-filtering including reverberation
cancellation (Equation \ref{eq:noise_estim_reverb}).
\end{enumerate}
No tests are presented using only one microphone because in this case,
the system would have no directional information at all to distinguish
between the speakers (it would recognise the same digits for all sources). 

\begin{figure}[th]
\includegraphics[width=0.5\columnwidth,keepaspectratio]{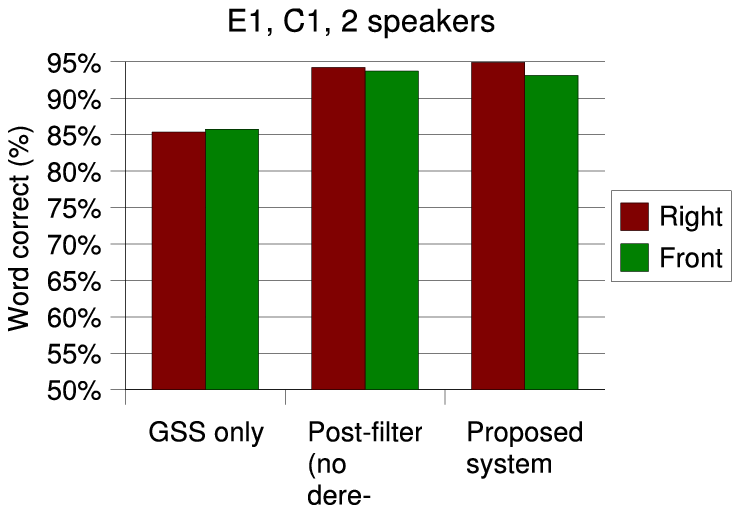}\includegraphics[width=0.5\columnwidth,keepaspectratio]{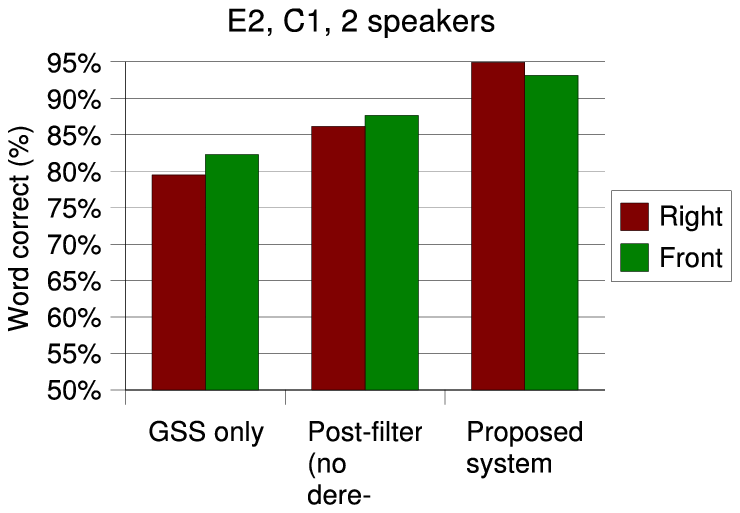}

\includegraphics[width=0.5\columnwidth,keepaspectratio]{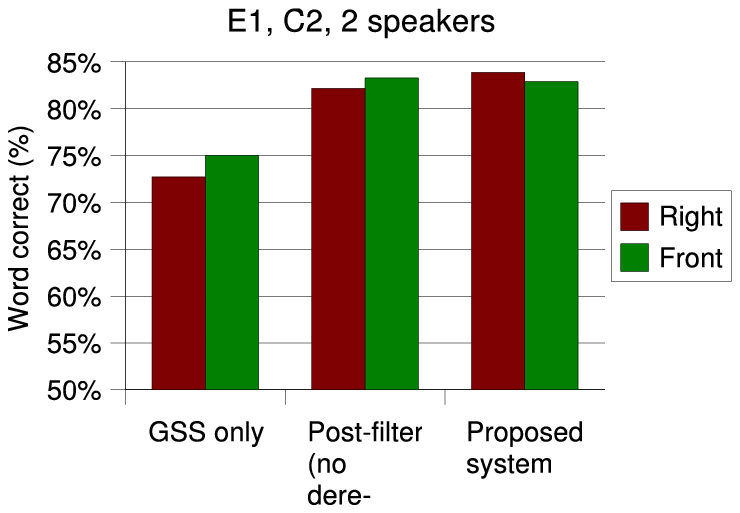}\includegraphics[width=0.5\columnwidth,keepaspectratio]{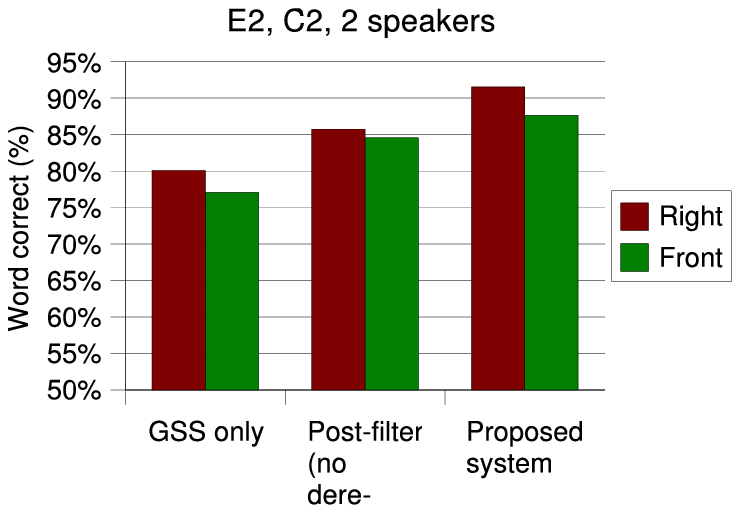}

\caption{Speech recognition results for two simultaneous speakers.\label{cap:nuance_2speakers}}
\end{figure}

\begin{figure}[th]
\includegraphics[width=0.5\columnwidth,keepaspectratio]{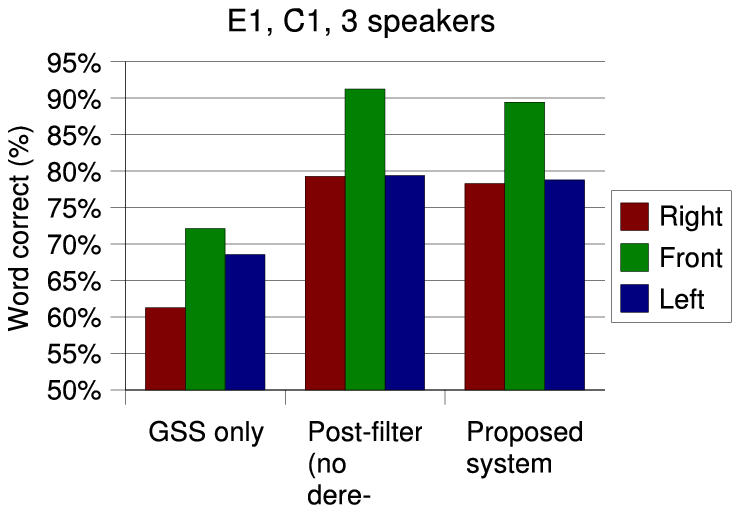}\includegraphics[width=0.5\columnwidth,keepaspectratio]{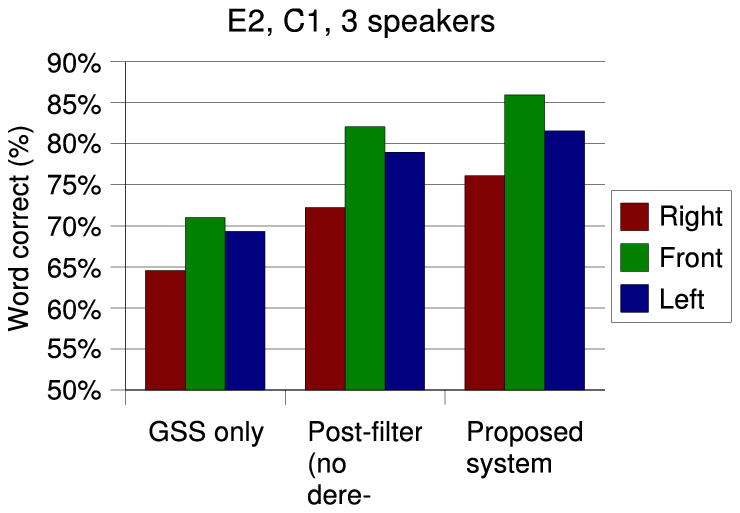}

\includegraphics[width=0.5\columnwidth,keepaspectratio]{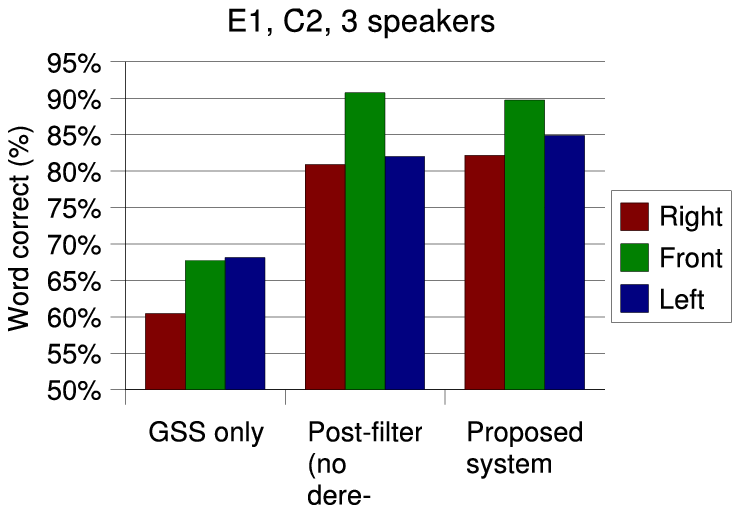}\includegraphics[width=0.5\columnwidth,keepaspectratio]{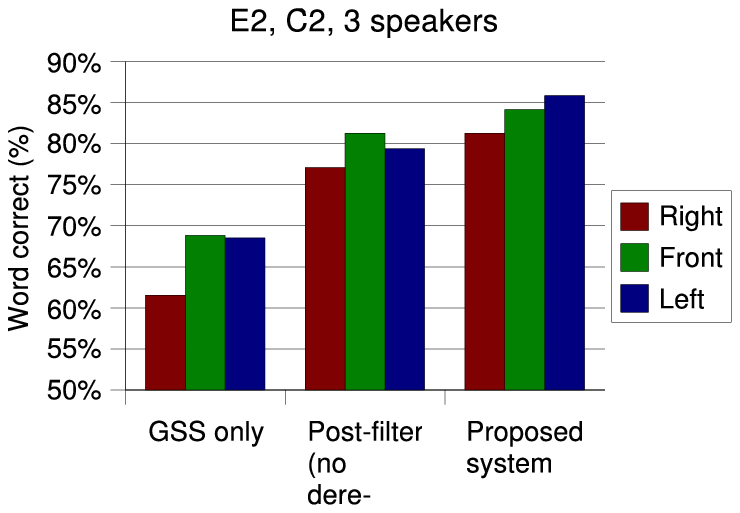}

\caption{Speech recognition results for three simultaneous speakers.\label{cap:nuance_3speakers}}
\end{figure}

Results for two simultaneous speakers are shown in Figure \ref{cap:nuance_2speakers},
while results for three simultaneous speakers are shown in Figure
\ref{cap:nuance_3speakers}. The average recognition rate (word correct
averaged over all sources) is 83\% for three simultaneous speakers
and 90\% for two simultaneous speakers. In all cases, the proposed
system provides a significant improvement over the use of the GSS
algorithm alone. The average reduction in word error rate is 51\%
relative\footnote{The relative improvement is computed as the difference in error rates,
divided by the error rate for the reference condition.} and is fairly constant across microphone configuration, environment
(reverberation conditions) and number of speakers. The results with
no reverberation cancellation show that reverberation cancellation
is responsible for half of the post-filter improvement in the E2 environment
(1 second reverberation time), but has no significant effect on the
E1 environment (350 ms reverberation time), even though individual
results vary slightly.

\begin{comment}
\begin{table}[th]
\begin{tabular}{|c|c|c|c|c|c|c|}
\hline 
 & \multicolumn{2}{c|}{right} & \multicolumn{2}{c|}{centre} & \multicolumn{2}{c|}{left}\tabularnewline
\hline 
 & GSS & PF & GSS & PF & GSS & PF\tabularnewline
\hline 
\hline 
E1, C1, 2 speakers &  & 95.1 &  & 93.0 &  & \tabularnewline
\hline 
E1, C1, 3 speakers &  & 78.3 &  & 86.6 &  & 79.1\tabularnewline
\hline 
E1, C2, 2 speakers &  & 82.3 &  & 81.2 &  & \tabularnewline
\hline 
E1, C2, 3 speakers &  & 80.4 &  & 87.3 &  & 83.4\tabularnewline
\hline 
E2, C1, 2 speakers &  & 89.9 &  & 91.6 &  & \tabularnewline
\hline 
E2, C1, 3 speakers &  & 74.0 &  & 84.1 &  & 81.3\tabularnewline
\hline 
E2, C2, 2 speakers &  & 90.5 &  & 89.1 &  & \tabularnewline
\hline 
E2, C2, 3 speakers &  & 81.3 &  & 82.2 &  & 82.9\tabularnewline
\hline 
\end{tabular}

\caption{Speech recognition results using direct ASR integration.}
\end{table}
\end{comment}

\subsubsection{Human Capabilities}

Just for the sake of comparison, we compare the auditory capabilities
developed to human capabilities in a \emph{cocktail party} context.
This is done by repeating the previous experiment with a person in
the place of the robot. Five different listeners placed in the E1
environment were told to transcribe the digits from the front speaker
only. We did not ask the listeners to transcribe all directions because
humans can usually only focus their attention on a single source of
interest. The use of connected digits for the test is convenient because
for this, humans cannot make use of higher level language knowledge.
Moreover, the listeners were all non-native English speakers. It is
shown in \cite{Peters1999} that in noisy conditions, non-native listeners
tend to obtain recognition accuracy results similar to those obtained
by an ASR. This indicates that our tests would be comparing mainly
the lower-level auditory capabilities (audio analysis) of humans to
our robot. 

\begin{figure}[th]
\begin{center}\includegraphics[width=0.7\columnwidth,keepaspectratio]{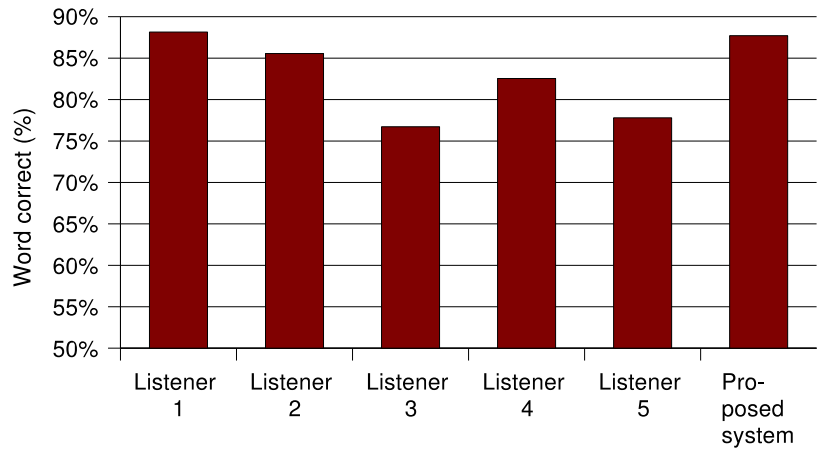}\end{center}

\caption{Human versus machine speech recognition accuracy.\label{cap:Human-vs.-machine}}
\end{figure}

The results in Figure \ref{cap:Human-vs.-machine} comparing human
accuracy show an important variability between speakers. Those results
indicate that in difficult \emph{cocktail party} conditions, our system
performs equally or better than humans. In fact, the best listener
obtained only 0.5\% (absolute) better accuracy, while the worst listener
had 11\% lower accuracy (absolute). We do not however expect these
results to hold for the case of real conversations, since in this
case, humans have an advantage over an ASR's language model (we are
good at ``guessing'' what words will come next in a conversation).
On the other hand, our system is able to listen to several conversations
simultaneously, while our listeners could not recognise much of what
originated from the other sources.

\subsubsection{Moving Sources}

\begin{figure}[th]
\begin{center}\includegraphics[width=0.55\columnwidth,keepaspectratio,angle=270]{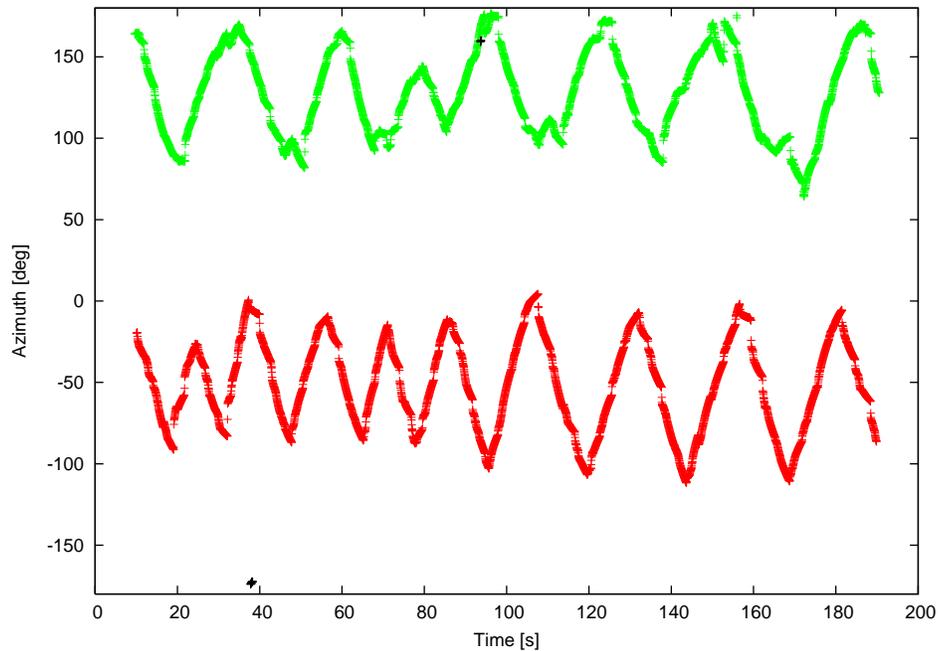}\end{center}

\caption[Source trajectories (azimuth as a function of time): recognition on two moving speakers.]{Source trajectories (azimuth as a function of time): recognition on two moving speakers. Two false detections are shown in black (around $t=40\:\mathrm{s}$ and $t=100\:\mathrm{s}$).}\label{cap:Source-trajectories:-recognition}

\begin{comment}
Source trajectories (azimuth as a function of time): recognition on
two moving speakers. Two false detections are shown in black (around
$t=40\:s$ and $t=100\:s$).\label{cap:Source-trajectories:-recognition}
\end{comment}
\end{figure}

We performed an experiment to measure speech recognition accuracy
when speakers are moving. Unfortunately, the fact that the AURORA
data used in the previous experiment is only available at 8 kHz makes
accurate tracking impossible because most of the speech bandwidth
is not available (see Section \ref{sub:Audio-Bandwidth}). For that
reason, the digit strings were spoken out loud in the E1 environment,
while wandering around the robot at a distance varying between one
and two metres. The estimated trajectories of both speakers are shown
in Figure \ref{cap:Source-trajectories:-recognition}. Over a period
of three minutes, only two false detections were found, one of which
can be explained as being the sound of a speaker's feet. The speech
recognition accuracy (word correct) is 96.0\% for the first speaker
(trajectory shown in green) and 85.1\% for the second speaker (trajectory
shown in red). The difference between the two results can be explained
by the fact that the first speaker's voice is louder.

\subsection{Missing Feature Theory Coupling\label{sub:ASR-Results-MFT}}

\begin{figure}[th]
\begin{centering}
\begin{center}\includegraphics[height=0.3\paperwidth]{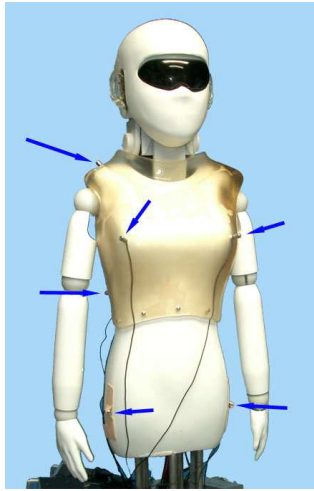}\end{center}
\par\end{centering}

\caption{SIG 2 robot with eight microphones (two are occluded).\label{cap:SIG-2-array}}
\end{figure}

Unlike other experiments, the work on speech recognition integration
using missing feature theory is evaluated on the SIG2 humanoid robot
on which eight microphones were installed, as shown in Figure \ref{cap:SIG-2-array}.
These results were obtained during a collaboration with Kyoto University
and, while they reflect a slightly older version of the separation
algorithm, we believe they are nonetheless valid and interesting. 

Because the ASR needs to be modified to support MFT, it was not possible
to use Nuance as in Section \ref{sub:ASR-Results-Direct-Integration}
because it is not distributed with the source code. The speech recognition
engine used is based on the CASA Tool Kit (CTK) \cite{barker2001}
hosted at Sheffield University, U.K.\footnote{http://www.dcs.shef.ac.uk/research/groups/spandh/projects/respite/ctk/},
and uses 16 kHz audio as input. The same work was also applied to
the Julius \cite{Lee2001} Japanese ASR\footnote{http://julius.sourceforge.jp/},
but since preliminary experiments comparing both recognisers showed
better recognition accuracy using CTK, only these results are reported. 

In order to test the system, three Japanese voices (two males, one
female) are played simultaneously: one in front, one on the left,
and one on the right. In three different experiments, the angle between
the centre speaker and the side speakers are set to 30, 60, and 90
degrees. The speakers are placed two meters away from the robot. The
room in which the experiment took place has a reverberation time of
approximately 300 ms. The post-filter uses short-term spectral amplitude
(STSA) estimation \cite{EphraimMalah1984} since it was found to maximise
speech recognition accuracy. Speech recognition complexity is not
reported as it usually varies greatly between different engine and
settings. Japanese isolated word recognition is performed using speaker-independent
triphone \cite{triphones} acoustic models.

\subsubsection{Separated Signals}

\begin{figure*}[th]
\begin{center}\includegraphics[width=0.65\paperwidth,keepaspectratio]{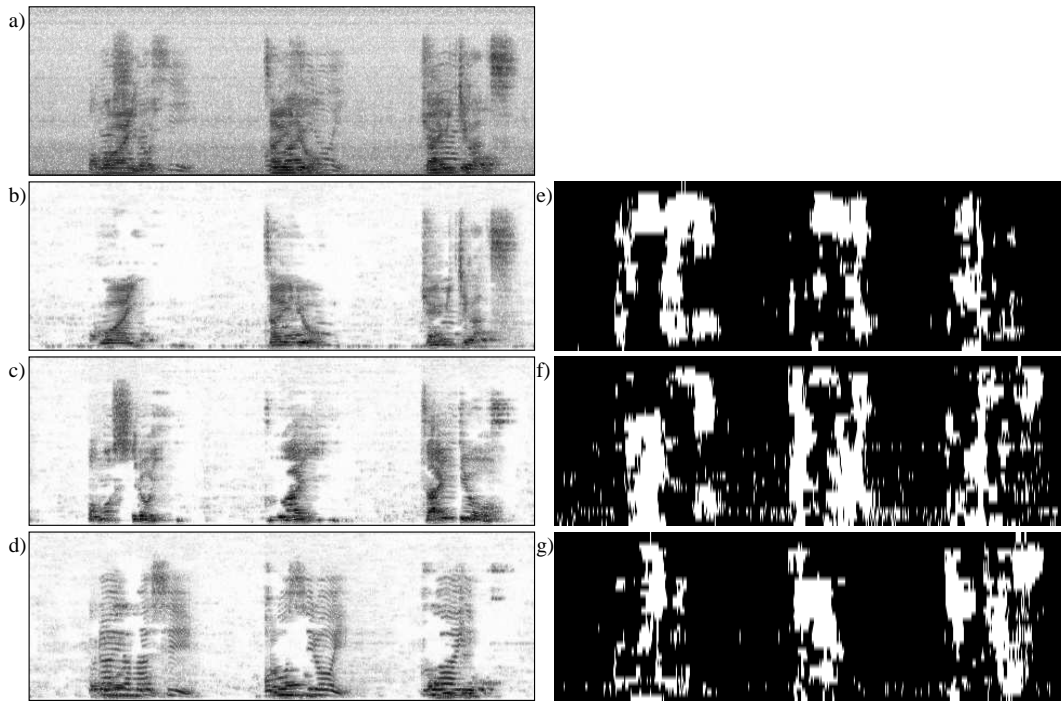}\end{center}

\caption[Missing feature masks for separation of three speakers.]{Missing feature masks for separation of three speakers, $90^{\circ}$ apart with post-filter: a) signal as captured at microphone \#1; b) separated right speaker; c) separated centre speaker; d) separated left speaker; e)\ --\ g) corresponding mel-frequency missing feature mask for static features with reliable features ($M_{\ell}(i)=1$) shown in black. Time is represented on the x-axis and frequency (0-8kHz) on the y-axis.}\label{cap:Spectrograms-with-MFM}

\begin{comment}
Spectrograms for separation of three speakers, $90^{\circ}$ apart
with post-filter: a) signal as captured at microphone \#1; b) separated
right speaker; c) separated centre speaker; d) separated left speaker;
e)~-~g) corresponding mel-frequency missing feature mask for static
features with reliable features ($M_{\ell}(i)=1$) shown in black.
Time is represented on the \emph{x-axis} and frequency (0-8kHz) on
the \emph{y-axis}.
\end{comment}
\end{figure*}

Spectrograms showing separation of the three speakers\footnote{Audio signals and spectrograms for all three sources are available
at: \texttt{http://www.gel.usherb.ca/laborius/projects/Audible/separation/sap/}} are shown in Figure \ref{cap:Spectrograms-with-MFM}, along with
the corresponding mask for static features. Even though the task involves
non-stationary interference with the same frequency content as the
signal of interest, we observe that our post-filter is able to remove
most of the interference. Informal subjective evaluation has confirmed
that the post-filter has a positive impact on both quality and intelligibility
of the speech. This is also confirmed by improved recognition results.

\subsubsection{Speech Recognition Accuracy}

We present speech recognition accuracy results obtained in three different
conditions:
\begin{enumerate}
\item Geometric Source Separation (GSS) only; 
\item GSS separation plus post-filter; 
\item GSS separation plus post-filter and missing feature mask. 
\end{enumerate}
Again, no tests are presented using only one microphone because in
this case, the system would have no directional information at all
to distinguish between the speakers (it would recognise the same digits
for all sources). The speech recognition accuracy on the clean (non-mixed)
data is usually very high (>95\%), so it is not reported either.

\begin{figure}[th]
\begin{center}\includegraphics[width=0.8\columnwidth,keepaspectratio]{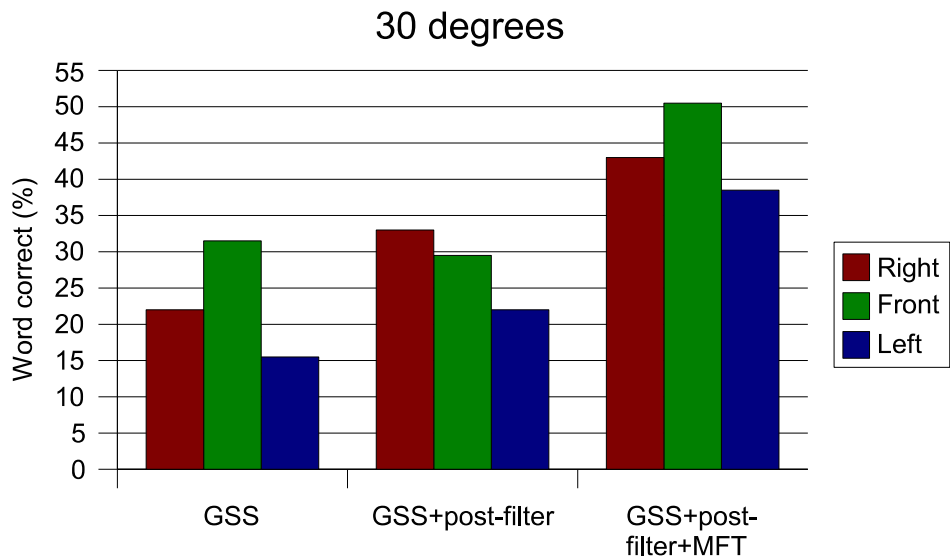}\end{center}

\caption{Speech recognition accuracy results for $30^{\circ}$ separation between
speakers.\label{cap:MFT30}}
\end{figure}

\begin{figure}[th]
\begin{center}\includegraphics[width=0.8\columnwidth,keepaspectratio]{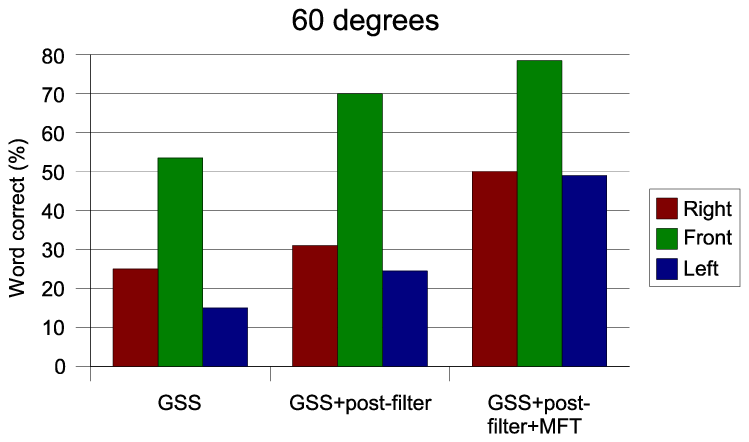}\end{center}

\caption{Speech recognition accuracy results for $60^{\circ}$ separation between
speakers.\label{cap:MFT60}}
\end{figure}

\begin{figure}[th]
\begin{center}\includegraphics[width=0.8\columnwidth,keepaspectratio]{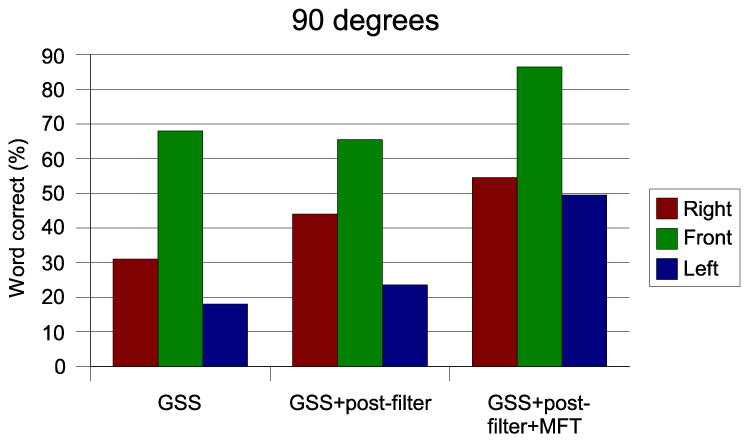}\end{center}

\caption{Speech recognition accuracy results for $90^{\circ}$ separation between
speakers.\label{cap:MFT90}}
\end{figure}

The test set contains 200 isolated common Japanese words spoken simultaneously
by two male speakers and one female speaker. There is no overlap with
the training set. Results are presented in Figures \ref{cap:MFT30},
\ref{cap:MFT60} and \ref{cap:MFT90}. The relatively poor results
with GSS only are mainly due to the highly non-stationary interference
coming from the two other speakers and the fact that the microphones
placement is constrained by the robot dimensions. The post-filter
brings an average reduction in relative error rate\footnote{The relative error rate is computed as the difference in errors divided
by the number of errors in the reference setup.} of 10\% over use of GSS alone. When the post-filter is combined with
missing feature theory, the total improvement becomes 38\%. The important
difference in recognition accuracy as a function of direction (left,
centre, right) is mainly due to differences in playback level, resulting
in different SNR levels after GSS. It can also be observed from the
results that accuracy decreases as the separation between sources
decreases. This can be explained by the fact that it is more difficult
to find a demixing matrix for narrow angles, making the GSS step less
efficient. Note that these results cannot be compared to those obtained
in Section \ref{sub:ASR-Results-Direct-Integration} because the task
and experimental setup are completely different.

\begin{comment}
\begin{table}[th]
\caption{Speech recognition accuracy results for $30^{\circ}$interval \label{cap:Speech-recognition-accuracy30}}

\centering{}%
\begin{tabular}{|c|c|c|c|}
\hline 
Source position &  left ($-30^{\circ}$) & centre ($0^{\circ}$) &  right ($+30^{\circ}$)\tabularnewline
\hline 
GSS &  15.5 &  31.5 &  22.0\tabularnewline
\hline 
GSS+post-filter &  22.0 &  29.5 &  33.0\tabularnewline
\hline 
GSS+post-filter+MFT &  38.5 &  50.5 &  43.0 \tabularnewline
\hline 
\end{tabular}
\end{table}

\begin{table}[th]
\caption{Speech recognition accuracy results for $60^{\circ}$interval\label{cap:Speech-recognition-accuracy60}}

\centering{}%
\begin{tabular}{|c|c|c|c|}
\hline 
Source position &  left ($-60^{\circ}$) & centre ($0^{\circ}$) &  right ($+60^{\circ}$)\tabularnewline
\hline 
GSS &  15.0 &  53.5 &  25.0\tabularnewline
\hline 
GSS+post-filter &  24.5 &  70.0 &  31.0\tabularnewline
\hline 
GSS+post-filter+MFT &  49.0 &  78.5 &  50.0 \tabularnewline
\hline 
\end{tabular}
\end{table}

\begin{table}[th]
\caption{Speech recognition accuracy results for $90^{\circ}$interval\label{cap:Speech-recognition-accuracy90}}

\centering{}%
\begin{tabular}{|c|c|c|c|}
\hline 
Source position &  left ($-90^{\circ}$) & centre ($0^{\circ}$) &  right ($+90^{\circ}$)\tabularnewline
\hline 
GSS &  18.0 &  68.0 &  31.0\tabularnewline
\hline 
GSS+post-filter &  23.5 &  65.5 &  44.0\tabularnewline
\hline 
GSS+post-filter+MFT &  49.5 &  86.5 &  54.5 \tabularnewline
\hline 
\end{tabular}
\end{table}
\end{comment}

\section{Discussion\label{sec:ASR-Integration-Discussion}}

In this chapter, we demonstrate a complete multi-microphone speech
recognition system that integrates all stages of source separation
and recognition so as to maximise accuracy in the context of simultaneous
speakers. We also demonstrate that the multi-source post-filter described
in Section \ref{sec:Post-filter} can be optionally used to enhance
speech recognition accuracy by providing feature reliability information
(in the form of a missing feature mask) to a missing feature theory-based
ASR. This optional missing feature mask is designed so that only spectral
regions dominated by interference are marked as unreliable. 

When the separation system is coupled directly to the ASR without
the use of a missing feature mask, the speech recognition rate is
on average 83\% for three simultaneous speakers and 90\% for two simultaneous
speakers. The result also clearly demonstrates the importance of the
post-filter and reverberation cancellation for reverberant environments.
Moreover, the results obtained are equal or better than those obtained
by human listeners in the same conditions. Good results are also obtained
when the speakers are moving.

When using a missing feature theory-based ASR, we have obtained a
reduction of 38\% (relative) in error rate compared to separation
with GSS only. In this case, the post-filter alone contributes to
a 10\% reduction in error rate. This shows that speech recognition
on simultaneous speakers can be enhanced when using the missing feature
framework in conjunction with a multi-source post-filter. 

\begin{comment}
The ASR is able to perform recognition tasks that listeners could
not do. 

In future work, we plan to perform the speech recognition with moving
speakers and increase the system's robustness in reverberant environments,
in the hope of developing new capabilities for natural communication
between robots and humans.
\end{comment}

In the future, we believe it is possible for the technique to be generalised
to the use of cepstral features (which provide better accuracy) instead
of spectral features. One way to achieve this would be to go one step
further to increase speech recognition robustness by explicitly considering
the uncertainty in the feature values. Instead of considering features
as reliable or unreliable, the acoustic model would take into consideration
both the estimated features and their variance, as proposed in \cite{Deng2002,Droppo2002}.

\chapter{Conclusion\label{cha:Conclusion}}

In this thesis, a complete auditory system for a mobile robot is presented.
The system is composed of subsystems for localising sources, separating
the different sources, as well as performing speech recognition. The
sound localisation system is composed of two parts: a steered beamformer
implemented in the frequency domain that is able to ``listen'' for
sources in all possible directions; a particle filter using a probabilistic
model to track multiple sound sources by combining past and present
estimations from the steered beamformer.

The sound source separation algorithm is also composed of two parts.
The first part is an implementation of the Geometric Source Separation
(GSS) algorithm that provides partial separation using the microphone
array audio data and the location of the sources. The sources separated
by the GSS algorithm are then sent to the second part, a multi-source
post-filter, that uses a log-domain spectral estimator to further
remove noise, interference and reverberation. The novelty of our post-filter
is that it estimates the spectrum of interferences by using the other
separated sources. 

The output of the sound source separation subsystem is used for performing
speech recognition. The audio can be sent directly to an ASR for recognition.
Optionally, it is possible to use internal values from the post-filter
to compute a missing feature mask, representing the reliability of
the spectral features. That mask can then be sent to the ASR so as
to ignore unreliable features and increase recognition rate.

We can now comment on the original mobile robotics constraints imposed
on the system, as outlined in Chapter \ref{cha:Overview-of-the-Thesis}:
\begin{itemize}
\item \textbf{Limited computational capabilities}. The system is able to
function in real-time consuming between 40\% and 70\% (depending on
the number of sources present) of the CPU cycles on a Pentium-M 1.6
GHz. Using machine-dependant optimisations could probably help to
further reduce the computational requirements.
\item \textbf{Need for real-time processing with reasonable delay}. The
delay added by the localisation system is 50 ms, which is small enough
for any application (and no longer than typical human reflexes). The
delay for the separation is even less, so it is only limited by the
delay in localisation information. In cases where it is needed, an
additional delay of 500 ms to the localisation algorithm also provides
smoother tracking, while still making the (rougher) short-delay estimations
available for source separation and fast reflex actions.
\item \textbf{Weight and space constraints}. The additional weight of the
system is composed only of the eight microphones (very light), the
soundcard and the embedded computer. Of these, the soundcard and computer
could be made even smaller by the use of a DSP card. The space constraints
are also respected, since each microphone is a 2.5 cm square (could
be even further reduced by using smaller components) and can be placed
directly on the surface of the robot's frame, without having to modify
the robot in any way.
\item \textbf{Noisy operating environment (both point source, diffuse sources,
robot noise)}. We have shown that the system works in noisy environments,
with several people talking at the same time. Also, the system is
shown to perform reliably even in a room where the reverberation time
is 1 second.
\item \textbf{Mobile sound sources}. The system is tested in conditions
where the sound sources are moving and is still able to perform both
tracking and separation on those sources.
\item \textbf{Mobile reference system (robot can move)}. We demonstrate
that the system functions when the robot is moving. Also, because
no absolute reference is used, this case is equivalent to the case
where all sources are moving (the system does not know whether the
robot or the sources are moving).
\item \textbf{Adaptability}. The system is demonstrated in two different
configurations on the Spartacus robot. Also, a collaboration with
Kyoto University has shown that the system could work on the SIG2
robot without any modification.
\end{itemize}
In light of this, we can say that all the goals set for a mobile robot
auditory system are successfully achieved.

\section{Future Work}

An important aspect of speech-based human-robot interaction that has
not been addressed in this thesis is dialog management. While it is
important for a robot to be able to recognise speech in real-life
environments, it is equally important for the robot to know what to
say and what to expect as a response. The latter means that the robot
should be able to update its speech recognition context (vocabulary,
grammar and/or language model) dynamically. Also, it would be desirable
for a robot to have good conversational and socialising skills. As
part of a more complete dialog system, it would be interesting for
the robot to still be able to listen even when it is speaking. This
would require using echo cancellation to subtract the speech from
the robot from the signal captured at the microphones.

Since the current system is able to recognise what is being said by
a speaker, the logical extension would be to recognise who the speaker
is. This would involve running speaker identification algorithms on
the separated outputs. Also, it would be useful to recognise non-speech
sounds occurring in the environment. These include people walking,
doors closing, phones ringing, etc. This could be used to direct the
attention of the robot, e.g. by directing a camera to events occurring
in its surroundings.

Another area of research in mobile robot audition would be to apply
human-inspired audition techniques with the use of a microphone array.
This includes making use of the head-related transfer function (HRTF)
along with interaural intensity difference (IID) to obtain additional
information about the signals. In the same spirit, it would be interesting
to adapt computational auditory scene analysis (CASA) and neural based
techniques \cite{Pichevar2004} to the use of microphone arrays.

Finally, on a more practical aspect, a useful improvement would consist
of implementing the system described in this thesis on a DSP platform.
This would allow integration with a larger variety of robots by alleviating
the size and computational requirements.

\section{Perspectives}

While the system developed for this thesis is especially targeted
at mobile robotics, many of the advancements made are also applicable
to other fields or research areas. First, the localisation algorithm
developed can be easily adapted to a video-conferencing application.
It would thus allow the camera to automatically follow the person
speaking, even if that person is moving, as is done by Mungamuru and
Aarabi \cite{Mungamuru2004}.

The sound separation system can also be applied to different fields.
The robustness to noise would make it suitable for use in noisy environments
(such as automobiles) in order to improve speech recognition accuracy.
The ability to perform recognition simultaneously on different speakers
would also make it possible to generate a real-time text version of
a meeting (e.g., for deaf participants), as well as performing automatic
generation of meeting minutes or lecture transcription. Also, the
localisation information combined with speech recognition results
would allow automatic annotation of audio content and could be used
in conjunction with the Annodex\footnote{http://www.annodex.net/}
annotation format \cite{Pfeiffer2003}.

Now that we have demonstrated that it is possible for a mobile robot
to have auditory capabilities close to that of humans (and in some
cases maybe better), we believe that mobile robots will be able to
interact more naturally with humans in an unconstrained environments.
This assertion will soon be verified in practice as the Spartacus
robot will attend the 2005 AAAI conference as part of the AAAI challenge\footnote{http://palantir.swarthmore.edu/aaai05/robotChallenge.htm}.

\singlespacing\bibliographystyle{/home/jm/phd/jmvalin/doc/these/IEEEtran}
\bibliography{iros,BiblioAudible,localize,pub2}

\end{document}